\documentclass[10pt]{article} 
\usepackage[accepted]{tmlr}

\usepackage{amsmath,amsfonts,bm}

\def\eqref#1{equation~\ref{#1}}

\def\1{\bm{1}}

\DeclareMathAlphabet{\mathsfit}{\encodingdefault}{\sfdefault}{m}{sl}
\SetMathAlphabet{\mathsfit}{bold}{\encodingdefault}{\sfdefault}{bx}{n}

\usepackage{microtype}
\usepackage{graphicx}
\usepackage{subfigure}
\usepackage{booktabs} 

\usepackage{hyperref}
\usepackage{url}

\usepackage{amsmath}
\usepackage{amssymb}
\usepackage{mathtools}
\usepackage{amsthm}

\usepackage[capitalize]{cleveref}
\crefname{figure}{Fig.}{Figs.}
\Crefname{figure}{Fig.}{Figs.}

\crefname{table}{Tab.}{Tabs.}
\Crefname{table}{Tab.}{Tabs.}

\crefname{section}{Sec.}{Secs.}
\Crefname{section}{Sec.}{Secs.}

\usepackage{latexsym}

\usepackage[T1]{fontenc}
\usepackage[utf8]{inputenc}
\usepackage{microtype}
\usepackage{inconsolata}

\usepackage{graphicx}
\usepackage{float}
\usepackage{amssymb}  
\usepackage{pifont}   
\usepackage{inconsolata}
\usepackage{booktabs}
\usepackage{subcaption} 
\usepackage[export]{adjustbox}
\usepackage{listings}

\usepackage{xcolor} 
\usepackage{colortbl}
\usepackage[most]{tcolorbox}
\tcbset{
    fontupper=\ttfamily\small,    
    fontlower=\ttfamily\small,    
    halign=left    
}

\usepackage{mdframed} 
\usepackage{marginnote} 
\usepackage{xspace}
\usepackage{soul} 
\usepackage{tikz}
\usepackage{mdframed}
\usepackage{cleveref}
\usepackage{enumitem}
\usepackage{fontawesome}
\usepackage{multirow}
\usepackage{makecell}
\usepackage{longtable}

\definecolor{lightgray}{HTML}{EFEFEF}
\definecolor{darkgray}{HTML}{4D4D4D}
\definecolor{sophieblue}{HTML}{ADD8E6}
\definecolor{nephewyellow}{HTML}{FFE4B5}
\definecolor{blockpurple}{HTML}{DDA0DD}
\definecolor{toyteal}{HTML}{40E0D0}
\definecolor{greenhighlight}{HTML}{98FB98}
\definecolor{peachorange}{HTML}{FFDAB9} 
\definecolor{lightblue}{HTML}{B0E0E6}

\definecolor{fact1}{HTML}{BAE6FF}  
\definecolor{fact2}{HTML}{FFD9BE} 
\definecolor{fact3}{HTML}{d4bbff} 
\definecolor{fact4}{HTML}{FFD6E8} 
\definecolor{fact5}{HTML}{fddc69} 
\definecolor{fact6}{HTML}{A7F0BA} 
\definecolor{fact7}{HTML}{c6c6c6} 
\definecolor{fact8}{HTML}{FFAB47}  
\definecolor{fact9}{HTML}{33D6FF} 
\definecolor{fact10}{HTML}{FF5733}
\definecolor{fact11}{HTML}{FF5733}

\definecolor{derive1}{HTML}{E6F7FF}  
\definecolor{derive2}{HTML}{FFEBD6}  
\definecolor{derive3}{HTML}{E9DEFF}  
\definecolor{derive4}{HTML}{FFEEF5}  
\definecolor{derive5}{HTML}{FFF6D6}  

\newcommand{\xmlfactopen}[1]{%
  \ifnum#1<5
    \textcolor{fact#1!300}{<fact#1>}%
  \else
    \textcolor{fact#1!100}{<fact#1>}%
  \fi
}

\newcommand{\xmlfactclose}[1]{%
  \ifnum#1<5
    \textcolor{fact#1!300}{</fact#1>}%
  \else
    \textcolor{fact#1!100}{</fact#1>}%
  \fi
}

\definecolor{ForestGreen}{RGB}{34, 139, 34}
\definecolor{Amethyst}{HTML}{232843} 
\definecolor{LimeGreen}{HTML}{5D616E} 
\definecolor{DarkOrange}{HTML}{A25DBF}
\definecolor{DeepSkyBlue}{HTML}{CC224D}
\definecolor{GoldenYellow}{HTML}{F06FB0} 
\definecolor{darkorange}{RGB}{255, 140, 0} 

\definecolor{HoTColor}{HTML}{F06FB0} 
\definecolor{CoTColor}{HTML}{232843} 
\definecolor{RQColor}{HTML}{5D616E} 
\definecolor{TQColor}{HTML}{A25DBF} 
\definecolor{TAColor}{HTML}{CC224D} 

\newmdenv[
  skipabove=\topsep,
  skipbelow=\topsep,
  middlelinewidth=1pt,
  roundcorner=5pt,
  frametitlebackgroundcolor=gray!20,
  backgroundcolor=gray!5
]{custommdframed}

\newcommand{\subsec}[1]{\noindent\textbf{#1}~~}

\newcommand{\correct}[1]{\textcolor{ForestGreen}{\texttt{#1}}}
\newcommand{\incorrect}[1]{\textcolor{red}{\texttt{#1}}}

\newcommand{\increase}[1]{(\textcolor{ForestGreen}{+#1})}
\newcommand{\increasenoparent}[1]{\textcolor{ForestGreen}{+#1}}
\newcommand{\increasenoparentwithspace}[1]{\hspace{9mm} \textcolor{ForestGreen}{+#1}}
\newcommand{\decrease}[1]{(\textcolor{red}{-#1})}
\newcommand{\decreasenoparent}[1]{\textcolor{red}{-#1}}
\newcommand{\decreasenoparentwithspace}[1]{\hspace{9mm} \textcolor{red}{-#1}}

\newcommand{\greencheck}{\textcolor{ForestGreen}{\ding{51}}}
\newcommand{\redcross}{\textcolor{red!70!black}{\ding{55}}}

\newcommand{\eg}{e.g.\xspace}
\newcommand{\ie}{i.e.\xspace}
\newcommand{\hot}{HoT\xspace}

\newcommand{\hotcolored}{\textcolor{HoTColor}{\hot}\xspace}

\newcommand{\cotcolored}{\textcolor{CoTColor}{CoT}\xspace}
\newcommand{\rqcolored}{\textcolor{RQColor}{R-Q}\xspace}
\newcommand{\tqcolored}{\textcolor{TQColor}{T-Q}\xspace}
\newcommand{\tacolored}{\textcolor{TAColor}{T-A}\xspace}

\newcommand{\llmjudge}{LLM-as-a-Judge\xspace}

\newcommand{\model}[1]{{\texttt{#1}\xspace}}
\definecolor{gpt_green}{RGB}{22,163,127} 
\definecolor{gemini_blue}{RGB}{81,134,209} 
\definecolor{sonnet3_brown}{RGB}{204,154,123} 
\definecolor{sonnet35_brown}{RGB}{216, 119, 87} 
\definecolor{qwen_violet}{RGB}{191, 123, 234} 

\newcommand{\geminiflash}{\model{Gemini-\textcolor{gemini_blue}{1.5}-Flash}\xspace}
\newcommand{\geminipro}{\model{Gemini-\textcolor{gemini_blue}{1.5}-Pro}\xspace}
\newcommand{\geminilogo}{\raisebox{-1pt}{\includegraphics[scale=0.82]{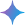}}\xspace}

\newcommand{\geminiflashlogo}{\raisebox{-1pt}{\includegraphics[scale=0.04]{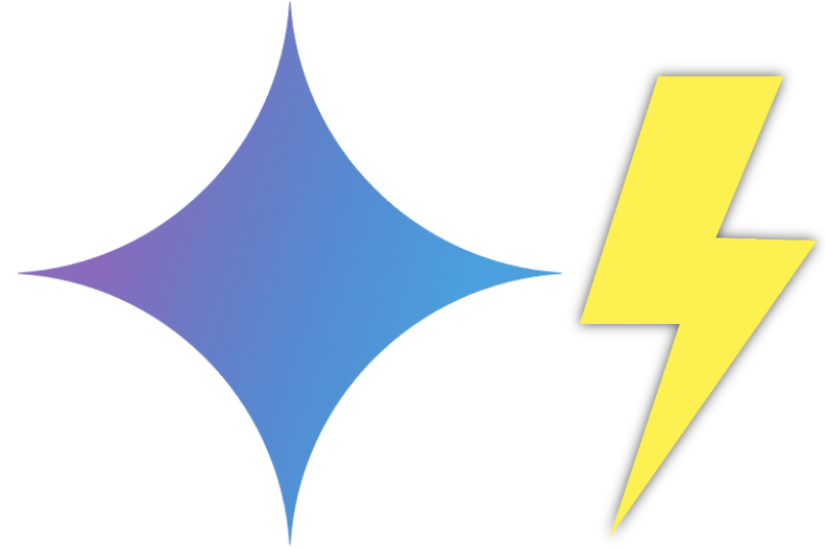}}\xspace}

\newcommand{\llamaA}{\model{Llama-\textcolor{gemini_blue}{3.1}-8B}\xspace}
\newcommand{\llamaB}{\model{Llama-\textcolor{gemini_blue}{3.1}-70B}\xspace}
\newcommand{\llamaC}{\model{Llama-\textcolor{gemini_blue}{3.1}-405B}\xspace}

\newcommand{\llamaSmalllogo}{\hspace{0em}\raisebox{-1pt}{\includegraphics[scale=0.028]{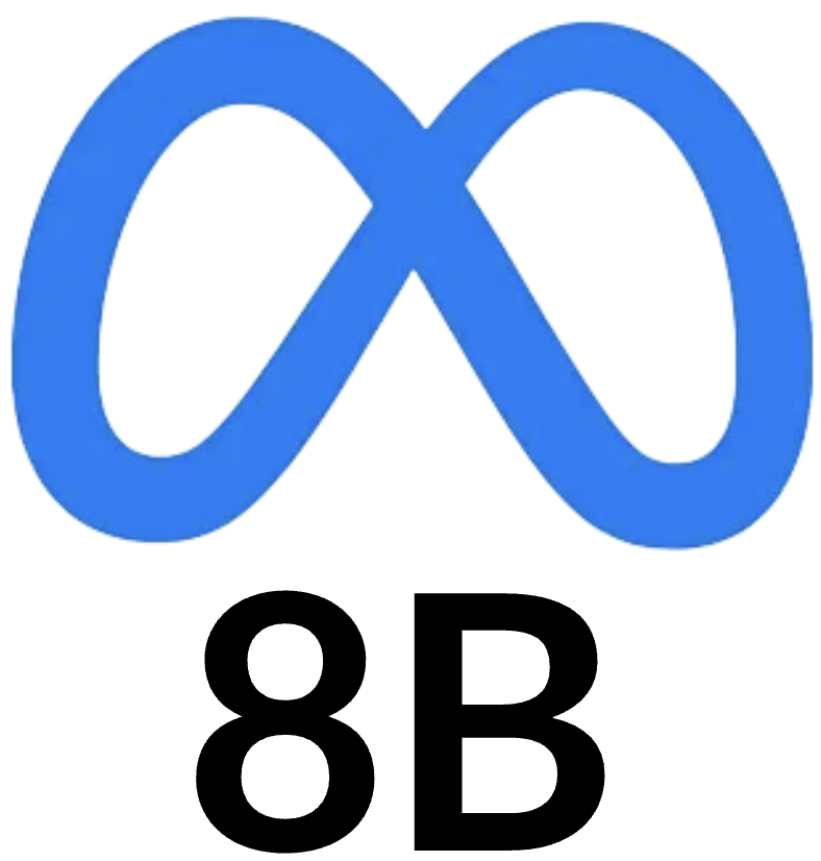}}\xspace}

\newcommand{\llamaMedlogo}{\hspace{0em}\raisebox{-1pt}{\includegraphics[scale=0.028]{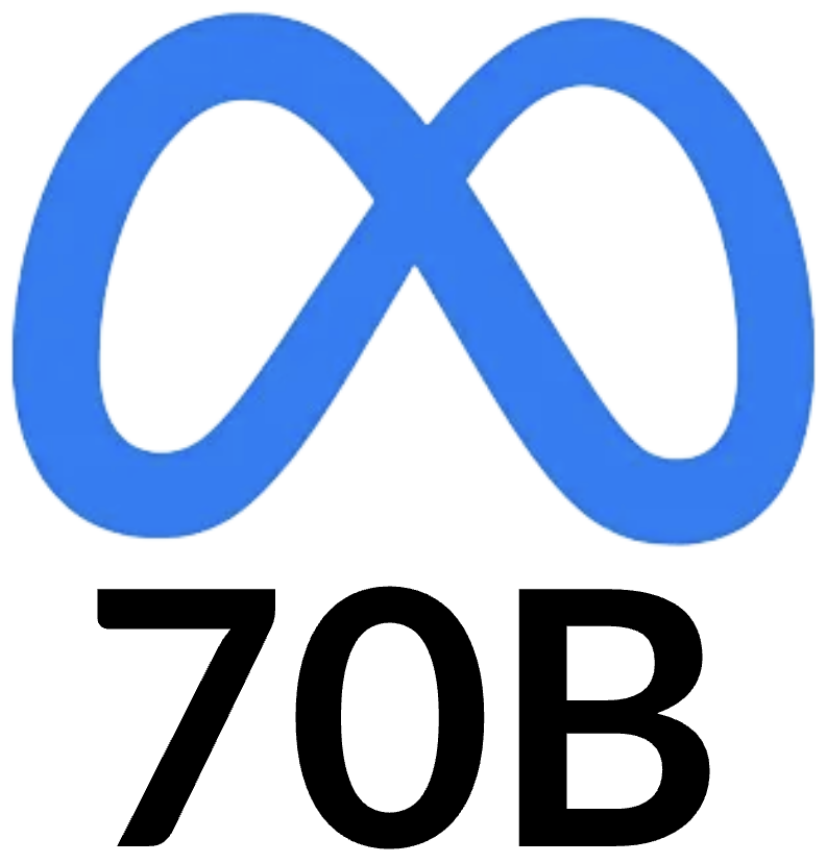}}\xspace}

\newcommand{\llamaBiglogo}{\hspace{0em}\raisebox{-1pt}{\includegraphics[scale=0.028]{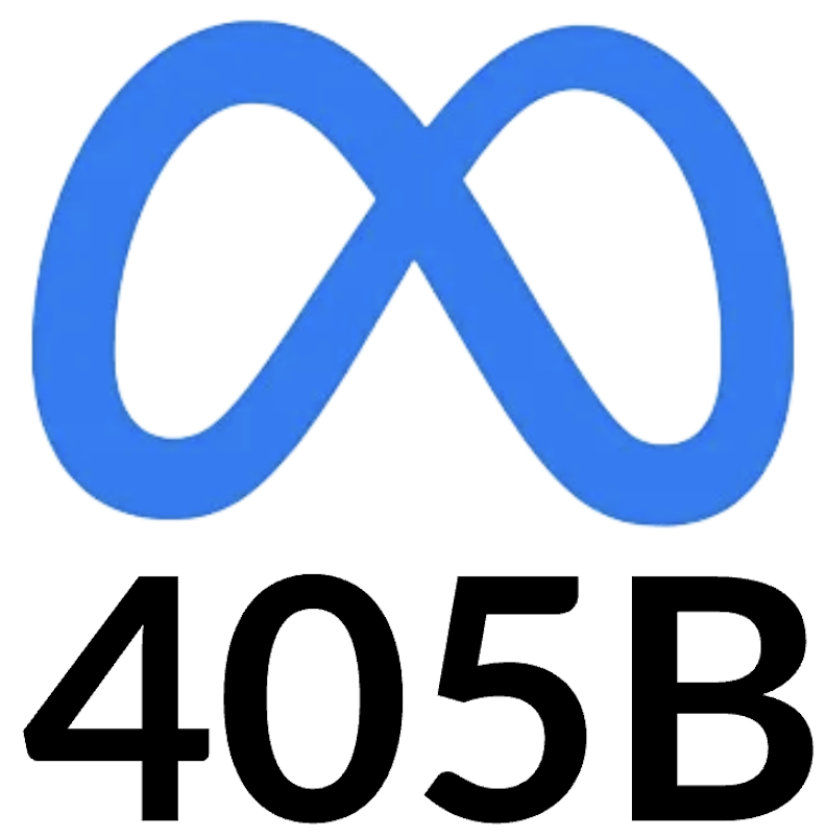}}\xspace}

\newcommand{\gpto}{\model{GPT-\textcolor{gpt_green}{4o}}\xspace}
\newcommand{\gptmini}{\model{GPT-\textcolor{gpt_green}{4o-mini}}\xspace}
\newcommand{\gptlogo}{\raisebox{-1pt}{\includegraphics[scale=0.02]{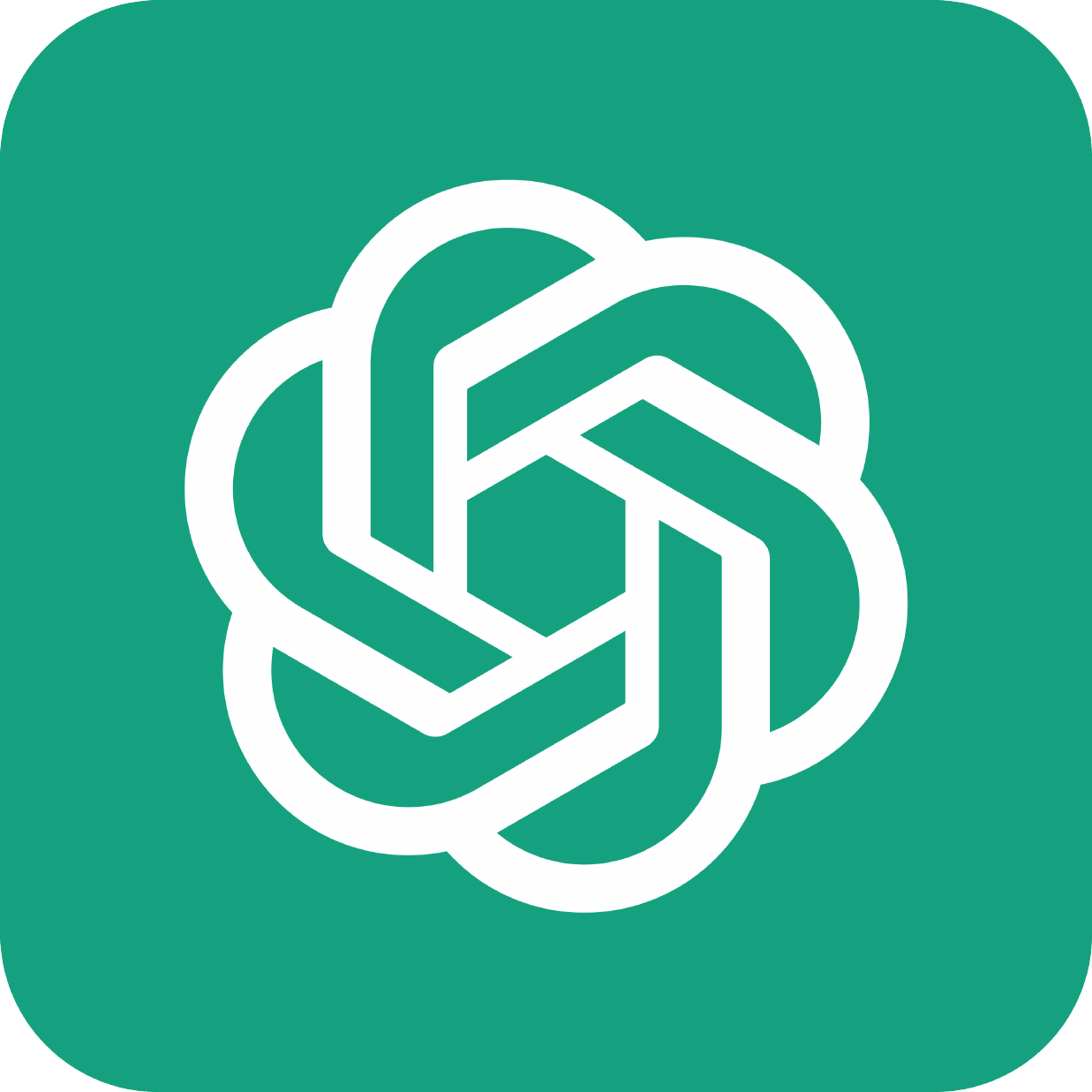}}\xspace}

\newcommand{\gptminilogo}{\raisebox{-1pt}{\includegraphics[scale=0.034]{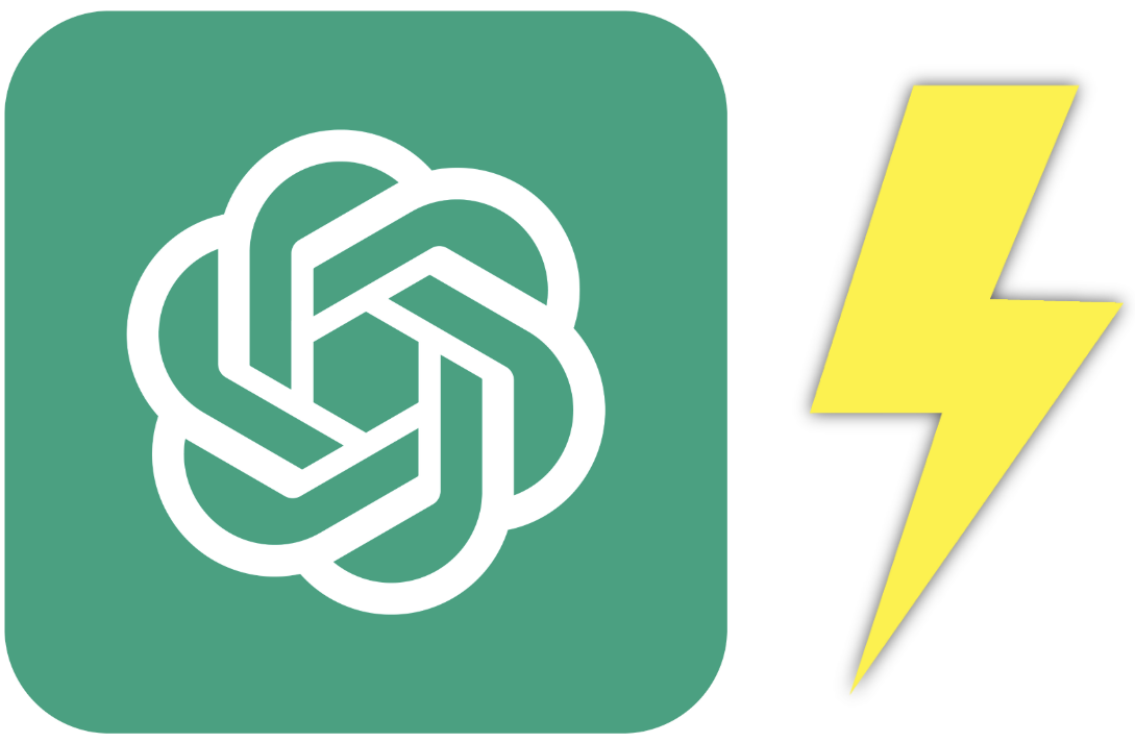}}\xspace}

\newcommand{\qwen}{\model{Qwen-\textcolor{qwen_violet}{2.5}-Coder-32B}\xspace}

\newcommand{\qwenlogo}{{\includegraphics[scale=0.034]{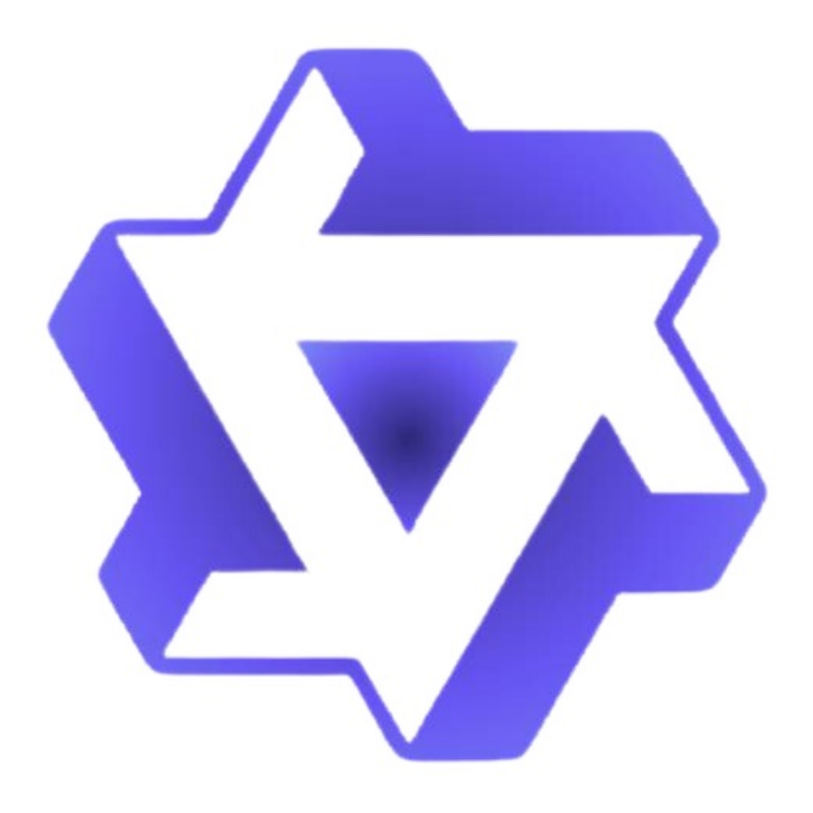}}\xspace}

\newcommand{\robot}{\raisebox{-2pt}{\includegraphics[scale=0.13]{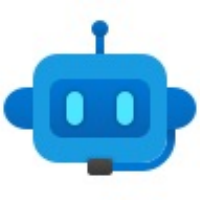}}\hspace{0.3em}\xspace}
\newcommand{\human}{\raisebox{-2pt}{\includegraphics[scale=0.13]{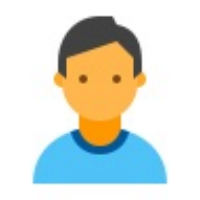}}\hspace{0.3em}\xspace}

\newcommand{\deepseek}{\raisebox{-1pt}{\includegraphics[scale=0.048]{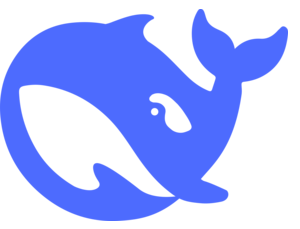}}\xspace}

\newcommand{\llamavision}{\raisebox{-1pt}{\includegraphics[scale=0.038]{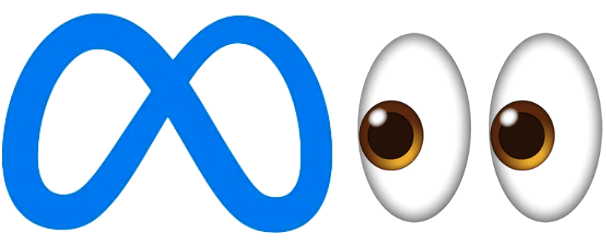}}\xspace}

\newcommand{\qwq}{QwQ-32B-Preview\xspace}

\newcommand{\projectrepo}{\href{https://github.com/anguyen8/hot}{Github}\xspace}

\definecolor{yellow_sizes}{RGB}{255, 252, 130}

\newcommand{\cleanpapertitle}{HoT: Highlighted Chain of Thought for Referencing Supporting Facts from Inputs}

\usepackage{hyperref}
\usepackage{url}

\title{HoT: Highlighted Chain of Thought for Referencing Supporting Facts from Inputs}

\author{\name Tin Nguyen\thanks{Equal contribution} \email ttn0011@auburn.edu \\
      \addr Auburn University
      \AND
      \name Logan Bolton\footnotemark[1] \email ldb0046@auburn.edu \\
      \addr Auburn University
      \AND
      \name Mohammad Reza Taesiri \email mtaesiri@gmail.com \\
      \addr University of Alberta
      \AND
      \name Trung Bui \email bhtrung@gmail.com\\
      \addr 
      \AND
      \name Anh Totti Nguyen\thanks{Corresponding author} \email anh.ng8@gmail.com \\
      \addr Auburn University}

\def\month{06}  
\def\year{2026} 
\def\openreview{\url{https://openreview.net/forum?id=abm6pDTbT1}} 

\begin{document}

\maketitle

\begin{abstract}
An Achilles heel of Large Language Models (LLMs) is their tendency to hallucinate non-factual statements. 
A response mixed of factual and non-factual statements poses a challenge for humans to verify and accurately base their decisions on. To combat this problem, we propose Highlighted Chain-of-Thought Prompting (HoT), a technique for prompting LLMs to generate responses with XML tags that ground facts to those provided in the question.
That is, given an input question, LLMs would first re-format the question to add XML tags highlighting key facts, and then, generate a response with highlights over the facts referenced from the input. 
Compared to vanilla chain of thought prompting (CoT), HoT reduces the rate of hallucination and separately improves LLM accuracy consistently on over \textbf{22 tasks} from arithmetic, reading comprehension, to logical reasoning.
When asking humans to verify LLM responses, highlights help time-limited participants to more accurately and efficiently recognize when LLMs are correct.
Yet, surprisingly, when LLMs are wrong, HoTs tend to fool users into believing that an answer is correct. Code is available at: \projectrepo.
\end{abstract}

\section{Introduction}
\label{sec:introduction}




Chains of Thoughts (CoT) enable LLMs to generate step-by-step solutions to questions, improving both (1) accuracy on many tasks \citep{wei2022chain} that benefit from problem decomposition and (2) transparency in how a model arrives at a final answer.
However, a major weakness of LLMs is their tendency to hallucinate non-factual statements \citep{zhang2023siren}, making it difficult for humans to verify whether an LLM answer is correct.

\begin{figure}[t]   
    \centering
    \includegraphics[width=\columnwidth]{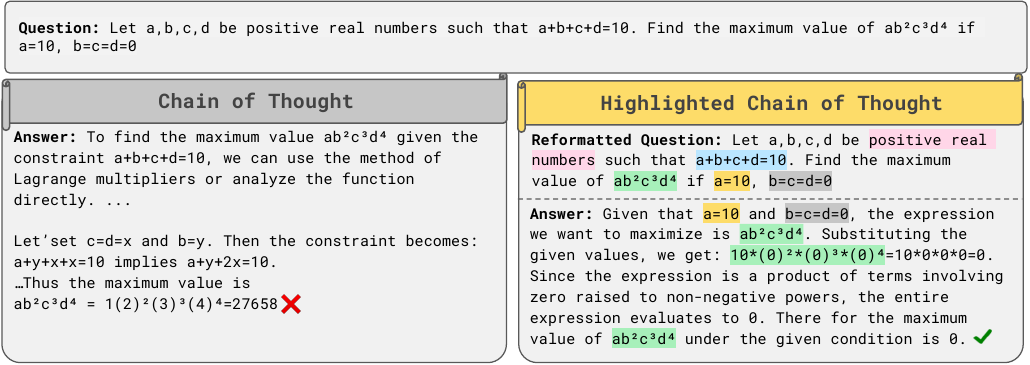}
    \caption{
    CoT and HoT (ours) responses for a MATH500 question in ReasoningTrap benchmarks \citep{jang2025reasoning}, both generated by \geminiflash~\geminiflashlogo. \textbf{Left}: CoT misses the key constraint \sethlcolor{fact7}\hl{$b=c=d=0$}, giving an incorrect answer. \textbf{Right}: HoT (a re-formatted question and answer) applies the key constraint \sethlcolor{fact7}\hl{$b=c=d=0$} to the expression $ab^2c^3d^4$, yielding the correct answer of 0. The full reasoning traces of both methods are provided in \cref{tab:gcot_vs_cot_math500_conditioned_math}.
    }
    \label{fig:teaser}
\end{figure}

Based on this observations, we aim to mitigate it by proposing Highlighted Chain-of-Thought (\hot), a prompting technique that instructs LLMs to generate a CoT answer but with in-line XML tags, grounding in-response facts to in-question facts (\cref{fig:teaser}).
We hypothesize that HoT may improve LLM accuracy on downstream tasks and also user verification accuracy.

Existing methods attempt to combat hallucination and improve verifiability by forcing LLMs to cite websites \citep{perplexityai, chatgptsearch}, documents \citep{bai2024longcite} or paragraphs \citep{cohen2024contextcite} to support statements in the response.
Yet, there is no work that enables LLMs to generate regular CoTs but with references back to the in-question facts.


In \hot, first, an LLM re-formats the input question to wrap XML tags around key facts.
Second, it generates its response but with XML tags around the facts that come from the input, enabling colored highlights (\cref{fig:teaser}).
Such highlights enable users to trace which statements in the response correspond to which facts in the input, which we hypothesize to make human verification faster and more accurate.

We test \hot on \textbf{5} LLMs including \geminiflash (\geminiflashlogo), \geminipro (\geminilogo) \citep{reid2024gemini}, \llamaB (\llamaMedlogo), \llamaC (\llamaBiglogo) \citep{dubey2024llama}, and \gpto (\gptlogo) \citep{gpt4o} across \textbf{22} tasks from arithmetic, logical reasoning, reading comprehension, long-context to puzzle, conditioned maths questions.
Our main findings are:

\begin{enumerate}[itemsep=0pt, parsep=0pt]

\item HoT substantially reduces hallucination rate compared to CoT (from 21.22\% to 14.92\%) across multiple datasets, as measured by SelfCheckGPT \citep{manakul-etal-2023-selfcheckgpt} (\cref{sec:hallucination}).

\item Compared to few-shot CoT, \hot consistently improves LLM accuracy across \textbf{5 LLMs} and \textbf{22 datasets}. 
On average, across arithmetic, question-answering, logical reasoning, long-context, and puzzle tasks, \hot gains \increasenoparent{2.61}, \increasenoparent{2.58}, \increasenoparent{2.53}, \increasenoparent{2.03}, \increasenoparent{11.38} percentage points (pp), respectively (\cref{sec:main_results}).

\item HoT outperforms other advanced prompting techniques (Least-to-Most (LtM) \citep{zhou2022least}, Tree-of-Thought (ToT) \citep{yao2023tree}, Self-Refine \citep{madaan2023self}, and Chain-of-Verification (CoVE) \citep{dhuliawala2023chain}) on challenging datasets including r-GSM, Seven Objects, and Date. 
Furthermore, when combined with Self-Consistency \citep{wang2022self}, ComplexCoT \citep{fu2022complexity} methods, HoT achieves superior performance than each of these methods alone and HoT alone (\cref{sec:hot_beat_other_methods}).

\item \emph{Both} components of HoT, repeating the question and highlighting facts via XML, independently improve LLM accuracy (\cref{sec:prompt_variation_exp}).

\item The colored highlighting in HoT responses improves the users' speed in verifying the accuracy of LLM answers by $\sim$25\% (62.38s $\to$ 47.26s) and also their accuracy (\cref{sec:human_verification}).
\item Fine-tuning small LLMs on HoT responses not only enhances accuracy but also strengthens their ability to attend more effectively to input context compared to base and CoT fine-tuned models (\cref{sec:attention_entropy}).




\end{enumerate}

\section{Related Work}
\label{sec:related_work}

\subsec{Generating references to documents}
Recent works have trained LLMs to answer questions by generating responses that include citations to the documents from which they extract supporting information \citep{cohen2024contextcite, bai2024longcite, gao-etal-2023-enabling, press2407citeme, taylor2022galactica, bohnet2022attributed}.
Another approach is to generate citations post-hoc, \ie, by having one LLM generate the answer first and another LLM search for citations that support facts in the answers \citep{ramu2024enhancing, sancheti2024post, dasigi-etal-2021-dataset}.
Commercial LLM-powered search engines recently also rolled out their citation feature \citep{anthropic,chatgptsearch,bingsearch,perplexityai}, which references web pages and online documents that support statements in the responses.

Unlike above works, we focus on generating references to phrase-level and sentence-level facts (\cref{fig:teaser,fig:cot_vs_hot_puzzletrivial}) in the question instead of references to context documents or paragraphs.



\subsec{Prompting techniques} 
Interestingly, inputting the exact question \emph{twice} to an LLM improves its answer accuracy slightly \citep{xu-etal-2024-reading}.
Similarly, given a question, asking an LLM to first repeat the question and then answer it also improves accuracy compared to vanilla CoT \citep{mekala-etal-2024-echoprompt}.
Our work is similar 
in that we also ask LLMs to re-generate the question; however, the difference is that our re-generated question contains XML tags around key facts.
Furthermore, unlike the above two works, \hot-prompted LLMs include tags in the \emph{answer}, which further improve LLM accuracy as shown in our ablation study (\cref{sec:prompt_variation_exp}).

\subsec{Retrieval-Augmented Generation (RAG)} Some systems first retrieve relevant documents given an input question and then feed the documents along with the original question to an LLM for generating an answer \citep{asai2023selfrag, jin2024flashrag}.
While they generate references to the retrieved documents \citep{Liu_LlamaIndex_2022}, \hot generates references to fact phrases in the input question.
In RAG systems, feeding HTML tags in web pages instead of plain text improves the accuracy of the retrieved knowledge \citep{tan2024htmlrag}.
Functionally similar to HTML tags, our XML tags around key facts here make LLMs answer more accurately.



\subsec{Span prediction} is a core NLP task that requires a model to read a question and a paragraph and predict the (start, end) index of the answer embedded in the paragraph  \citep{rajpurkar-etal-2016-squad,dasigi2019quoref}.
Similarly to span prediction, we ask LLMs to identify key phrases in the input question.
However, \hot instructs LLMs to perform an extra step of generating the answer with references to the selected spans (\ie, highlighted in-question facts).


\subsec{Impact of highlights on human cognition}
In cognitive science, studies have found that selective emphasis techniques, such as text highlighting, can improve comprehension and learning for humans \citep{fowler1974effectiveness,ramirez2019understanding}. 
In contrast, inappropriate highlighting could harm human accuracy in reading comprehension \citep{gier2009harmful,ramirez2019understanding}.
To our knowledge, we are the first to study how key-fact highlights may help users verify the accuracy of an LLM's answer.

\section{Method}\label{sec:method}
Our goal is to prompt LLMs to produce a response consisting of two parts: 
(1) a version of the original question with XML tags (\texttt{<fact1>}, \texttt{</fact1>}, etc.) wrapped around key facts in the input query and 
(2) an answer that explicitly links statements in the answer to the highlighted facts in the question (\cref{fig:teaser}). 
\textbf{Our hypothesis} is that asking the LLM to decide which facts are worth wrapping in XML tags encourages the model to better attend to these important facts, thereby reducing hallucinations (\cref{sec:hallucination}) and improving the accuracy of the final answer (\cref{sec:main_results}). 
Additionally, these XML tags can be transformed through regex and CSS to become highlighted in the GUI when the LLM answer is presented to users for verification.
We experiment with different XML tag names and find that \texttt{<fact\{}\textit{i}\}\texttt{></fact}\{\textit{i}\}> where \textit{i} = 1,2,3 results in the best LLM accuracy (\cref{sec:tag_ablation}). 


\subsection{Highlighted Chain-of-Thought (\hot)}
\label{sec:hot_introduction}
\begin{figure*}[t]
\centering
    \includegraphics[width=\columnwidth]{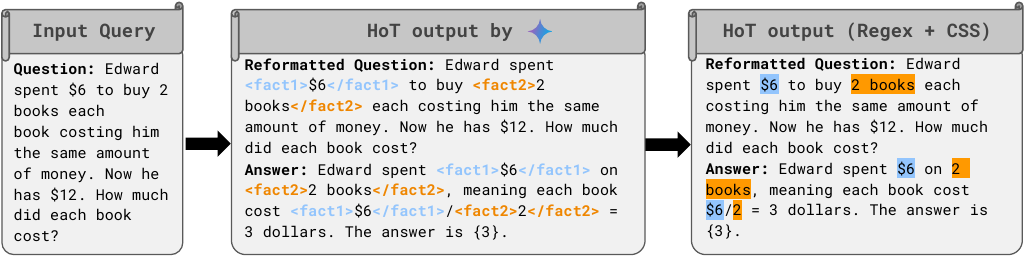}
    \caption{LLMs generate HoT responses by wrapping XML tags around the information that the model determines is the most important. 
Regex and CSS are then used to visualize the highlights for user readability (see the code to convert XML tags to highlights in \cref{sec:code}).}
\label{fig:cot_vs_gcot}
\end{figure*}

To prompt LLMs to generate HoTs (\cref{fig:cot_vs_gcot}), we design the following prompt structure (\cref{fig:prompt_structure}) and use it for all datasets.
First, the 8-shot demonstration examples (which are CoT demonstrations but with XML tags) would show LLMs how to insert tags and answer questions.
Second, the HoT instruction would be a short, explicit request that asks LLMs to insert tags into questions and answer it. 

\begin{figure}[ht!]
\begin{minipage}{0.6\textwidth}
\begin{tcolorbox}[boxrule=0pt, frame empty]
{
\texttt{\{Question 1\}} {\{Reformatted Q1\}} {\{Answer 1\}}\\
\texttt{\{Question 2\}} {\{Reformatted Q2\}} {\{Answer 2\}}\\
...\\
\texttt{\{Question 8\}} {\{Reformatted Q8\}} {\{Answer 8\}}\\
\textcolor{fact2!300}{
\texttt{\{Question\}}\\
}
\textcolor{fact1!300}{
\texttt{\{\hot Instruction\}}
}
}
\end{tcolorbox}
\end{minipage}%
\hfill
\begin{minipage}{0.35\textwidth}
\caption{8-shot \hot examples are provided in addition to the explicit directions (\textcolor{fact1!300}{\texttt{\small\hot Instruction}}) (see \cref{sec:system_prompt}) to help LLMs understand the expected format. 
See \cref{tab:full_fewshot_prompt_example} for one entire example prompt.}
\label{fig:prompt_structure}
\end{minipage}
\end{figure}

For each dataset, we make separate 8-shot examples \citep{fewshot} to demonstrate tag insertion, as instructing LLMs to do this zero-shot is non-trivial, especially since different datasets may require highlighting different types of linguistic structures.




\subsection{Annotating HoT in-context examples}

Generating demonstrations (\ie, few-shot HoT in-context examples) for each dataset from scratch can be time-consuming for humans.
Therefore, we propose a 3-step, LLM-assisted approach for humans to quickly create high-quality HoT demonstrations for a given dataset, and apply the same procedure for all datasets. 
We begin by manually creating 15 \hot question-answer pairs across multiple domains. 
These HoT examples will then be used to generate extra demonstrations tailored to each dataset.

\subsec{Step 1: Humans to insert highlights into HoT meta demonstrations} We take, in total, 15 different CoT question-answer pairs from multiple datasets in a variety of domains including arithmetic, logical reasoning and question-answering tasks.
Then, we manually \textbf{add XML tags to these questions} following the leave-one-out principle \citep{li2016understanding} for identifying important tokens.
That is, a fact (to be wrapped around, \eg, \texttt{<fact1></fact1>}) is a key phrase in the question that when removed would render the question unanswerable. 
Any details that are not directly relevant to answering the final question should not be tagged.




After tagging all key phrases in the question, we examine the answer to find semantically matching phrases referring to the same entity (\eg, \sethlcolor{fact6}\hl{``$ab^2c^3d^4$''} vs. \sethlcolor{fact6}\hl{``$10*(0)^2*(0)^3*(0)^4$''} in \cref{fig:teaser}) and surround them with XML tags.
That is, every tag in the answer must correspond to an existing tag in the question.
These 15 human-annotated examples (see \projectrepo repo) are then used as \textbf{meta demonstrations} for LLMs to generate few-shot demonstrations for a specific dataset (Step 2).

\subsec{Step 2: LLMs to generate CoT responses for 8 questions in a given dataset} To fairly compare HoT and CoT under the common 8-shot CoT setting, we need to create 8 HoT demonstrations for each dataset.
To do that, we 
(a) ask \gpto to generate standard CoT answers for 8 random questions in each dataset; and 
(b) convert these CoT examples into HoT examples in Step 3, leveraging the 15 meta demonstrations (generated in Step 1).

\subsec{Step 3: LLMs to insert XML tags into CoT responses for a given dataset} We prompt \gpto with the 15 demonstrations from Step 1 as few-shot examples, along with a \textbf{question} from Step 2, to generate a reformatted question containing XML tags (see prompts in \cref{sec:annotation_app}), producing 8 tagged questions.
Finally, using the same 15 demonstrations and each tagged question, we instruct \gpto to generate a tagged answer, resulting in 8 tagged question-answer pairs per dataset.

\section{Evaluation over 22 benchmarks and 5 LLMs}
\label{sec:datasets}
We evaluate our method across \textbf{22 tasks} spanning five domains: arithmetic, question answering, logical reasoning, reading comprehension, and hard, long-context benchmarks. 
See \cref{sec:datasets_app} for more dataset details (\eg, number of instances).
We test \geminiflashlogo, \geminilogo, \llamaMedlogo, \llamaBiglogo, and \gptlogo using their default temperatures.
The configuration details of each model are listed in \cref{sec:llm_providers_details}.

\paragraph{4 Arithmetic} We test on arithmetic tasks taken from \citep{wei2022chain}: SVAMP \citep{patel2021nlp}, and AQUA \citep{ling2017program}. We also examine the performance of our method on R-GSM \citep{chen2024premise}, which focuses on changing the question premise order to challenge LLM reading comprehension in an arithmetic setting. Finally, we evaluate on GSM-Symbolic dataset \citep{gsmsymbolic}, which changes the original GSM8K questions through a symbolic template (\ie changing the numerical values). 

\subsec{5 Logical reasoning} We choose five tasks from BigBench Hard \citep{bigbenchhard}: Logical Deduction Five Objects, Logical Deduction Seven Objects, Reasoning about Colored Objects, Causal Judgement, and Navigate. 
For brevity, we refer to these datasets as Five Objects, Seven Objects, Colored Objects, Judgement, and Navigate. 

\subsec{3 Question answering} We choose StrategyQA \citep{strategyQA}, SpartQA \citep{mirzaee2021spartqa}, and Date \citep{bigbenchhard} to evaluate our method on question-answering tasks.

\subsec{2 Reading comprehension} We use the Break and Census subsets of the DROP reading comprehension benchmark \citep{dua-etal-2019-drop}. 

\subsec{5 Hard, long-context}To validate \hot on more challenging and long-context tasks, we evaluate \hot on GPQA Diamond \citep{rein2024gpqa} and four BigBench Extra Hard (BBEH) \citep{kazemi2025big} tasks: Time Arithmetic, Spatial Reasoning, Shuffle Objects, and Causal Judgement.

\subsec{3 ReasoningTrap datasets}  
LLMs often default to familiar reasoning templates, a tendency referred to as \emph{reasoning rigidity} \citep{jang2025reasoning}.  
To evaluate whether HoT can overcome this limitation, we test on two complementary datasets.  
The first, \textbf{PuzzleTrivial}, consists of logic puzzles that are subtly modified from their original forms, requiring models to move beyond rote solution patterns.  
The second, \textbf{2 Conditioned Math}, is derived from the AIME and MATH500 benchmarks, where problems are augmented with additional constraints. This design yields mathematically challenging tasks that specifically test a model’s ability to adapt its reasoning under nonstandard conditions.


\section{Results}
\label{sec:results}
\subsection{\hot prompting makes LLMs hallucinate less and answer questions more consistently}
\label{sec:hallucination}





To better quantify and understand how the grounding effect in HoT improves accuracy, we use the SelfCheckGPT framework \citep{manakul-etal-2023-selfcheckgpt} to measure whether \hot mitigates hallucinations. This method outputs a hallucination likelihood score that is correlated with the prevalence of hallucinations in LLM-generated responses. We evaluate hallucination across five representative datasets from distinct domains to cover different types of reasoning: r-GSM and GSM-Symbolic (arithmetic), SpartQA (question answering), Break (reading comprehension), and Seven Objects (logical reasoning).

\subsec{Experiment} 
The core idea supporting SelfCheckGPT (\cref{sec:selfcheckgpt}) is that when an LLM truly knows a fact, multiple stochastic generations from the same prompt should yield consistent statements. 
In contrast, hallucinated facts are often inconsistent across multiple independent responses. Given an LLM response $R$, SelfCheckGPT generates $N$ additional samples $\{S_1, S_2, \dots, S_N\}$ from the same prompt and uses a \emph{judge} LLM (here, GPT-4o) to evaluate the consistency of each sentence $r_i$ in $R$ with respect to the $N$ samples.

We run \geminiflash\geminiflashlogo using HoT prompting on a set of five diverse benchmarks: r-GSM, GSM-Symbolic, SpartQA, Break, and Seven Objects, and compute the answers' hallucination and consistency rates.
We repeat for CoT prompting and compare the two methods.

\begin{table}[H]
\centering
\begin{minipage}{0.48\textwidth}
\centering
\caption{HoT prompting makes \geminiflash~\geminiflashlogo \textbf{hallucinate consistently less} over a diverse set of tasks.
Table shows the SelfCheckGPT hallucination scores. 
(Lower is better.)}
\resizebox{\columnwidth}{!}{
\begin{tabular}{l|ccccc|c}
\hline
Prompt & R-GSM & GSM-Symbolic & SpartQA & Break & Seven Objects & Avg \\
\hline
CoT  & 12.75 & 44.60 & 17.33 & 8.57 & 22.87 & 21.22 \\
HoT  & \textbf{7.01} & \textbf{36.18} & \textbf{16.24} & \textbf{4.88} & \textbf{10.31} & \textbf{14.92} \\
\hline
\end{tabular}
}
\label{tab:mitigate_hallucinations_table}
\end{minipage}
\hfill
\begin{minipage}{0.48\textwidth}
\centering
\caption{The rates (\%) of \emph{unanimous} responses from \geminiflash~\geminiflashlogo across 5 independent runs show that \textbf{HoT prompting makes LLMs more consistent in the final answers}.}
\resizebox{\columnwidth}{!}{
\begin{tabular}{l|ccccc}
\hline
Prompt & r-GSM & GSM-Symbolic & SpartQA & Break & Seven Objects \\
\hline
CoT & 89.55 & 89.55 & 93.75 & 84.75 & 51.20 \\
\hline
HoT & \textbf{95.00} & \textbf{93.00} & \textbf{97.00} & \textbf{87.25} & \textbf{62.40} \\
\hline
\end{tabular}
}
\label{tab:unanimous_responses}
\end{minipage}
\end{table}

\subsec{Results} 
HoT consistently reduces hallucination scores for \geminiflashlogo across all 5 datasets (from 21.22 $\rightarrow$ 14.92 in \cref{tab:mitigate_hallucinations_table}).
This improvement occurs because CoT more often misses key facts in the input question, leading to higher hallucination scores (e.g., \cref{tab:hallucination_example_rgsm,tab:hallucination_example_seven_objects}).

The relationship between hallucination reduction and accuracy increase is not necessarily linear. 
For example, on Seven Objects, \geminiflashlogo’s SelfCheckGPT hallucination score nearly halves under \hot, yet accuracy increases by only \increasenoparent{4.40} points (see \cref{tab:mitigate_hallucinations_table,tab:full_performance_logical}).
This is because a model's reasoning could be highly consistent (as measured by SelfCheckGPT), but its final answer does not match the groundtruth for a task (measured by task accuracy).
That is, hallucination reduction and accuracy measure different aspects of model behavior and are not perfectly correlated.

Given that HoT produces more unanimous responses than CoT (e.g., \increasenoparent{11.2} in Seven Objects in \cref{tab:unanimous_responses}), HoT achieves lower hallucination scores regardless of correctness, while CoT's varied responses appear more hallucinatory despite slightly lower accuracy.
\subsection{\hot prompting consistently improves LLM accuracy over CoT across 5 models \& 22 datasets}
\label{sec:main_results}

Since \hot makes \geminiflash provide much more consistent answers and hallucinate less over independent runs (\cref{sec:hallucination}), we test whether \hot can help LLMs improve \textbf{accuracy} over regular question-answering tasks.

We hypothesize that HoT prompting could encourage LLMs to identify and then leverage facts in the question in their chain of thoughts, thereby improving accuracy and reducing hallucinations.

\subsec{Experiment} We compare 8-shot CoT and 8-shot \hot on five LLMs, including the proprietary models \gpto (\gptlogo), \geminiflash (\geminiflashlogo), \geminipro (\geminilogo) and open-source models \llamaB (\llamaMedlogo), \llamaC (\llamaBiglogo) across 22 tasks. 
Both the CoT and \hot few-shot demonstration examples use the same questions and answers. 
The key difference is that \hot examples contain XML tags, while CoT examples do not.

\subsec{Results}
\hot consistently improves over CoT across most datasets and models.
On average, \hot improves LLM accuracy on arithmetic, question answering, logical reasoning, and long-context tasks by \increasenoparent{2.61} (\cref{tab:full_performance_math}), \increasenoparent{2.58} (\cref{tab:full_performance_qa_rc}), \increasenoparent{2.53} (\cref{tab:full_performance_logical}), and \increasenoparent{2.03} pp (\cref{tab:long_input_results}), respectively. Especially, HoT significantly increases LLM accuracy on three hard datasets: Puzzle and Conditioned Math (AIME, MATH500) by \increasenoparent{11.38} (\cref{tab:tricky_datasets_results}).

On recent adversarial, more challenging versions of GSM8K (\ie, r-GSM and GSM-Symbolic), \hot shows substantial improvements of \increasenoparent{3.09} and \increasenoparent{0.87} pp (\cref{tab:full_performance_math}), respectively.

\hot also outperforms CoT on benchmarks involving long and complex reasoning. Specifically, HoT achieves a \increasenoparent{5.55} over CoT on GPQA Diamond (38.38\% vs. 32.83\%) and a \increasenoparent{4.50} gain on BBEH Causal Judgement (53.00\% vs. 48.50\%) (\cref{tab:long_input_results}). In contrast, on BBEH Spatial Reasoning and BBEH Shuffle Objects, where questions are extremely long, HoT does not show a significant advantage over CoT.

Compared to CoT, HoT reduces contradictory errors by \increasenoparent{10.00} on StrategyQA, \increasenoparent{26.70} on SpartQA, and \increasenoparent{3.30} on AQUA (see \cref{tab:gcot_vs_cot_strategyqa}). Furthermore, in AQUA, CoT answers in \geminiflashlogo fail to even produce the answer in 7.7\% of cases, while HoT has no such failures.

\textbf{Qualitative insights:} The substantial gains on Puzzle and Conditioned Math (\increasenoparent{11.38} in \cref{tab:tricky_datasets_results}) and most other hard datasets can be explained from an observed phenomenon that \textbf{HoT encourages LLMs to make use of the important facts already given in the input} more (e.g., \cref{fig:teaser}, \cref{tab:gcot_vs_cot_puzzle_trivial}).
In contrast, with vanilla CoT prompting, LLMs may overlook important facts, \eg not using \sethlcolor{fact7}\hl{$b=c=d=0$} in the calculation (\cref{fig:teaser}), leading to incorrect answers. This ability to pay attention to facts is important for adversarial datasets (\eg PuzzleTrivial, AIME, MATH500 (Conditioned Math), r-GSM).

\begin{table*}[ht]
\caption{HoT outperforms CoT across 3 \textbf{ReasoningTrap} tasks, with significant gains in MATH500 (Conditioned Math) (\increasenoparent{18.00} \geminiflashlogo, an example shown in \cref{fig:teaser}) and PuzzleTrivial (\increasenoparent{12.50} \geminiflashlogo, an example shown in \cref{tab:gcot_vs_cot_puzzle_trivial}).}
\centering
\newcolumntype{g}{>{\columncolor{gray!10}}l}
\newcolumntype{G}{>{\columncolor{red!20}}l}
\small
\resizebox{0.8\columnwidth}{!}{
\begin{tabular}{lgggg|g}
\hline
\rowcolor{white} & Prompt & PuzzleTrivial & \makecell{AIME\\(Conditioned Math)} & \makecell{MATH500\\(Conditioned Math)} & Task Mean ($\Delta$) \\ \hline

\rowcolor{white}
\multirow{2}{*}{
  \raisebox{-0.3\height}{
    \hspace*{-0.45em}
    \geminiflashlogo
  }
} & CoT & 66.25 & 44.12 & 36.00 & 48.79 \\
          & \hot ($\Delta$) & \textbf{73.75} \increase{7.50} & \textbf{52.94} \increase{8.82} & \textbf{54.00} \increase{18.00} & \textbf{60.23} \increase{11.44} \\
\midrule
\rowcolor{white}
\multirow{2}{*}{
  \raisebox{-0.3\height}{
    \hspace*{-0.45em}
    \geminilogo
  }
} & CoT & 65.00 & 41.12 & 44.00 & 40.50 \\
         & \hot ($\Delta$) & \textbf{71.25} \increase{6.25} & \textbf{50.00} \increase{8.88} & \textbf{56.00} \increase{12.00} & \textbf{59.08} \increase{18.58} \\
\midrule
\rowcolor{white}
\rowcolor{white} 
\multirow{2}{*}{
  \raisebox{-0.3\height}{
    \hspace*{0.03em}
    \llamaMedlogo
  }
} & CoT & 38.75 & 20.58 & 28.00 & 29.11 \\
        & \hot ($\Delta$) & \textbf{45.00} \increase{6.25} & \textbf{26.47} \increase{5.89} & \textbf{38.00} \increase{10.00} & \textbf{36.49} \increase{7.38} \\
\midrule
\rowcolor{white}
\rowcolor{white} 
\multirow{2}{*}{
  \raisebox{-0.3\height}{
    \hspace*{0.03em}
    \llamaBiglogo
  }
} & CoT & 50.00 & 17.65 & 24.00 & 30.55 \\
         & \hot ($\Delta$) & \textbf{62.50} \increase{12.50} & \textbf{23.50} \increase{5.85} & \textbf{30.00} \increase{6.00} & \textbf{38.67} \increase{8.12} \\
\midrule
\rowcolor{white} \multicolumn{2}{l}{Model Mean $\Delta$} & \increasenoparentwithspace{8.13} & \increasenoparentwithspace{7.36} & \increasenoparentwithspace{11.50} & \increasenoparentwithspace{11.38} \\
\bottomrule
\end{tabular}
}
\label{tab:tricky_datasets_results}
\end{table*}

\begin{table*}[ht]
\centering
\caption{Over all 5 LLMs, \hot consistently improves accuracy over CoT across four \textbf{arithmetic} tasks. 
         Notably, \hot achieves the largest performance gains in AQUA (\increasenoparent{14.64} for \geminiflashlogo) and R-GSM (\increasenoparent{12.73} for \geminilogo).}
\resizebox{0.9\textwidth}{!}{%
\newcolumntype{g}{>{\columncolor{gray!10}}l}
\begin{tabular}{lggggg|g}
\hline
\rowcolor{white} Model & Method & SVAMP & AQUA & R-GSM & GSM-Symbolic & Task Mean ($\Delta$)\\
\hline
\rowcolor{white}
\multirow{2}{*}{\raisebox{-0.3\height}{\hspace*{-0.45em}\geminiflashlogo}}
 & CoT              & 92.80 & 81.42 & 85.91 & 81.67 & 85.45\\
 & \hot ($\Delta$)  & \textbf{94.80}\increase{2.00} & \textbf{96.06}\increase{14.64} & \textbf{86.36}\increase{0.45} & \textbf{83.67}\increase{2.00} & \textbf{90.22}\increase{4.77}\\
\hline
\rowcolor{white}
\multirow{2}{*}{\raisebox{-0.3\height}{\hspace*{0.03em}\geminilogo}}
 & CoT               & 94.70 & 82.68 & 78.18 & 85.67 & 85.31\\
 & \hot ($\Delta$)   & \textbf{95.80}\increase{1.10} & \textbf{91.73}\increase{9.05} & \textbf{90.91}\increase{12.73} & \textbf{86.33}\increase{0.66} & \textbf{91.19}\increase{5.89}\\
\hline
\rowcolor{white}
\multirow{2}{*}{\raisebox{-0.3\height}{\hspace*{0.03em}\llamaMedlogo}}
 & CoT              & 93.00 & 90.94 & 89.09 & 82.33 & 88.84\\
 & \hot ($\Delta$)   & \textbf{95.10}\increase{2.10} & 87.01\decrease{3.93} & 89.09 & 82.33 & 88.38\decrease{0.46}\\
\hline
\rowcolor{white}
\multirow{2}{*}{\raisebox{-0.3\height}{\hspace*{0.03em}\llamaBiglogo}}
 & CoT               & 93.70 & 87.40 & 90.91 & \textbf{90.33} & 90.59\\
 & \hot ($\Delta$)   & \textbf{95.50}\increase{1.80} & \textbf{88.98}\increase{1.58} & \textbf{91.82}\increase{0.91} & 90.00\decrease{0.33} & \textbf{91.58}\increase{0.99}\\
\hline
\rowcolor{white}
\multirow{2}{*}{\raisebox{-0.3\height}{\hspace*{0.03em}\gptlogo}}
 & CoT               & 94.60 & 79.13 & 89.09 & 87.33 & 87.54\\
 & \hot ($\Delta$)  & \textbf{95.20}\increase{0.60} & \textbf{82.68}\increase{3.55} & \textbf{90.45}\increase{1.36} & \textbf{89.33}\increase{2.00} & \textbf{89.42}\increase{1.88}\\
\hline
\rowcolor{white}\multicolumn{2}{l}{Model Mean $\Delta$}
  & \increasenoparentwithspace{1.52}
  & \increasenoparentwithspace{4.98}
  & \increasenoparentwithspace{3.09}
  & \increasenoparentwithspace{0.87}
  & \increasenoparentwithspace{2.61}\\
\hline
\end{tabular}%
}
\label{tab:full_performance_math}
\end{table*}

\begin{table*}[ht]
\centering
\caption{Over all 5 LLMs, \hot consistently improves accuracy over CoT across {three} \textbf{QA} tasks (StrategyQA, SpartQA, Date Understanding) and {two} \textbf{reading comprehension} tasks (Break \& Census). 
The largest gains are observed in StrategyQA (\increasenoparent{15.07} for \llamaMedlogo) and SpartQA (\increasenoparent{11.88} for \geminiflashlogo).
}
\resizebox{\textwidth}{!}{%
\small
\newcolumntype{g}{>{\columncolor{gray!10}}l}
\begin{tabular}{lgggggg|g}
\hline
\rowcolor{white} Model & Method   & StrategyQA & SpartQA & Date & Break & Census & Task Mean ($\Delta$)\\ 
\hline

\rowcolor{white} 
\multirow{2}{*}{
  \raisebox{-0.3\height}{
    \hspace*{-0.45em}
    \geminiflashlogo
  }
}
& CoT  & 76.55 & 47.28 & 85.24 & 83.61 & 90.00  & 76.54\\ 
& \hot ($\Delta$) & \textbf{79.74} \increase{3.19} & \textbf{59.16} \increase{11.88} & \textbf{85.79} \increase{0.55} & \textbf{86.25} \increase{2.64} & 90.00 & \textbf{80.19} \increase{3.65}\\ 
\hline

\rowcolor{white} 
\multirow{2}{*}{
  \raisebox{-0.3\height}{
    \hspace*{0.03em}
    \geminilogo
  }
}
& CoT  & 81.75 & 61.88  & 93.31  & 86.39  & 91.75 & 83.02\\ 
& \hot ($\Delta$) & \textbf{83.45} \increase{1.70} & \textbf{64.85} \increase{2.97} & \textbf{95.82} \increase{2.51} & \textbf{87.36} \increase{0.97} & \textbf{92.50} \increase{0.75} & \textbf{84.80} \increase{1.78}\\ 
\hline

\rowcolor{white} 
\multirow{2}{*}{
  \raisebox{-0.3\height}{
    \hspace*{0.03em}
    \llamaMedlogo
  }
}
& CoT  & 69.30  & 66.09 & 91.36 & 88.75 & 94.25  & 81.95\\ 
& \hot ($\Delta$) & \textbf{84.37} \increase{15.07} & \textbf{67.08} \increase{0.99} & \textbf{91.92} \increase{0.56} & \textbf{88.89} \increase{0.14} & 94.25 & \textbf{85.30} \increase{3.35}\\ 
\hline

\rowcolor{white}
\multirow{2}{*}{
  \raisebox{-0.3\height}{
    \hspace*{0.03em}
    \llamaBiglogo
  }
}
& CoT & 85.33 & 69.80 & 95.54 & 90.28 & 93.50 & 86.89\\ 
& \hot ($\Delta$) & \textbf{88.43} \increase{3.10} & \textbf{72.28} \increase{2.48} & \textbf{97.49} \increase{1.95} & 90.28 & \textbf{94.50} \increase{1.00} & \textbf{88.60} \increase{1.71}\\ 
\hline

\rowcolor{white} 
\multirow{2}{*}{
  \raisebox{-0.3\height}{
    \hspace*{0.03em}
    \gptlogo
  }
}
& CoT & 83.89 & 55.00 & 96.66 & 86.75 & 86.25 & 81.81\\ 
& \hot ($\Delta$) & \textbf{85.37} \increase{1.48} & \textbf{59.75} \increase{4.75} & \textbf{97.21} \increase{0.55} & \textbf{87.50} \increase{0.75} & \textbf{90.75} \increase{4.50}& \textbf{84.12} \increase{2.41}\\ 
\hline
\rowcolor{white} \multicolumn{2}{l}{Model Mean $\Delta$} & \increasenoparentwithspace{4.91} & \increasenoparentwithspace{4.61} & \increasenoparentwithspace{1.22} & \increasenoparentwithspace{0.90} & \increasenoparentwithspace{1.25} & \increasenoparentwithspace{2.58}
\\ \hline
\end{tabular}%
}
\label{tab:full_performance_qa_rc}
\end{table*}

\begin{table*}[ht]
\caption{\hot outperforms CoT across five \textbf{logical reasoning} tasks from BigBench Hard with notable gains in Judgement (\increasenoparent{15.5} for \gptlogo) and Five Object (\increasenoparent{6.00} for \geminilogo).}
\resizebox{\textwidth}{!}{%
\small
\newcolumntype{g}{>{\columncolor{gray!10}}l}
\newcolumntype{R}{>{\columncolor{red!20}}l}
\centering
\begin{tabular}{lgggggg|g}
\hline
\rowcolor{white} Model & Method   & Five Objects & Seven Objects & Colored Objects & \makecell{Judgement} & Navigate & Task Mean ($\Delta$)\\ 
\hline

\rowcolor{white}
\multirow{2}{*}{
  \raisebox{-0.3\height}{
    \hspace*{-0.45em}
    \geminiflashlogo
  }
}
& CoT& 78.80& 74.80  & 94.00 & 71.66 & 92.80 & 82.41\\ 
& \hot ($\Delta$) & \textbf{82.00} \increase{3.20} & \textbf{79.20} \increase{4.40} & \textbf{95.20} \increase{1.20} & 71.66  & 92.80 & \textbf{84.17} \increase{1.76}\\ 
\hline

\rowcolor{white}
\rowcolor{white} 
\multirow{2}{*}{
  \raisebox{-0.3\height}{
    \hspace*{0.03em}
    \geminilogo
  }
}
& CoT & 92.80 & 86.00 & 96.40 & 74.87 & 92.00  & 88.41\\ 
& \hot ($\Delta$) & \textbf{98.80} \increase{6.00} & \textbf{88.80} \increase{2.80} & \textbf{97.20} \increase{0.80} & \textbf{75.40} \increase{0.53} & \textbf{96.40} \increase{4.40} & \textbf{91.32} \increase{2.91}\\ 
\hline

\rowcolor{white}
\rowcolor{white} 
\multirow{2}{*}{
  \raisebox{-0.3\height}{
    \hspace*{0.03em}
    \llamaMedlogo
  }
}
& CoT & 92.80  & 79.60  & 92.00 & 67.91  & 87.60 & 83.98\\ 
& \hot ($\Delta$) & \textbf{94.00} \increase{1.20} & \textbf{83.60} \increase{4.00} & \textbf{93.60} \increase{1.60} & \textbf{71.12} \increase{3.21} & \textbf{90.40} \increase{2.80} & \textbf{86.54} \increase{2.56}\\ 
\hline

\rowcolor{white}
\rowcolor{white} 
\multirow{2}{*}{
  \raisebox{-0.3\height}{
    \hspace*{0.03em}
    \llamaBiglogo
  }
}
& CoT & 95.60 & 89.60 & 96.80  & 67.91 & 95.20 & 89.02\\ 
& \hot ($\Delta$) & \textbf{97.20} \increase{1.60} & \textbf{90.00} \increase{0.40} & \textbf{97.20} \increase{0.40} & \textbf{74.33} \increase{6.42} & \textbf{97.20} \increase{2.00} & \textbf{91.19} \increase{2.16}\\ 
\hline

\rowcolor{white}
\rowcolor{white} 
\multirow{2}{*}{
  \raisebox{-0.3\height}{
    \hspace*{0.03em}
    \gptlogo
  }
}
& CoT  & \textbf{93.60} & 85.20 & 98.40 & 73.80 & 96.40  & 89.48\\ 
& \hot ($\Delta$) & 92.80 \decrease{0.80} & \textbf{86.40} \increase{1.20} & \textbf{98.80} \increase{0.40} & \textbf{89.30} \increase{15.5} & 96.40 & \textbf{92.74} \increase{3.26}\\ 
\hline
\rowcolor{white} \multicolumn{2}{l}{Model Mean $\Delta$} & \increasenoparentwithspace{2.24}&\increasenoparentwithspace{2.56}&\increasenoparentwithspace{0.88}&\increasenoparentwithspace{5.13}&\increasenoparentwithspace{1.84} & \increasenoparentwithspace{2.53}
\\ \hline
\end{tabular}%
}
\label{tab:full_performance_logical}
\end{table*}

\begin{table*}[htbp]
\caption{HoT outperforms CoT across five hard, \textbf{long-context} tasks, with notable gains in Causal Judgement (\increasenoparent{4.50} \geminiflashlogo) and GPQA Diamond tasks (\increasenoparent{5.55} \geminiflashlogo).}
\centering
\newcolumntype{g}{>{\columncolor{gray!10}}l}
\newcolumntype{G}{>{\columncolor{red!20}}l}
\small
\resizebox{\columnwidth}{!}{
\begin{tabular}{lgggggg|g}
\hline
\rowcolor{white} & Prompt & \makecell{BBEH\\Time Arithmetic} & \makecell{BBEH\\Shuffle Objects} & \makecell{BBEH\\Spatial Reasoning} & \makecell{BBEH\\Causal Judgement} & \makecell{GPQA\\ Diamond} & Task Mean ($\Delta$) \\ \hline

\rowcolor{white}
\multirow{2}{*}{
  \raisebox{-0.3\height}{
    \hspace*{-0.45em}
    \geminiflashlogo
  }
} & CoT & 31.00 & 3.00 & 9.50 & 48.50 & 32.83 & 24.97 \\
          & \hot ($\Delta$) & \textbf{32.00} \increase{1.00} & \textbf{7.00} \increase{4.00} & \textbf{10.00} \increase{0.50} & \textbf{53.00} \increase{4.50} & \textbf{38.38} \increase{5.55} & \textbf{28.08} \increase{3.11} \\
\midrule
\rowcolor{white}
\multirow{2}{*}{
  \raisebox{-0.3\height}{
    \hspace*{-0.45em}
    \geminilogo
  }
} & CoT & 46.00 & 17.00 & 10.00 & 53.50 & 41.92 & 33.68 \\
         & \hot ($\Delta$) & \textbf{48.50} \increase{2.50} & \textbf{19.00} \increase{2.00} & \textbf{10.50} \increase{0.50} & \textbf{58.00} \increase{4.50} & \textbf{43.43} \increase{1.51} & \textbf{35.89} \increase{2.21} \\
\midrule
\rowcolor{white}
\rowcolor{white} 
\multirow{2}{*}{
  \raisebox{-0.3\height}{
    \hspace*{0.03em}
    \llamaMedlogo
  }
} & CoT & 26.50 & 14.00 & 10.00 & 47.00 & 21.21 & 23.74 \\
        & \hot ($\Delta$) & \textbf{29.00} \increase{2.50} & \textbf{15.00} \increase{1.00} & 10.00 & \textbf{50.00} \increase{3.00} & \textbf{25.25} \increase{4.04} & \textbf{25.85} \increase{2.11} \\
\midrule
\rowcolor{white}
\rowcolor{white} 
\multirow{2}{*}{
  \raisebox{-0.3\height}{
    \hspace*{0.03em}
    \llamaBiglogo
  }
} & CoT & 41.00 & 17.00 & 10.00 & 56.00 & 23.23 & 29.45 \\
         & \hot ($\Delta$) & \textbf{41.50} \increase{0.50} & 17.00 & \textbf{11.00} \increase{1.00} & 56.00 & \textbf{25.25} \increase{2.02} & \textbf{30.15} \increase{0.70} \\
\midrule
\rowcolor{white} \multicolumn{2}{l}{Model Mean $\Delta$} & \increasenoparentwithspace{1.63} & \increasenoparentwithspace{1.75} & \increasenoparentwithspace{0.5} & \increasenoparentwithspace{3.00} & \increasenoparentwithspace{3.28} & \increasenoparentwithspace{2.03} \\
\bottomrule
\end{tabular}
}
\label{tab:long_input_results}
\end{table*}

\subsection{Span-Level LLM Judging Reveals Lower Hallucination Rates with HoT}
\label{sec:hallucination_span_detection}

SelfCheckGPT \citep{manakul-etal-2023-selfcheckgpt} primarily measures response consistency (see \cref{sec:hallucination}) and does not directly localize which parts of an output are hallucinated. To obtain span-level hallucination annotations, we instead use an LLM-as-a-judge setup (see \cref{sec:hallucination_examples}), which can flag specific hallucinated spans.

Aligning with \citep{cossio2025comprehensive,li2024fine}, we use four broad hallucination types: contradiction, missing context, calculation error, and logical error.

\subsec{1. Contradiction}: Answer directly contradicts a specific fact in the question.\\
\subsec{2. Missing Context}: Answer ignores crucial information from the question that affects the reasoning.\\
\subsec{3. Calculation Error}: The arithmetic computation itself is mathematically incorrect.\\
\subsec{4. Logical Error}: The reasoning step is logically flawed or misinterprets the problem.

\subsec{Experiment} We apply LLM-as-Judge using Gemini-2.5-Flash to evaluate \geminiflash HoT responses across five diverse benchmarks: r-GSM, GSM-Symbolic, SpartQA, Break, and Seven Objects. Hallucination rate is computed as the proportion of questions containing at least one hallucinated span, and the same evaluation is repeated for CoT responses. We report both hallucination rate and task accuracy to assess whether HoT reduces hallucinations while preserving or improving performance compared to CoT (see \cref{tab:judge_hallucination_reduction,tab:judge_accuracy_improvement}).

\begin{table}[ht]
\centering
\begin{minipage}{0.48\textwidth}
\centering
\caption{HoT reduces the hallucination rate measured as the fraction of responses containing hallucinations.}
\label{tab:judge_hallucination_reduction}
\resizebox{\textwidth}{!}{%
\begin{tabular}{lccccc}
\hline
Hallucination Rate $\downarrow$ & R-GSM & GSM-Symbolic & SpartQA & Break & Seven Objects \\
\hline
CoT & 12.00 & 14.50 & 80.00 & 14.50 & 21.50 \\
HoT & 6.00 & 14.50 & 74.00 & 14.00 & 12.50 \\
$\Delta(\text{CoT} - \text{HoT})$ & 6.00 & 0.00 & 6.00 & 0.50 & 9.00 \\
\hline
\end{tabular}
}
\end{minipage}
\hfill
\begin{minipage}{0.48\textwidth}
\centering
\caption{Accuracy comparison between CoT and HoT across different benchmarks.}
\label{tab:judge_accuracy_improvement}
\resizebox{\textwidth}{!}{%
\begin{tabular}{lccccc}
\hline
Accuracy $\uparrow$ & R-GSM & GSM-Symbolic & SpartQA & Break & Seven Objects \\
\hline
CoT & 78.18 & 85.67 & 61.88 & 86.39 & 86.00 \\
HoT & 90.91 & 86.33 & 64.85 & 87.36 & 88.80 \\
$\Delta(\text{HoT} - \text{CoT})$ & 12.73 & 0.66 & 2.97 & 0.97 & 2.80 \\
\hline
\end{tabular}
}
\end{minipage}
\end{table}

\subsec{Results} We observe that while HoT reduces the hallucination rate in Seven Objects by \textbf{9.00} points but only improves the accuracy by \textbf{2.80} points. One possible explanation is that the Seven Objects dataset requires complex multi-step reasoning to reach the final answer. Hallucinations often appear in intermediate reasoning steps, yet the model can still produce the correct final answer. Consequently, fixing these hallucinations yields only minimal accuracy improvements.

For reading comprehension datasets (Break, SpartQA), hallucinations primarily arise from misusing or ignoring passage information, so we evaluate them using Contradiction and Missing Context. In contrast, hallucinations in arithmetic and logical reasoning datasets (r-GSM, GSM-Symbolic, Seven Objects) mainly result from faulty computation or invalid inference, and are therefore labeled as Calculation Error and Logical Error. This dataset-specific labeling aligns evaluation with each dataset’s dominant failure modes.

\subsection{\hot outperforms CoT and other advanced prompting methods}
\label{sec:hot_beat_other_methods}

Given that HoT provides consistent gains over CoT prompting, sometimes substantial on hard and adversarial benchmarks (\cref{sec:main_results}), it is natural to ask: Q1: How would HoT fare against other advanced prompting techniques?
Q2: Can HoT be used in tandem with other techniques to even further improve LLM accuracy?


We first examine whether HoT outperforms the Repeated Question prompting (RQ) \citep{mekala-etal-2024-echoprompt}, then extend our comparison by integrating both HoT and CoT with Self-Consistency (SC) and the ComplexCoT strategy. 
That is, SC takes multiple LLM responses to the same prompt and selects the most frequent answer among these outputs, whereas ComplexCoT selects the most complex answer.

Additionally, we evaluate HoT against multi-step promptings including Least-to-Most (LtM), Tree-of-Thought (ToT), Self-Refine, and Chain-of-Verification (CoVE) (see details at \cref{sec:prompting_details}).

\subsec{Experiment} We evaluate HoT, CoT, RQ, CoT + SC and HoT + SC, ComplexCoT, and ComplexHoT (HoT + ComplexCoT) across five independent runs on 17 benchmarks (7 Arithmetic, 5 Logical Reasoning, 3 Question Answering, and 2 Reading Comprehension, details of each dataset are in \cref{sec:datasets_app}) and two models (\geminiflashlogo, \geminilogo) (\cref{tab:full_performance_math,tab:full_performance_qa_rc,tab:full_performance_logical}) and report the mean and standard deviation (std) to ensure reliability. 
Mean accuracies across all benchmarks are in \cref{tab:advanced_prompting_table}.

We also compare HoT against four other state-of-the-art prompting techniques on 3 datasets (r-GSM, Seven Objects, and Date) across 5 runs and two models (\geminiflashlogo, \geminilogo). 
The reported mean accuracies are in \cref{tab:advanced_prompting_table_2}.

\label{sec:advanced_prompting}

\textbf{Result} 
Over 17 datasets, HoT outperforms both RQ and CoT (\cref{tab:advanced_prompting_table}).
Interestingly, HoT even surpasses CoT + SC by \increasenoparent{1.02}  (87.24 vs. 86.22; \cref{tab:advanced_prompting_table}).

When combining HoT with SC, the resultant HoT + SC method surpasses HoT alone and even CoT+SC.
Similarly, ComplexHoT outperforms HoT alone and ComplexCoT alone consistently (\cref{tab:advanced_prompting_table}).
That is, HoT provides a \textbf{distinct benefit} to LLMs as their accuracy continues to improve as HoT prompting is used in combination with other approaches that reward consistency (SC) and longer thinking (ComplexCoT).

\begin{table}[H]
\caption{Over 5 runs across 17 benchmarks, HoT consistently outperforms both CoT and Repeating Questions (RQ), and even CoT + Self-Consistency (SC), and ComplexCoT.
ComplexHoT and HoT + SC also outperforms their counterparts (ComplexCoT and CoT + SC) showing that HoT can complement these methods.
}
\centering
\resizebox{0.8\textwidth}{!}{
\begin{tabular}{lccc|cc|cc}
\hline
Model & CoT & RQ & HoT & CoT + SC & HoT + SC & ComplexCoT & ComplexHoT \\ \hline 
\raisebox{-0.2\height}{
    \hspace*{-0.45em}
    \geminiflashlogo
  } & 83.21$\pm$0.82 & 83.51$\pm$1.98 & \textbf{85.17$\pm$1.85} & 83.68 & \textbf{87.18} & 83.79 & \textbf{86.37} \\
\raisebox{-0.2\height}{
    \hspace*{0.03em}
    \geminilogo
  } & 87.61$\pm$1.09 & 88.35$\pm$1.52 & \textbf{89.31$\pm$0.92} & 88.76 & \textbf{90.39} & 89.11 & \textbf{90.71} \\\hline  
  \textbf{Mean $\pm$ std} & 85.41$\pm$0.96 & 85.93$\pm$1.75 & \textbf{87.24$\pm$1.38} & 86.22 & \textbf{88.79} & 86.45 & \textbf{88.54} \\ \hline
\end{tabular}
}
\label{tab:advanced_prompting_table}
\end{table}

On average over 5 runs and 3 datasets, HoT alone is still the most performing method compared to all other advanced prompting methods of CoT, LtM, CoVE, Self-Refine, and ToT (\cref{tab:advanced_prompting_table_2}). Under other prompting techniques, we observe LLMs often miss critical facts (\eg, overlooking temporal indicators like ``yesterday'' in Date), causing incorrect answers.
In contrast, LLMs tend to focus better on key facts under HoT prompting.

\begin{table}[ht]
\caption{Mean ± standard deviation over 5 runs across r-GSM, Seven Objects, and Date comparing HoT against LtM, ToT, Self-Refine, and CoVE. HoT consistently outperforms these advanced methods.
}
\centering
\resizebox{0.75\textwidth}{!}{
\begin{tabular}{lccccc|c}
\hline
Model & CoT & LtM & CoVE & Self-Refine & ToT & HoT \\
\hline
\raisebox{-0.3\height}{
    \hspace*{-0.45em}
    \geminiflashlogo
  } & 81.41$\pm$0.90 & 75.94$\pm$1.42 & 74.07$\pm$1.34 & 70.32$\pm$0.72 & 81.98$\pm$1.70 & \textbf{84.21$\pm$1.45} \\
\raisebox{-0.3\height}{
    \hspace*{0.03em}
    \geminilogo
  } & 88.94$\pm$2.07 & 84.38$\pm$0.62 & 81.54$\pm$1.27 & 81.74$\pm$1.66 & 87.39$\pm$0.92 & \textbf{91.87$\pm$0.91} \\
\hline
\textbf{Mean} & 85.18$\pm$3.77 & 80.16$\pm$4.22 & 77.81$\pm$3.73 & 76.03$\pm$5.71 & 84.69$\pm$2.71 & \textbf{88.04$\pm$3.83} \\\hline
\end{tabular}
}
\label{tab:advanced_prompting_table_2}
\end{table}

\subsection{Ablation study: Repeating the question and adding \texttt{<fact>} tags improves accuracy}
\label{sec:prompt_variation_exp}

As \hot consistently outperforms CoT over many tasks (\cref{sec:main_results,sec:hot_beat_other_methods}), a burning question is: What components contribute to \hot's success?

\hot consists of two steps: (a) regenerating the question but with XML tags around key facts; and (b) adding tags to the answer.
Interestingly, concurrent work showed that repeating the question alone can help improve LLM accuracy \citep{mekala-etal-2024-echoprompt}.
Therefore, we perform an ablation study to understand the impact of each \hot component: (1) Repeating the question alone (no tags); (2) Repeating and adding tags to the question; and (3) Adding tags to the answer, but not to the question.

\subsec{Experiment} We compare the following variations: 
(a) \textbf{\cotcolored}: Few-shot CoT baseline \citep{wei2022chain}; 
(b) \textcolor{RQColor}{\textbf{Repeated Question (R-Q)}}: Repeat the input question and then generate the regular CoT answer \citep{mekala-etal-2024-echoprompt}.\\
(c) \textcolor{TQColor}{\textbf{Tags in Question (T-Q)}}: Repeat the question and wrap key facts in it with XML tags while leaving the answer untagged.\\
(d) \textcolor{TAColor}{\textbf{Tags in Answer (T-A)}}: Repeat the question and wrap key facts in the answer with XML tags.\\
(e) \textcolor{HoTColor}{\textbf{Tags in Question and Answer (HoT)}}: The full \hot recipe, \ie wrapping XML tags around key facts in the reformatted question and also adding corresponding tags to the answer.

For each variation, we evaluate 4 models (\geminiflashlogo, \geminilogo, \llamaMedlogo, \llamaBiglogo) on 400 randomly sampled instances across 6 diverse, representative datasets (AQUA, StrategyQA, R-GSM, Seven Objects, Judgement, \& Navigate). 

\begin{figure}[ht]
\begin{minipage}{0.55\textwidth}
    \includegraphics[width=0.95\textwidth]{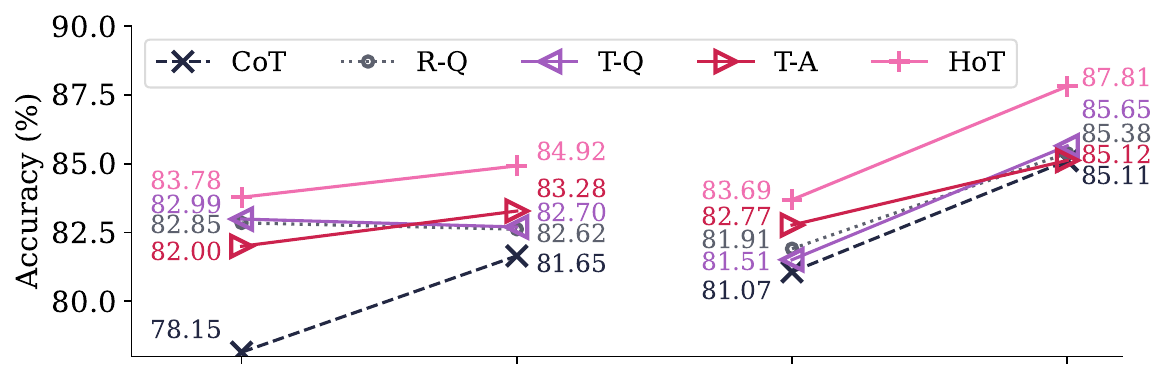}
    {\scriptsize
    \makebox[0.4\textwidth][c]{\raisebox{-0.3\height}{\geminiflashlogo}}%
    \makebox[0.06\textwidth][c]{\raisebox{-0.3\height}{\geminilogo}}%
    \makebox[0.4\textwidth][c]{\raisebox{-0.3\height}{\llamaMedlogo}}%
    \makebox[0.04\textwidth][c]{\raisebox{-0.3\height}{\llamaBiglogo}}%
    }
\end{minipage}%
\hfill
\begin{minipage}{0.45\textwidth}
    \caption{
    HoT ablation study: Every component---repeating the question (\rqcolored), adding tags to only question (\tqcolored), adding tags to only answer (\tacolored)---independently contributes to the overall accuracy of \hotcolored prompting (\textbf{\textcolor{HoTColor}{$+$}}).
    Each component also outperforms the vanilla \cotcolored (-$\times$-).
    $y$-axis shows mean accuracy across 6 datasets (the detailed accuracy of each dataset is in \cref{sec:detail_prompt_variation_app}).
    }
    \label{fig:comparison}
\end{minipage}
\end{figure}

\subsec{Results} 
First, on average across 6 datasets and 4 LLMs, every variation (\rqcolored, \tqcolored, \tacolored, and \hotcolored) outperforms the baseline \cotcolored (\cref{fig:comparison}).
That is, each HoT component is an improvement over CoT.
Second, for some smaller LLMs (\geminiflashlogo, \llamaMedlogo), adding each component monotonically increases accuracy while the trend is weaker for larger models (\geminilogo).

Interestingly, examining the results of the \tacolored method shows that, for larger LLMs (\ie \geminilogo and \llamaBiglogo), while we instruct them via demonstrations to insert tags exclusively to the answer (but not the question), these two models still generate, on average, 3.27 and 2.27 tags, respectively, in each question (\cref{tab:avg_tags}).
As the result, instructing LLMs to generate \textbf{tags exclusively in the answer (\tacolored) is sufficient for large LLMs to gain accuracy over not having tags at all (\cotcolored, and \rqcolored)}.

We also find that intentionally scrambling tags in the QA pairs in the few-shot examples significantly reduces HoT’s accuracy across datasets (\decreasenoparent{2.13} pp \cref{sec:misaligned_tag}). However, even with mismatched tags, HoT still outperforms CoT by \increasenoparent{1.21} pp (\cref{sec:misaligned_tag}) perhaps because LLMs do not highlight tags randomly as shown in the demonstrations.

\subsection{\hot{} highlights help \human{} humans improve their speed of verifying LLM answers}
\label{sec:human_verification}

High-quality text highlights in classification tasks reduce decision time and effort, while low-quality highlights harm user accuracy \citep{ramirez2019understanding}.
Here, we also aim to assess how our highlights in HoT impact user decision verification accuracy and time.
That is, users are asked to decide whether the final answer is accurate given the input question, and the HoT answers.
To our knowledge, this would be the first study in the LLM literature that studies how highlights in chains of thought impact users on a downstream task.


\subsec{Experiment}
To evaluate the impact of highlights to humans on arithmetic and reading comprehension problems, we measure user accuracy in verifying whether LLM responses (of \llamaSmalllogo, \llamaMedlogo, \llamaBiglogo, \geminilogo and \geminiflashlogo) are accurate for GSM-Symbolic and DROP questions. 
This verification task is known as \textbf{distinction} \citep{kim2022hive} or \textbf{verification} task \citep{visual_correspondence,nguyen2021effectiveness}.

We select 30 incorrect and 30 correct HoT responses from each dataset, forming a balanced pool of questions with 120 cases of \hot. We then remove the XML \texttt{<fact>} tags from HoT responses, resulting in 120 HoT and 120 CoT responses for user verification.

We recruit 63 users, consisting of undergraduate and graduate students, each verifying 10 LLM responses to questions from GSM-Symbolic and DROP via an online interface (see \cref{sec:human_study}). Users are randomly assigned to see exclusively \hot or CoT responses and then have to predict if a response is \texttt{correct} or \texttt{incorrect}. 
To simulate real-world constraints, users are given a maximum of 2 minutes per question, after which, they are required to make a decision.

         
         
         

\begin{table}[h!]
    \centering
    \small
    \begin{minipage}{0.45\textwidth}
        \centering
        \resizebox{0.95\textwidth}{!}{
        \begin{tabular}{llcc}
        \hline
             Method & \makecell{Avg Time\\ (secs)} & \makecell{Accuracy (\%)\\ LLM is \correct{correct}}  & \makecell{Accuracy (\%)\\ LLM is \incorrect{incorrect}} \\\hline
             
             \hot & \textbf{47.26} & \textbf{84.48} ± 20.28  & 54.83 ± 30.13 \\
             
             CoT & 62.38 & 78.82 ± 28.26   & \textbf{72.21} ± 21.99       \\ \hline
             
        \end{tabular}
        }
    \end{minipage}%
    \hfill
    \begin{minipage}{0.55\textwidth}
        \caption{\human Users spend $\sim$25\% less time when verifying HoT answers (compared to no highlights, \ie CoT).
        Highlights make users accept LLM answers more, yielding higher accuracy in \texttt{correct} cases but worse in \texttt{incorrect} ones.}
        \label{tab:human_results}
    \end{minipage}
\end{table}

\subsec{Results}
On average, over all 240 cases including both \correct{correct} and \incorrect{incorrect} answers, \textbf{users spend nearly 25\% less time} (47.26 vs. 62.38 seconds, \ie \increasenoparent{15.12} seconds faster) when making decisions with highlights (\cref{tab:human_results}).
Interestingly, users perform better on \texttt{correct} cases (84.48 vs. 78.82) but worse on \texttt{incorrect} cases (54.83 vs. 72.21).
That is, \textbf{HoT highlights tend to make users believe that an answer is accurate} (\cref{sec:user_verification_visualization}), making them more likely to accept the answer.
This aligns with recent findings \citep{jazwinska2025ai,wu2025automated} showing that search engines and LLMs often generate misleading citations, causing users to wrongly trust their answers.

\subsec{In practice:} LLMs are often more correct than wrong. 
Therefore, the ratio of \texttt{correct}/\texttt{incorrect} cases is far above 50/50.
That is, we report user accuracy on these two subsets separately (\cref{tab:human_results}) as taking an average might lead to \emph{misleading} interpretation.
For real-world datasets, \textbf{highlights in \hot is still estimated to improve the user verification accuracy} over an entire dataset, \ie, both \texttt{correct} and \texttt{incorrect} cases (\cref{sec:human_estimated_accuracy}).


\subsection{Learning to highlights make LLMs improve OOD accuracy and pay more attention to key facts}
\label{sec:attention_entropy}

Our previous sections showed that HoT prompting improves the accuracy consistently for large LLMs including Gemini and Llama families.
However, prior research also found that smaller LLMs tend to struggle in following instructions \citep{fewshot}.
Similarly, our preliminary results reveal that HoT few-shot prompting does not help small LLMs of a few billion parameters---they fail to highlight facts appropriately.

To smaller LLMs generate highlighted chains of thoughts, we first use HoT few-shot prompting to generate SFT training examples and and repeat the same procedure to generate CoT examples separately.
Then, we finetune small open-weight LLMs on these two datasets and compare HoT-finetuned and CoT-finetuned LLMs against the baselines.
One might also ask: \textbf{Would fine-tuning improve LLM performance on out-of-distribution (OOD) datasets?} 
Our answer: Yes! We observe that HoT-finetuned models can recall input facts more accurately, resulting in more accurate answers (\cref{tab:puzzletrivial_acc}).

To shed light on why accuracy improves after finetuning on HoT examples, we perform an attention entropy analysis \citep{zhang2024attention} and find that HoT-finetuned models pay more attention to tagged facts than the baseline and CoT-finetuned models.

\subsec{Experiment} We finetune LLaMA-3.2-1B \citep{meta2024llama} and Qwen2.5-1.5B \citep{yang2024qwen2} on 3,187 CoT responses, and 3,187 HoT responses in total that \geminipro~\geminilogo generates for 17 datasets (\ie, those datasets previously shown in \cref{tab:full_performance_math,tab:full_performance_qa_rc,tab:full_performance_logical}).
Then, we compare the accuracy of 3 models (original model, model finetuned on CoT, and finetuned on HoT) across 5 OOD benchmarks (\cref{tab:puzzletrivial_acc}). 
We also measure the attention entropy of three models on PuzzleTrivial questions (\cref{tab:attention_entropy}) to gain insights into why HoT-training may improve LLM accuracy.

\begin{table*}[htbp]
\caption{HoT-finetuned LLMs consistently outperform both the LLMs finetuned on CoT and the base model across 5 OOD tasks including: 3 ReasoningTrap (PuzzleTrivial, AIME, and MATH500), and 2 long-context, logical tasks (BBEH Causal Judgement, BBEH Time Arithmetic), an example shown in \cref{fig:cot_vs_hot_puzzletrivial}).}
\centering
\newcolumntype{g}{>{\columncolor{gray!10}}l}
\newcolumntype{G}{>{\columncolor{red!20}}l}
\small
\resizebox{\columnwidth}{!}{
\begin{tabular}{lrrrrr|r}
\hline
\rowcolor{white}Model & PuzzleTrivial & \makecell{AIME\\(Conditioned Math)} & \makecell{MATH500\\(Conditioned Math)} & \makecell{BBEH\\ Causal Judgement} & \makecell{BBEH\\Time Arithmetic} & Mean ($\Delta$) \\ \hline

\rowcolor{white}Qwen-2.5-1.5B
 & 13.75 & 14.70 & 20.00 & 39.50 & 18.50 & 21.30  \\
 \rowcolor{white}+ SFT on CoT & 28.75 & 11.76 & 18.00 & 15.00 & 10.00 & 16.70 \\
 + SFT on HoT & \textbf{38.75} & \textbf{29.40} & \textbf{28.00} & \textbf{40.00} & \textbf{28.50} & \textbf{32.94} \\
\midrule
\rowcolor{white}Llama-3.2-1B
 & 11.20  & 2.90 & 10.00 & 21.00 & 30.50 & 15.12  \\
 \rowcolor{white}+ SFT on CoT & 13.75 & 5.90 & 2.00 & 29.50 & 45.00 & 19.24\\
 + SFT on HoT & \textbf{18.75} & \textbf{11.80}  & 10.00 & \textbf{32.00} & \textbf{52.50} & \textbf{27.02} \\
\bottomrule
\end{tabular}
}
\label{tab:puzzletrivial_acc}
\end{table*}

\begin{figure*}[h!]
    \centering
    \includegraphics[width=\columnwidth]{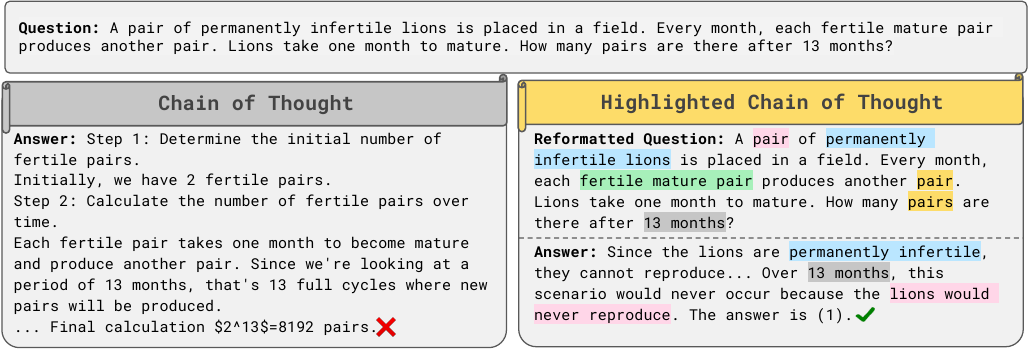}
    \caption{
    \textbf{Left:} After finetuned via SFT on CoT examples, Qwen-2.5-1.5B answers incorrectly an adversarial question from PuzzleTrivial as it does \emph{not} factor in the key fact of \sethlcolor{fact1}\hl{\emph{``permanently infertile lions''}}.
    \textbf{Right:}
    In contrast, HoT-finetuned counterpart LLM can highlight facts and answer correctly using the fact (\sethlcolor{fact4}\hl{\emph{``the lions would never reproduce''}}).
    }
\label{fig:cot_vs_hot_puzzletrivial}
\end{figure*}

\subsec{Results}
Llama-3.2-1B finetuned on HoT examples achieves the best accuracy compared to the CoT-finetuned model and the base model (27.02 vs 19.24 and 15.12). 
Similarly, Qwen-2.5-1.5B finetuned on HoT examples achieves the best accuracy compared to the CoT fine-tuned model and base model (32.94 vs 21.30 and 16.70) (see \cref{tab:puzzletrivial_acc}). 
After finetuning on HoT, LLMs does not need few-shot prompting anymore.
Instead, they generate answers directly with highlights over facts (see \cref{fig:cot_vs_hot_puzzletrivial}).

Interestingly, we find LLMs finetuned on HoT training examples demonstrate a more focus attention maps, \ie, consistently lower attention entropy (e.g., 1.524 for HoT-finetuned Llama-3.2-1B vs. 1.671 for the base Llama-3.2-1B; \cref{tab:attention_entropy}).
Qualitatively, over hard, adversarial datasets such as PuzzleTrivial, HoT-finetuned LLMs accurately recall the key facts that may be uncommon (\eg, \sethlcolor{fact1}\hl{\emph{``permanently infertile lions''}}; \cref{fig:cot_vs_hot_puzzletrivial}) to answer tricky questions correctly.
In contrast, the base LLMs or CoT-finetuned models often overlook such facts (\cref{fig:cot_vs_hot_puzzletrivial}; Left), yielding wrong answers.


\begin{table}[h]
\centering
\begin{minipage}{0.3\textwidth}
\centering
\small
\begin{tabular}{lcc}
\hline
\makecell{Attention\\Entropy} & Qwen2.5-1.5B & Llama-3.2-1B \\
\hline
Base & 2.150 & 1.671\\
+ SFT on CoT & 2.141 & 1.847 \\
+ SFT on HoT & \textbf{1.773} & \textbf{1.524} \\
\hline
\end{tabular}
\end{minipage}%
\hfill
\begin{minipage}{0.54\textwidth}
\caption{LLMs finetuned on HoT show lower attention entropy on PuzzleTrivial questions than the CoT-finetuned and base models.
Attention entropy is computed as 
$H(A) = -\sum_i p_i \log(p_i)$, 
where $A$ denotes the attention distribution over tokens 
and $p_i$ is the normalized attention weight assigned to token $i$.}
\label{tab:attention_entropy}
\end{minipage}
\end{table}

\section{Limitations} 
\textbf{First}, while larger models such as \llamaBiglogo, and \gptlogo are generally able to follow the format reliably, smaller models like \llamaSmalllogo and \geminiflashlogo can exhibit inconsistencies, sometimes repeating the few-shot examples (\cref{sec:small_models_app}). 
Smaller LLMs might incorrectly repeat the question (in \cref{sec:grounding_examples_app}) or fail to adhere to the tag format, as demonstrated in the few-shot examples (\cref{sec:small_models_app}).
\textbf{Second}, HoT prompting inherently requires LLMs to repeat the question, which costs extra tokens and could be problematic for long-context questions (see \cref{sec:token_overhead}).
\textbf{Third}, HoT prompting requires task-specific demonstrations, for which we manually create 8 examples per dataset.
However, highlighting is a costly human effort and may be non-trivial to even define for some domains.




\section{Discussions and Conclusion}

We further fine that HoT few-shot prompting causes both \llamaB and \llamaC to assign more attention mass to the XML tags tokens (\eg, \texttt{<fact1>}) than to other tokens outside the tags (\cref{sec:attention_allocation}).
Internally, such attention pattern may improve LLM focus on key facts, thus improving accuracy and reducing hallucination.
Additionally, finetuning thinking LLMs DeepSeek-R1 to highlight facts and intermediate results in their \emph{long reasoning chains} (see \cref{sec:computed_result_tagging,sec:hot_o1}) might improve their accuracy and help users interpret their complex reasoning chains faster.

In this paper, we find the highlights to reduce the verification speed of users substantially; however, do highlights improve the verification accuracy of another LLM judge?
Our preliminary results do not find clear evidence supporting this hypothesis (\cref{sec:llm_judge}), which requires further research.

We present \hot, a novel prompting approach that enables LLMs to directly reference text from the input question in their responses. 
\hot improves LLM accuracy on arithmetic, question answering, and logical reasoning tasks.
Furthermore, highlights in HoT answers also improve user verification speed.

\section{Broader Impact Statement}
\label{sec:broader_impact_statement}
HoT highlights supporting information alongside model answers, but it can also introduce risks. When a model presents highlighted evidence, the output may appear more trustworthy—even if the reasoning is flawed or the conclusion is incorrect. Our human study \cref{sec:human_verification}) shows that highlighting can increase users’ confidence in model responses, including incorrect ones. In such cases, the structured format may create a false sense of reliability. This is especially concerning in high-stakes areas such as medicine, law, or financial decision-making, where small errors can have serious consequences.

We emphasize that highlighting does not guarantee correctness. Users should treat HoT as a tool for inspection, not as proof that the answer is accurate. Future work should explore additional safeguards, such as independent verification, uncertainty estimation, and human review, to reduce the risk of over-trust. 

\section*{Acknowledgments}
We are thankful for the API research credits from \texttt{together.ai} to MRT. AN was supported by the NSF Grant No. 1850117 \& 2145767, and donations from NaphCare Foundation \& Adobe Research.

\clearpage

\renewcommand{\thesection}{\arabic{section}}
\renewcommand{\thesubsection}{\thesection.\arabic{subsection}}
\newcommand{\beginsupplementary}{%
	\setcounter{table}{0}
	\renewcommand{\thetable}{T\arabic{table}}%
	\setcounter{figure}{0}
	\renewcommand{\thefigure}{F\arabic{figure}}%
	\setcounter{section}{0}
}

\bibliography{main}
\bibliographystyle{tmlr}

\clearpage
\appendix
\newcommand{\suptitle}{Appendix for:\\\cleanpapertitle}

\newcommand{\maketitlesupp}{
    \newpage
    \onecolumn
        \null
        \begin{center}
        \parbox{0.75\textwidth}{\centering {\Large \bf \suptitle}}

            \bottomtitlebar
        \end{center}
}

\maketitlesupp

\section{\texorpdfstring{\human}  HHuman Study}\label{sec:human_study}
\subsection{User Instructions}
At the beginning of the quiz, the users were given the following instructions:
1. Judge whether each answer is \textbf{correct} or \textbf{incorrect} using only the provided reasoning.

2. Do not use external resources or scratch paper.

3. You will evaluate 10 randomly selected questions under a 20-minute time limit.

4. Complete the quiz on a computer (NO phones).

Following these instructions, participants were shown an example CoT or \hot response (correct or incorrect), depending on their assigned group. They then completed two trivial practice questions, presented in the same CoT or \hot format. Finally, participants were informed that the actual quiz was about to begin, with a maximum time limit of two minutes per question.

\subsection{Data Analysis}
During analysis, we observed that some participants answered most questions legitimately but selected answers randomly near the end to finish quickly. To mitigate the impact of such behavior while preserving valid earlier responses, we filtered out answers with response times under 5 seconds.

If a participant exceeded the 2-minute time limit for a question, the response was recorded as selecting the option indicating incorrect reasoning. We adopted this rule because failure to verify an answer within the allotted time suggests low confidence in its validity, making it reasonable to label the answer as incorrect.

\subsection{Interface Examples}
\label{sec:human_study_interface}
\begin{figure}[ht]
\centering
\begin{subfigure}
  \centering
  \includegraphics[width=0.48\textwidth]{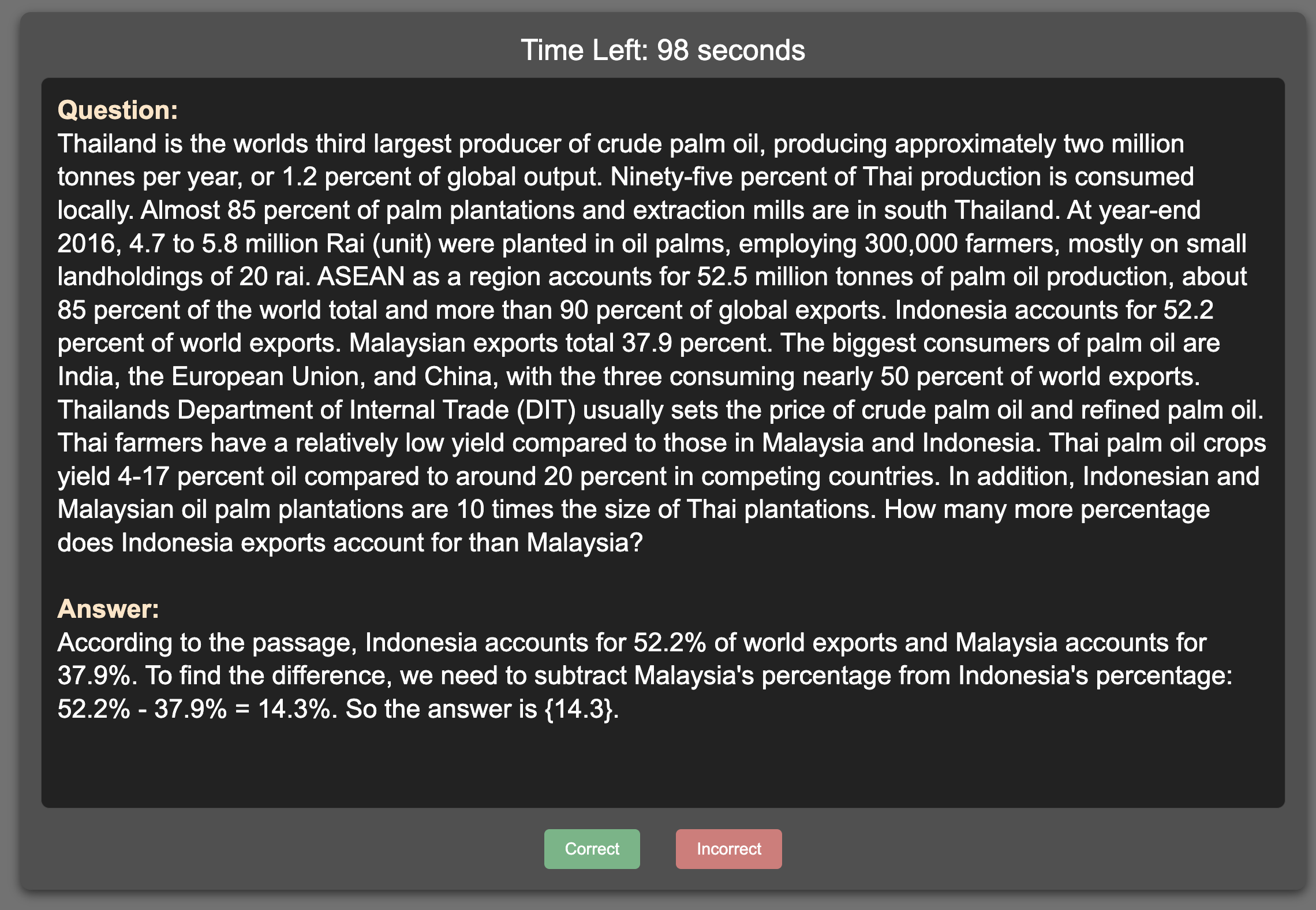}
  \label{fig:untagged_web}
\end{subfigure}\hfill
\begin{subfigure}
  \centering
  \includegraphics[width=0.48\textwidth]{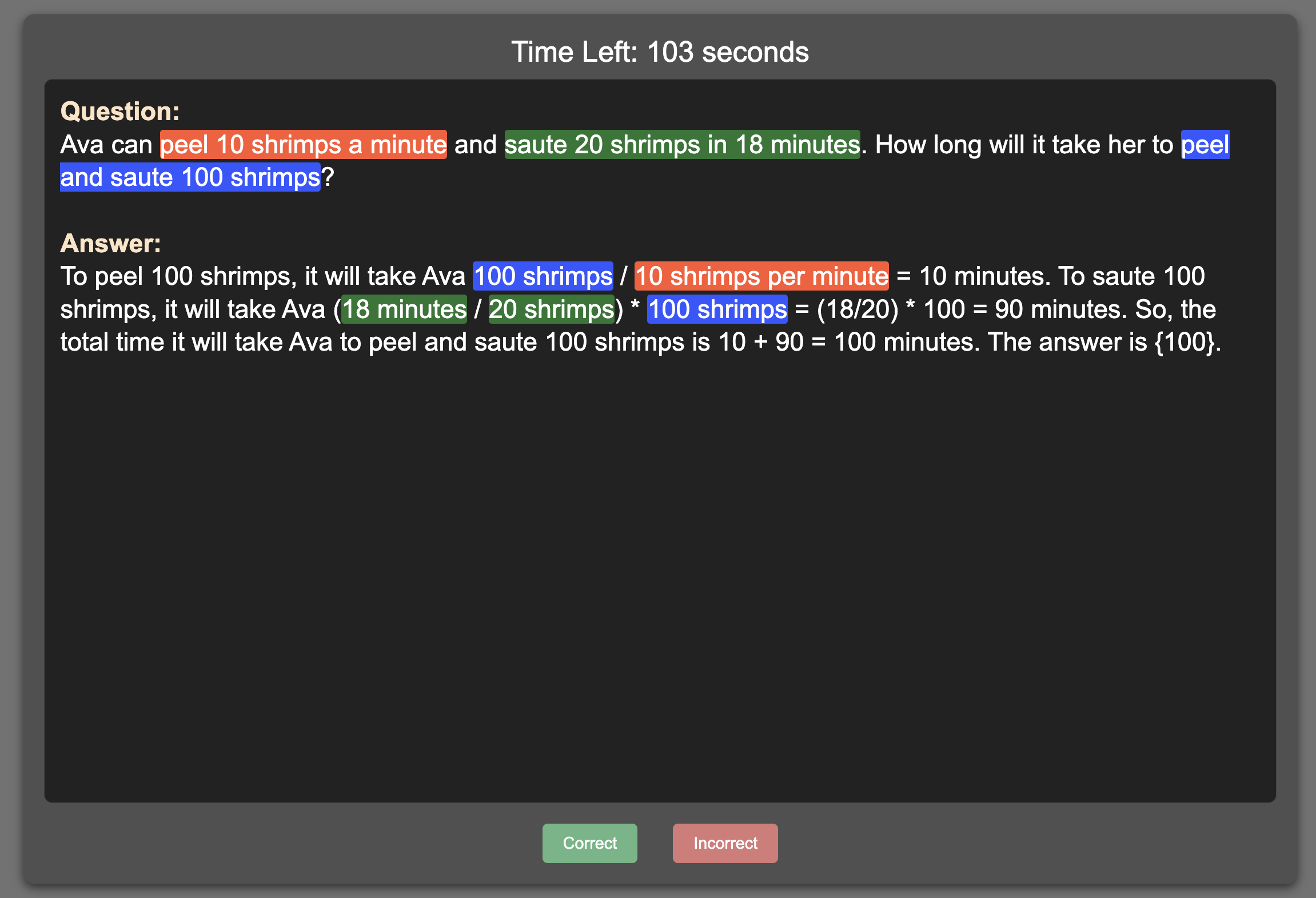}
  \label{fig:tagged_web}
\end{subfigure}
\caption{Left: Screenshot of a CoT question from the DROP dataset in the online interface. Right: Screenshot of a \hot\ question from the GSM-Symbolic dataset in the online interface.}
\label{fig:comparison}
\end{figure}



\clearpage
\section{Examples Where Users Misjudged the HoT Responses}
\label{sec:user_verification_visualization}
To better understand how HoT can sometimes lead users to accept incorrect answers, we analyze representative cases where users misjudged the model’s HoT output (\cref{tab:incorrectly_accept_1,tab:incorrectly_accept_2}), highlighting the specific failure modes that contributed to these errors.

\begin{table*}[h!]
\caption{Examples of LLMs where Users Incorrectly Accept LLM HoT Answers of DROP (Reading Comprehension task).}
\begin{tabular}{p{\textwidth}} 
\hline
DROP \textcolor{red}{The model misses some key facts (37-yard field goal is not highlighted by the LLM)} \\ \hline
\textbf{Question}: ... New Orleans would respond with Brees' 3-yard touchdown pass to tight end David Thomas.  \sethlcolor{fact1}\hl{The Saints would add onto their lead in the fourth quarter as kicker Garrett Hartley made a 46-yard and a 19-yard field goal.}  The 49ers would strike back as Gore got a 7-yard touchdown run, followed by Smith's 2-point conversion pass to tight end Vernon Davis.  In the end, New Orleans got the last laugh as Hartley booted the game-ending 37-yard field goal. How many field goals did Garrett Hartley kick in the fourth quarter?\\
\textbf{LLM Answer}: The passage states that \sethlcolor{fact1}\hl{in the fourth quarter, Garrett Hartley kicked a 46-yard and a 19-yard field goal}. This means he kicked two field goals in the fourth quarter. So the answer is \{2\}. {\color{red}{\ding{55}}} (\textbf{GT: 3}) \\ \hline
DROP \textcolor{red}{Misread the Question Requirements (the LLM should focus on the requirement ``the popular uprising start in?'')} \\ \hline
\textbf{Question}: The popular uprising included large areas of \sethlcolor{fact1}\hl{Lower Bavaria}, \sethlcolor{fact2}\hl{the Innviertel} and \sethlcolor{fact3}\hl{eastern Bavaria}. The lands adjoining towns and strongholds were captured by the rebels and the peasant uprising spread to \sethlcolor{fact4}\hl{the Bavarian Forest}, parts of \sethlcolor{fact5}\hl{the Upper Palatinate} and \sethlcolor{fact6}\hl{Kelheim on the Danube}. A Braunau Parliament meeting was held in December 1705 in the then-Bavarian Braunau.... How many regions did the popular uprising start in?\\
\textbf{LLM Answer}: According to the passage, the popular uprising started in \sethlcolor{fact1}\hl{Lower Bavaria}, \sethlcolor{fact2}\hl{the Innviertel}, and \sethlcolor{fact3}\hl{eastern Bavaria} initially, then spread to \sethlcolor{fact4}\hl{the Bavarian Forest}, \sethlcolor{fact5}\hl{the Upper Palatinate}, and \sethlcolor{fact6}\hl{Kelheim on the Danube}. So the answer is \{6\}. {\color{red}{\ding{55}}} (\textbf{GT: 3}) \\ \hline
\end{tabular}
\label{tab:incorrectly_accept_1}
\end{table*}

\begin{table*}[h!]
\caption{Examples of LLMs where Users Incorrectly Accept LLM HoT Answers of GSM-Symbolic (Arithmetic task).}
\begin{tabular}{p{\textwidth}} 
\hline
GSM-Symbolic \textcolor{red}{(Correct fact highlighting but missing reasoning steps - The LLM misses one step to subtract ¥330 to ¥84 = ¥246)}\\ \hline
\textbf{Question}: Arash is raising money for a music festival. He has applied for help from the \sethlcolor{fact1}\hl{youth club, which has decided to cover 1/5 of the cost of the music festival}. How much money is Arash missing if he has \sethlcolor{fact2}\hl{¥90} and the music festival costs \sethlcolor{fact3}\hl{¥420}?\\
\textbf{LLM Answer}: The youth club will cover \sethlcolor{fact1}\hl{1/5} of the cost of the music festival, which is \sethlcolor{fact1}\hl{1/5} * \sethlcolor{fact3}\hl{¥420} = ¥84. Arash has \sethlcolor{fact2}\hl{¥90}, so he has ¥90 - ¥84 = ¥6 more than the amount covered by the youth club. Since the music festival costs ¥420, Arash is missing ¥420 - ¥90 = ¥330. {\color{red}{\ding{55}}} (\textbf{GT: 246}) \\ \hline
DROP \\ \hline
GSM-Symbolic \textcolor{red}{(Incorrect arithmetic errors - The LLM incorrectly computed 17×241=4100)} \\ \hline
\textbf{Question}: A fog bank rolls in from the ocean to cover a city. \sethlcolor{fact1}\hl{It takes 241 minutes to cover every 5 miles of the city}. If \sethlcolor{fact2}\hl{the city is 85 miles across from the oceanfront to the opposite inland edge}, how many minutes will it take for the fog bank to cover the whole city?\\
\textbf{LLM Answer}: To find the total time it will take for the fog bank to cover the whole city, we need to divide the total distance across the city by the distance covered in the given time and then multiply by the given time. The total distance across the city is \sethlcolor{fact2}\hl{85 miles} and the distance covered in the given time is \sethlcolor{fact1}\hl{5 miles}. So, the number of times the given distance fits into the total distance is \sethlcolor{fact2}\hl{85}/\sethlcolor{fact1}\hl{5} = 17. Now, we multiply this by the given time, which is \sethlcolor{fact1}\hl{241 minutes}. Therefore, the total time it will take for the fog bank to cover the whole city is 17 * \sethlcolor{fact1}\hl{241 minutes} = 4100 minutes... {\color{red}{\ding{55}}} (\textbf{GT: 4097})\\ \hline
\end{tabular}
\label{tab:incorrectly_accept_2}
\end{table*}

\clearpage
\section{Tagging-Induced Token Overhead}
\label{sec:token_overhead}
In this section, we analyze the token overhead introduced by HoT to quantify how much additional context it consumes and to assess its impact on long-context reasoning.

\begin{table}[h]
\centering
\caption{Token Overhead Across Long-context Datasets. Overal, although the tags add about $61.80\%$ overhead on average, they lead to clear improvements in both accuracy and traceability. Note: RQ is Reformatted Question.}
\resizebox{\columnwidth}{!}{
\begin{tabular}{lccccccc}
\hline
\textbf{Dataset} & Model & {Question} & {CoT} & {HoT} & {HoT} & {HoT} & {Overhead} \\
 &  & {Avg} & {Avg} & {Total} & {RQ} & {Answer} & (HoT RQ/HoT Total) \\
\hline
Census & \geminilogo & 343 & 62 & 469 & 361 & 105 & 77.1\% \\
Break & \geminilogo & 345 & 72 & 481 & 371 & 108  & 77.0\% \\
bbeh\_spatial\_reasoning & \geminilogo & 1445 & 594 & 1810 & 1200 & 608 & 66.3\% \\
bbeh\_time\_arithmetic & \geminilogo & 324 & 498 & 796 & 256 & 537 & 32.2\% \\
bbeh\_shuffle\_objects & \geminilogo & 3244 & 911 & 1073 & 608 & 462 & 56.6\% \\
\hline
\textbf{Mean} $\pm$ \textbf{Std} & - & $1140 \pm 1136$ & $427 \pm 324$ & $926 \pm 495$ & $559 \pm 340$ & $364 \pm 215$ & $61.8\% \pm 16.7$ \\
\hline
\end{tabular}
}
\label{tab:token_overhead}
\end{table}

To compute token overhead, we compute the relative increase in
token length introduced by HoT. Let $L_{\text{RQ}}$ denote
the tokenized length of the reformatted question and $L_{\text{HoT\_total}}$ the
total tokenized length of the HoT (including reformatted question and answer). The overhead for that is:
\[
\text{Overhead} = \frac{L_{\text{RQ}}}{L_{\text{HoT\_total}}} \times 100.
\]
We compute the mean and std overhead across all long-context datasets in \cref{tab:token_overhead}, showing that HoT introduces moderate additional tokens while preserving strong performance and improving traceability.

\section{Estimated User Verification Accuracy}
\label{sec:human_estimated_accuracy}

Because testing users on all 17 datasets (\cref{tab:full_performance_math,tab:full_performance_qa_rc,tab:full_performance_logical}) is infeasible, here, we estimate the verification accuracy of users when highlights are displayed or not (\ie HoT vs. CoT).

We estimate the verification accuracy $P$ of users on a given dataset by the following equation:
\[
P = M_c \times H_c + (1 - M_c) \times H_i
\]
where $M_c$ is the actual LLM accuracy for the dataset (\cref{tab:full_performance_math,tab:full_performance_qa_rc,tab:full_performance_logical}), 
$H_c$ and $H_i$ are the \human human accuracy scores on the \texttt{correct} and \texttt{incorrect} cases, respectively (\cref{tab:human_results}).

Since LLMs tend to have relatively high accuracy of $\geq$80\% on most 17 datasets, we find that the averaged estimated verification accuracy of \hot across all types of datasets 
(\cref{sec:projected_accuracy_app}) is higher to CoT by an average of \increasenoparent{3.33} across all five LLMs (\cref{tab:avg_projected_accuracy}).

\begin{table}[h]
\centering
\begin{minipage}{0.6\textwidth}
\centering
\small
\resizebox{\columnwidth}{!}{%
\newcolumntype{g}{>{\columncolor{gray!10}}l}
\begin{tabular}{lgggg}
\hline
\rowcolor{white} Model & \makecell{Prompt} & \makecell{Arithmetic\\tasks\\Mean ($\Delta$)} & \makecell{Reading Comp.\\tasks\\Mean ($\Delta$)} & \makecell{Logical Reasoning\\tasks\\Mean ($\Delta$)} \\ 
\hline

\rowcolor{white}
\multirow{2}{*}{
  \raisebox{-0.3\height}{
    \hspace*{-0.45em}
    \geminiflashlogo
  }
}
 & CoT & 78.15  & 77.40  & 77.63  \\ 
 & \hot ($\Delta$) & \textbf{82.44} \increase{4.28} & \textbf{79.25} \increase{1.85} & \textbf{79.80} \increase{2.16} \\ 
\hline

\rowcolor{white}
\multirow{2}{*}{
  \raisebox{-0.3\height}{
    \hspace*{0.03em}
    \geminilogo
  }
}
 & CoT  & 78.15  & 77.76 & 78.03  \\ 
 & \hot ($\Delta$) & \textbf{82.69} \increase{4.54} & \textbf{80.44} \increase{2.68} & \textbf{81.92} \increase{3.89} \\ 
\hline

\rowcolor{white}
\multirow{2}{*}{
  \raisebox{-0.3\height}{
    \hspace*{0.03em}
    \llamaMedlogo
  }
}
& CoT & 78.29  & 77.72  & 77.74  \\ 
& \hot ($\Delta$) & \textbf{82.14} \increase{3.86} & \textbf{80.47} \increase{2.75} & \textbf{80.50} \increase{2.30} \\ 
\hline

\rowcolor{white}
\multirow{2}{*}{
  \raisebox{-0.3\height}{
    \hspace*{0.03em}
    \llamaBiglogo
  }
}
 & CoT  & 78.37  & 78.01  & 78.07  \\ 
 & \hot ($\Delta$) & \textbf{82.75} \increase{4.38} & \textbf{81.39} \increase{3.38} & \textbf{81.88} \increase{3.81} \\ 
\hline

\rowcolor{white}
\multirow{2}{*}{
  \raisebox{-0.3\height}{
    \hspace*{0.03em}
    \gptlogo
  }
}
 & CoT  & 78.25  & 77.70  & 78.10  \\ 
 & \hot ($\Delta$) & \textbf{82.33} \increase{4.08} & \textbf{80.21} \increase{2.50} & \textbf{82.34} \increase{3.53} \\ 
\hline

\rowcolor{white}
\multicolumn{1}{l}{Mean $\Delta$} & 
 & \increasenoparentwithspace{4.23} & \increasenoparentwithspace{2.63} & \increasenoparentwithspace{3.14} \\
\hline
\end{tabular}%
}
\end{minipage}%
\hfill
\begin{minipage}{0.38\textwidth}
\caption{The estimated \human human verification accuracy on all 17 benchmarks is higher with highlights (\hot) rather than without highlights (CoT). 
More results in \cref{sec:projected_accuracy_app}.}
\label{tab:avg_projected_accuracy}
\end{minipage}
\end{table}

\clearpage
\subsection{Detail of the estimated user verification accuracy}
\label{sec:projected_accuracy_app}
The following tables (from \cref{tab:full_projected_performance_math} to \cref{tab:full_projected_performance_logical}) are the estimated verification accuracies of humans if they were to verify each of the CoT and \hot results that we gather in our benchmarking. We calculate these results based off the recorded verification accuracies in the human study (\cref{tab:human_results}) and the equation described in \cref{sec:human_verification}.

\begin{table*}[ht]
\centering
\caption{Estimated human verification accuracy for \hot on arithmetic tasks.}
\resizebox{\textwidth}{!}{%
\newcolumntype{g}{>{\columncolor{gray!10}}l}
\begin{tabular}{lgggggggg|g}
\hline
\rowcolor{white} Model      & Prompt   & GSM8K & MultiArith & ASDiv & SVAMP & AQUA & R-GSM & GSM-Symbolic & Task Mean ($\Delta$) \\ 
\hline

\rowcolor{white} 
\multirow{2}{*}{
  \raisebox{-0.3\height}{
    \hspace*{-0.45em}
    \geminiflashlogo
  }
}
 & CoT            &  78.40 & 78.73 & 78.60 & 78.32 & 77.57 & 77.87 & 77.59 & 78.15 \\ 
  & \hot ($\Delta$) & \textbf{82.72} \increase{4.32} & \textbf{84.30} \increase{5.57} & \textbf{83.65} \increase{5.05} & \textbf{82.96} \increase{4.64} & \textbf{83.33} \increase{5.76} & \textbf{80.45} \increase{2.58} & \textbf{79.65} \increase{2.06} & \textbf{82.44}\increase{4.29} \\ 
\hline

\rowcolor{white} 
\multirow{2}{*}{
  \raisebox{-0.3\height}{
    \hspace*{0.03em}
    \geminilogo
  }
}
& CoT & 78.50 & 78.63 & 78.63 & 78.45 & 77.66 & 77.36 & 77.85 & 78.15
\\ 
& \hot ($\Delta$) & \textbf{83.15} \increase{4.65} & \textbf{84.25} \increase{5.62} & \textbf{83.91} \increase{5.28} & \textbf{83.25} \increase{4.80} & \textbf{82.04} \increase{4.38} & \textbf{81.80} \increase{4.44} & \textbf{80.44} \increase{2.59} & \textbf{82.69} \increase{4.54} \\ 
\hline

\rowcolor{white} 
\multirow{2}{*}{
  \raisebox{-0.3\height}{
    \hspace*{0.03em}
    \llamaMedlogo
  }
}
& CoT  & 78.45 & 78.74 & 78.56 & 78.34 & 78.20 & 78.08 & 77.63 & 78.29 \\ 
& \hot ($\Delta$) & \textbf{82.90} \increase{4.45} & \textbf{84.30} \increase{5.56} & \textbf{83.62} \increase{5.06} & \textbf{83.04} \increase{4.70} & \textbf{80.64} \increase{2.44} & \textbf{81.26} \increase{3.18} & \textbf{79.25} \increase{1.62} & \textbf{82.14}\increase{3.85}\\ 
\hline

\rowcolor{white}
\multirow{2}{*}{
  \raisebox{-0.3\height}{
    \hspace*{0.03em}
    \llamaBiglogo
  }
}
& CoT             & 78.52 & 78.72 & 78.61 & 78.38 & 77.97 & 78.20 & 78.16 & 78.37 \\ 
& \hot ($\Delta$) & \textbf{83.26} \increase{4.74} & \textbf{84.25} \increase{5.53} & \textbf{83.72} \increase{5.11} & \textbf{83.16} \increase{4.78} & \textbf{81.23} \increase{3.26} & \textbf{82.07} \increase{3.87} & \textbf{81.53} \increase{3.37} & \textbf{82.75} \increase{4.38} \\ 
\hline

\rowcolor{white} 
\multirow{2}{*}{
  \raisebox{-0.3\height}{
    \hspace*{0.03em}
    \gptlogo
  }
}
& CoT             & 78.50 & 78.73 & 78.62 & 78.44 & 77.42 & 78.08 & 77.96 & 78.25 \\ 
& \hot ($\Delta$) & \textbf{83.22} \increase{4.72} & \textbf{84.30} \increase{5.57} & \textbf{83.38} \increase{4.76} & \textbf{83.07} \increase{4.63} & \textbf{79.36} \increase{1.94} & \textbf{81.66} \increase{3.58} & \textbf{81.33} \increase{3.37} & \textbf{82.33} \increase{4.08} \\ 
\hline 

\rowcolor{white}
\multicolumn{2}{l}{Model Mean $\Delta$} & \increasenoparentwithspace{4.58} & \increasenoparentwithspace{5.57} & \increasenoparentwithspace{5.05} & \increasenoparentwithspace{4.71} & \increasenoparentwithspace{3.56} & \increasenoparentwithspace{3.53} & \increasenoparentwithspace{2.60} & \increasenoparentwithspace{4.23}
\\ \hline
\end{tabular}%
}
\label{tab:full_projected_performance_math}
\end{table*}

\begin{table*}[ht]
\centering
\caption{Estimated human verification accuracy for \hot on reading comprehension tasks.}
\resizebox{\textwidth}{!}{%
\small
\newcolumntype{g}{>{\columncolor{gray!10}}l}
\begin{tabular}{lgggggg|g}
\hline
\rowcolor{white} Model & Prompt   & StrategyQA & SpartQA & Date & Break & Census & Task Mean ($\Delta$)\\ 
\hline

\rowcolor{white} 
\multirow{2}{*}{
  \raisebox{-0.3\height}{
    \hspace*{-0.45em}
    \geminiflashlogo
  }
}
& CoT  & 77.25 & 75.32 & 77.83 & 77.72 & 78.14 & 77.45 \\ 
& \hot ($\Delta$) & \textbf{78.48} \increase{1.23} & 72.37 \decrease{2.95} & \textbf{80.28} \increase{2.45} & \textbf{80.42} \increase{2.70} & \textbf{81.53} \increase{3.39} & \textbf{79.25} \increase{1.80} \\ 
\hline

\rowcolor{white} 
\multirow{2}{*}{
  \raisebox{-0.3\height}{
    \hspace*{0.03em}
    \geminilogo
  }
}
& CoT  & 77.60 & 76.28 & 78.36 & 77.90 & 78.26 & 77.76 \\ 
& \hot ($\Delta$) & \textbf{79.58} \increase{1.98} & 74.06 \decrease{2.22} & \textbf{83.26} \increase{4.90} & \textbf{80.75} \increase{2.85} & \textbf{82.27} \increase{4.01} & \textbf{80.44} \increase{2.68} \\ 
\hline

\rowcolor{white} 
\multirow{2}{*}{
  \raisebox{-0.3\height}{
    \hspace*{0.03em}
    \llamaMedlogo
  }
}
& CoT  & 76.77 & 76.56 & 78.23 & 78.06 & 78.42 & 77.72 \\ 
& \hot ($\Delta$) & \textbf{79.86} \increase{3.09} & 74.72 \decrease{1.84} & \textbf{82.10} \increase{3.87} & \textbf{81.20} \increase{3.14} & \textbf{82.79} \increase{4.37} & \textbf{80.47} \increase{2.75} \\ 
\hline

\rowcolor{white}
\multirow{2}{*}{
  \raisebox{-0.3\height}{
    \hspace*{0.03em}
    \llamaBiglogo
  }
}
& CoT & 77.83 & 76.81 & 78.51 & 78.16 & 78.37 & 78.01 \\ 
& \hot ($\Delta$) & \textbf{81.06} \increase{3.23} & 76.27 \decrease{0.54} & \textbf{83.75} \increase{5.24} & \textbf{81.61} \increase{3.45} & \textbf{82.87} \increase{4.50} & \textbf{81.39} \increase{3.38} \\ 
\hline

\rowcolor{white} 
\multirow{2}{*}{
  \raisebox{-0.3\height}{
    \hspace*{0.03em}
    \gptlogo
  }
}
& CoT & 77.74 & 75.83 & 78.58 & 77.93 & 77.89 & 77.70
\\ 
& \hot ($\Delta$) & \textbf{80.15} \increase{2.41} & 72.55 \decrease{3.28} & \textbf{83.67} \increase{5.09} & \textbf{80.79} \increase{2.86} & \textbf{81.75} \increase{3.86} & \textbf{80.21} \increase{2.51} \\ 
\hline

\rowcolor{white}
\multicolumn{2}{l}{Model Mean $\Delta$} & \increasenoparentwithspace{2.39} & \decreasenoparentwithspace{2.17} & \increasenoparentwithspace{4.31} & \increasenoparentwithspace{3.00} & \increasenoparentwithspace{4.03} & \increasenoparentwithspace{2.63}
\\ \hline
\end{tabular}%
}
\label{tab:full_projected_performance_qa_rc}
\end{table*}

\begin{table*}[ht]
\caption{Estimated human verification accuracy for \hot on the subset of logical tasks in BBH.}
\resizebox{\textwidth}{!}{%
\small
\newcolumntype{g}{>{\columncolor{gray!10}}l}
\centering
\begin{tabular}{lgggggg|g}
\hline
\rowcolor{white} Model & Prompt   & \makecell{BBH: deduction \\five \\objects} & \makecell{deduction \\seven \\objects} & \makecell{reasoning \\colored \\objects} & \makecell{causal\\judgement} & navigate & Task Mean ($\Delta$)\\ 
\hline

\rowcolor{white}
\multirow{2}{*}{
  \raisebox{-0.3\height}{
    \hspace*{-0.45em}
    \geminiflashlogo
  }
}
& CoT& 77.40 & 77.14 & 78.40 & 76.93 & 78.32 & 77.63 \\ 
& \hot ($\Delta$) & \textbf{79.15} \increase{1.75} & 78.32 \increase{1.18} & \textbf{83.07} \increase{4.67} & 76.08 \decrease{0.85} & \textbf{82.36} \increase{4.04} & \textbf{79.80} \increase{2.17} \\ 
\hline

\rowcolor{white}
\rowcolor{white} 
\multirow{2}{*}{
  \raisebox{-0.3\height}{
    \hspace*{0.03em}
    \geminilogo
  }
}
& CoT & 78.32 & 77.88 & 78.56 & 77.14 & 78.27 & 77.99 \\ 
& \hot ($\Delta$) & \textbf{84.14} \increase{5.82} & \textbf{81.17} \increase{3.29} & \textbf{83.67} \increase{5.11} & 77.19 \increase{0.05} & \textbf{83.43} \increase{5.16} & \textbf{81.92} \increase{3.93}
\\ 
\hline

\rowcolor{white}
\rowcolor{white} 
\multirow{2}{*}{
  \raisebox{-0.3\height}{
    \hspace*{0.03em}
    \llamaMedlogo
  }
}
& CoT & 78.32 & 77.45 & 78.27 & 76.68 & 77.98 & 77.74 \\ 
& \hot ($\Delta$) & \textbf{82.72} \increase{4.40} & \textbf{79.63} \increase{2.18} & \textbf{82.60} \increase{4.33} & 75.92 \decrease{0.76} & \textbf{81.65} \increase{3.67} & \textbf{80.50} \increase{2.76} \\ 
\hline

\rowcolor{white}
\rowcolor{white} 
\multirow{2}{*}{
  \raisebox{-0.3\height}{
    \hspace*{0.03em}
    \llamaBiglogo
  }
}
& CoT & 78.51 & 78.11 & 78.59 & 76.68 & 78.48 & 78.07 \\ 
& \hot ($\Delta$) & \textbf{83.67} \increase{5.16} & \textbf{81.53} \increase{3.42} & \textbf{83.67} \increase{5.08} & 76.88 \increase{0.20} & \textbf{83.67} \increase{5.19} & \textbf{81.88} \increase{3.81} \\ 
\hline

\rowcolor{white}
\rowcolor{white} 
\multirow{2}{*}{
  \raisebox{-0.3\height}{
    \hspace*{0.03em}
    \gptlogo
  }
}
& CoT  & 78.38 & 77.82 & 78.69 & 77.07 & 78.56 & 78.10 \\ 
& \hot ($\Delta$) & \textbf{82.36} \increase{3.98} & \textbf{80.46} \increase{2.64} & \textbf{84.14} \increase{5.45} & 81.32 \increase{4.25} & \textbf{83.43} \increase{4.87} & \textbf{82.34} \increase{4.24} \\ 
\hline
\rowcolor{white}
\multicolumn{2}{l}{Model Mean $\Delta$} & \increasenoparentwithspace{4.22}&\increasenoparentwithspace{2.54}&\increasenoparentwithspace{4.93}&\increasenoparentwithspace{0.58}&\increasenoparentwithspace{4.59} & \increasenoparentwithspace{3.14}
\\ \hline
\end{tabular}%
}
\label{tab:full_projected_performance_logical}
\end{table*}
\clearpage

\section{\hot causes Llama-3.1 variants to allocate more attention around important tokens}
\label{sec:attention_allocation}

In order to compute the attention score for the XML tags and their contents, across all heads for the final layer of the model, we measure the amount of attention on (1) the XML tag tokens (e.g., <fact1>, </fact1>); and also (2) the regular content tokens inside those XML tags. 

Let $$A^{L,h} \in \mathbb{R}^{T \times T}$$ be the attention matrix of the final layer (L) and head (h).

Let $S_{\text{content}}$ and $S_{\text{tags}}$ be the token indices for content inside <fact\#> spans and the XML tokens themselves, respectively.

\subsec{Per-head attention scores:}
\begin{equation}
Attn^h_{content} = \frac{1}{|S_{content}|} \sum_{i \in S_{content}} \sum_{j=1}^{T} A^{L,h}_{j,i}
\qquad
Attn^h_{tags} = \frac{1}{|S_{tags}|} \sum_{i \in S_{tags}} \sum_{j=1}^{T} A^{L,h}_{j,i}
\end{equation}

\subsec{Final scores (averaged across all H heads):}
\begin{equation}
Attn_{content} = \frac{1}{H} \sum_{h=1}^{H} Attn^h_{content}
\qquad
Attn_{tags} = \frac{1}{H} \sum_{h=1}^{H} Attn^h_{tags}
\end{equation}
For a direct comparison, for the CoT responses, we identify and compute the attention score for the key-fact tokens (that would have been wrapped in XML tags in HoT responses).

\begin{table}[H]
\centering
\caption{Comparison of total percent of attention (\%) allocated to tokens in Llama-3.1-70B and Llama-3.1-405B for CoT and \hot.}
\resizebox{0.50\columnwidth}{!}{
\begin{tabular}{lcc|c}
\toprule
Method & Llama-3.1-70B  & Llama-3.1-405B & Avg \\
\midrule
CoT Fact Tokens & 3.34 & 2.26 & 2.80\\
HoT Total       & \textbf{8.40} & \textbf{5.30} & \textbf{6.85}\\
\rowcolor{gray!10} \quad Fact Tokens & 2.79 & 1.85 & 2.32\\
\rowcolor{gray!10} \quad XML Tokens  & 5.61 & 3.45 & 4.53\\
\bottomrule
\end{tabular}
}
\label{tab:attention_allocation}
\end{table}

We hypothesize that the attention on the XML <fact\#> tags is higher because the embeddings for the these <fact\#> might actually contain the information of the content inside those opening and closing tags. Due to the self-attention mechanism, this is recently an observation in Transformers both in text (e.g. "Attention sink" \citep{attentionsink}) and images ("Vision Transformers need Registers" \citep{darcet2023vision}). In images, a Vision Transformer classifier's attention was seen to focus on a seemingly non-relevant background token; yet; these high-attention token embeddings actually contains the global information for the entire image.

\subsec{Experiment} To test this hypothesis, we analyzed attention patterns in \llamaMedlogo and \llamaBiglogo across responses in all datasets, comparing \hot responses with XML tags to equivalent CoT responses without tags. 

\subsec{Results} Our results reveal that XML tag tokens receive substantially higher attention than the fact tokens they enclose (\cref{tab:attention_allocation}). Surprisingly, fact tokens wrapped in XML tags receive less attention in \hot compared to CoT.

Rather than directly increasing attention to fact tokens, XML tags appear to redirect attention away from non-essential tokens in the reasoning process. This redistribution of attention effectively increases the relative focus on important fact tokens, which could be a contributing cause to improved model performance.

\section{Tagging arbitrary phrases in the answer hurts HoT accuracy}
\label{sec:misaligned_tag}

Given that the key facts are already accurately tagged in the question, an interesting question is whether associated facts in the answer need to be accurately tagged for \hot to work.
Does the correspondence between tags in the question and those matching in the answer matter?

\begin{figure}[ht]
  \centering
  \includegraphics[width=\textwidth]{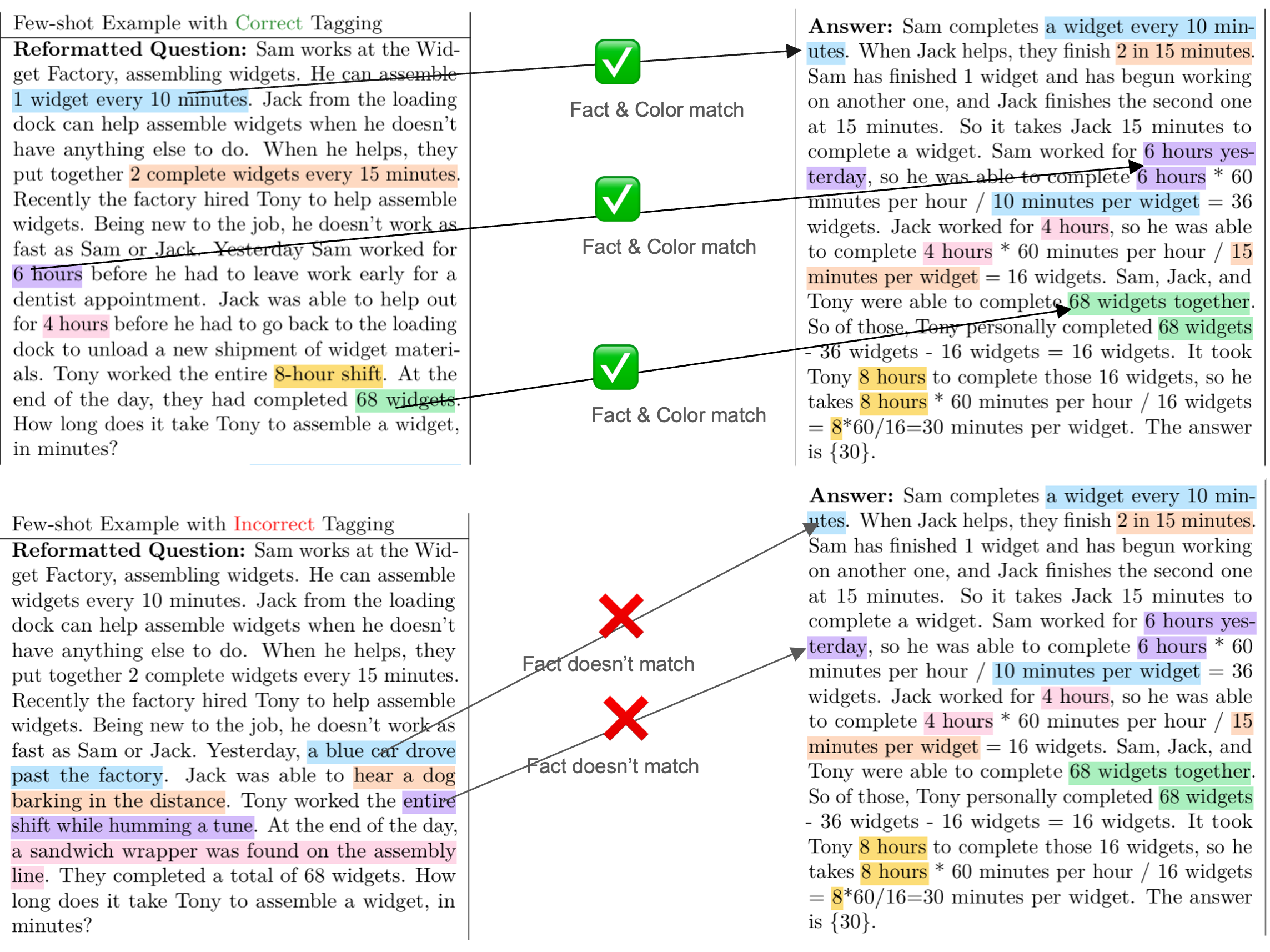}
\caption{On the first row, the tags in the answers correspond to the tags in the question, ensuring alignment. On the second row, the tags in the question and answer do not match, leading to inconsistencies, e.g. \sethlcolor{fact3}\hl{entire shift while humming a tune} in the question does not match with \sethlcolor{fact3}\hl{6 hours yesterday} in the answer.}
\label{fig:incorrect_tags}
\end{figure}

Here, we attempt to understand the importance of the correspondence by keeping \hot few-shot demonstrations intact, but relocate the tags in all answers to a random phrase (see examples in \cref{fig:incorrect_tags}).
How would this mismatch impact LLM accuracy on the task?

\begin{figure*}[!ht]
    \centering
    \begin{subfigure}
    \centering
    \captionsetup{labelformat=empty} 
    (a)
    
    \includegraphics[width=\columnwidth]{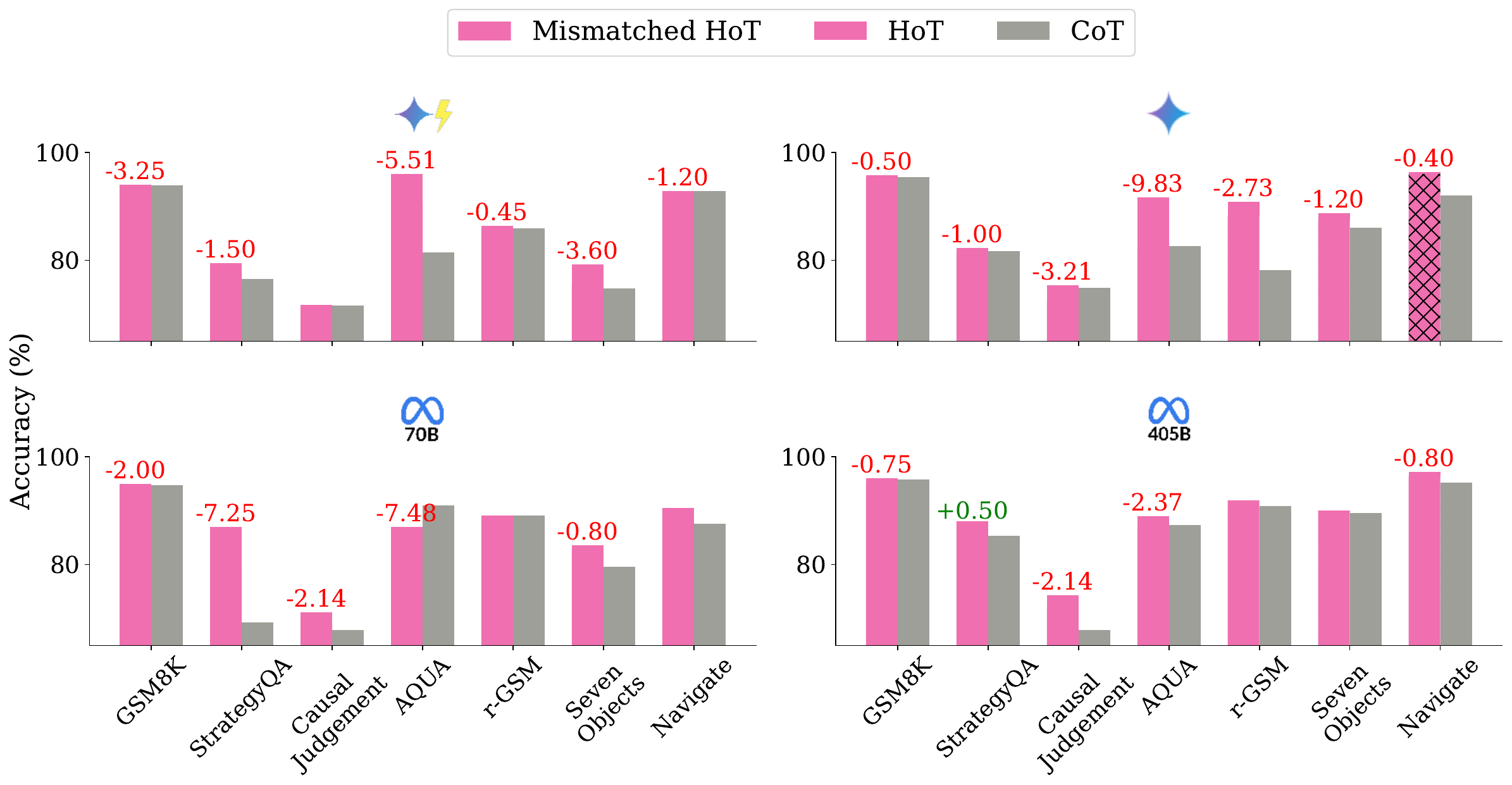}
    \end{subfigure}

    \begin{subfigure} 
        \centering
        \captionsetup{labelformat=empty}
        \resizebox{0.75\columnwidth}{!}{%
            \begin{tabular}{ccccc|c}
                \hline
                Model & \geminiflashlogo & \geminilogo & \llamaMedlogo & \llamaBiglogo & Model Mean $\downarrow$ \\
                \hline
                \multicolumn{6}{c}{\cellcolor{gray!25}{(b) Mean $\Delta$(mismatched \hotcolored - \hotcolored) accuracy across four models.}} \\ \hline
                 Task Mean $\downarrow$ & \decreasenoparent{2.22} & \decreasenoparent{2.70} & \decreasenoparent{2.81} & \decreasenoparent{0.79} & \decreasenoparent{2.13} \\
                \hline
                \multicolumn{6}{c}{\cellcolor{gray!25}{(c) Mean $\Delta$(mismatched \hotcolored - \cotcolored) accuracy across four models.}} \\ \hline
                Task Mean $\downarrow$ & 1.00 & 1.64 & 0.63 & 1.56 & 1.21 \\ \hline
            \end{tabular}
        }
    \end{subfigure}
    
    \caption{
    \textbf{(a)} Moving tags in the answer of HoT to random phrases (\ie, Mismatched HoT) results in substantial accuracy drops across many datasets \eg, this $\Delta$ is \decreasenoparent{3.25} for \geminiflashlogo.
    \\\textbf{(b)} The average accuracy drop of Mismatched \hotcolored w.r.t. \hotcolored is \decreasenoparent{2.13} pp. 
    \\\textbf{(c)} On average across 4 LLMs, Mismatched \hotcolored still outperforms vanilla \cotcolored by a mean gain of \increasenoparent{1.21} pp.}
    \label{fig:incorrect_tag_results}
\end{figure*}

\subsec{Experiment} We compare the accuracy of 4 models (\geminiflashlogo, \geminilogo, \llamaMedlogo, and \llamaBiglogo) using few-shot prompts with original \hot tags against those with mismatched tags. 
To generate examples with mismatched tags, we intentionally alter the original \hot few-shot examples by changing the tag locations in the answers. 
We run on 7 datasets (GSM8K, StrategyQA, Causal Judgement, AQUA, Seven Objects, Navigate).

\subsec{Results} Mismatched tagging reduces \hotcolored accuracy by \decreasenoparent{2.13} pp in average (\cref{fig:incorrect_tag_results}b).
However, the impact varies substantially across different datasets.
 For example, the accuracy drop is much higher in AQUA \decreasenoparent{9.93} pp compared to GSM8K \decreasenoparent{0.50} pp. This might be because, in AQUA, when exposed to few-shot examples with mismatched tags, LLMs are less likely to generate tags in their responses, resulting in a performance drop compared to when they are exposed to matched tags.

Even when the question and answer tags are mismatched, \hotcolored still outperforms \cotcolored by \increasenoparent{1.21} pp (see \cref{fig:incorrect_tag_results}c). Despite the misalignment, the LLM can still learn from few-shot examples and apply correct tagging to new questions, preserving some of HoT’s advantages over CoT.
This also explains the marginal decrease in some datasets, such as Navigate (\decreasenoparent{0.80}) and Seven Objects (\decreasenoparent{0.50}) pp (\cref{fig:incorrect_tag_results}a).



\clearpage
\section{\texorpdfstring{\robot \llmjudge} \hot does not hurt LLM-as-a-Judge accuracy.}\label{sec:llm_judge}

\subsec{Motivation}
The relevance of \llmjudge methods has grown as VLMs are increasingly used for automatic feedback in post-training fine-tuning \citep{dubey2024llama} and evaluation \citep{deepseekr1}. Since \hot improves human verification compared to CoT (\cref{tab:human_results}), we investigate whether it similarly benefits multimodal models’ verification ability.

Human users view \hot outputs with \texttt{<fact>} tokens replaced by color highlighting to reduce reading burden. In contrast, VLMs can easily process dense text. Therefore, to evaluate whether \hot improves VLM verification, we test models on both raw \hot text and screenshots of regex-parsed, color-highlighted \hot outputs presented as images to multimodal models.

\paragraph{Experimental Setup} We collect equal numbers of \hot responses from \llamaA, \llamaB, \llamaC, \geminipro, and \geminiflash on GSM-Symbolic and DROP, yielding 500 total LLM responses, evenly split between correct and incorrect answers. To construct the CoT set, we remove all XML tags from these 500 \hot responses, producing a corresponding set of 500 CoT QA pairs.

We test \gpto, \gptmini (\gptminilogo), \geminipro, \geminiflash, Deepseek R1 (\deepseek) (which is not multimodal) and Llama 3.2 90B (vision) (\llamavision) on their ability to accurately verify answers in \hot and CoT format. The prompts used for text-based and image-based verification are in \cref{fig:visual_hot_prompts}.

\begin{figure*}[ht!]
\begin{tcolorbox}[title=Text-based and Image-based Prompts for visual \hot Verification]
TEXT\_ONLY\_PROMPT = "Read the question and answer below. Verify if the answer is correct or is flawed. After thinking and solving provide your final answer as a yes and no in curly braces. {yes} means the answer is correct, {no} means the answer is flawed."

IMAGE\_TEXT\_PROMPT = "First examine the question and answer in the image. Then verify if the answer is correct or is flawed. After thinking and solving provide your final answer as a yes and no in curly braces. {yes} means the answer is correct, {no} means the answer is flawed."
\end{tcolorbox}
\caption{Text-based and Image-based prompts used in verifying HoT answers.}
\label{fig:visual_hot_prompts}
\end{figure*}

\begin{table}[h]
\centering
\begin{minipage}{0.6\textwidth}
\centering
\small
\resizebox{\columnwidth}{!}{%
\newcolumntype{g}{>{\columncolor{gray!10}}l}
\begin{tabular}{lgggggg}
\hline
\rowcolor{white} Model & \makecell{Prompt} & 
\multicolumn{2}{g}{\makecell{Accuracy When\\ Answer is Correct \greencheck}} & 
\multicolumn{2}{g}{\makecell{Accuracy When \\ Answer is Incorrect \redcross}}  \\ 
\cmidrule(lr){3-4} \cmidrule(lr){5-6} 
\rowcolor{white} & & Text & Image & Text & Image \\ 
\hline

\rowcolor{white}
\multirow{2}{*}{\raisebox{-0.3\height}{\hspace*{-0.45em}\geminiflashlogo}}
 & CoT & \textbf{63.86}  & \textbf{71.54} & 64.40 & 58.00 \\ 
 & \hot & 53.41 \decrease{10.54} & 54.00 \decrease{-17.54} & \textbf{66.94} \increase{2.54} & 67.07 \increase{9.07} \\ 
\hline

\rowcolor{white}
\multirow{2}{*}{\raisebox{-0.3\height}{\hspace*{0.03em}\geminilogo}}
 & CoT & \textbf{63.60} & 87.60 & 63.60 & 34.80\\ 
 & \hot & 61.60 \decrease{2.00} & \textbf{86.80} \decrease{0.80} & \textbf{68.00} \increase{4.40} & \textbf{36.40} \increase{1.60}\\ 
\hline

\rowcolor{white}
\multirow{2}{*}{\raisebox{-0.3\height}{\hspace*{0.03em}\gptminilogo}}
 & CoT & 83.60 & 83.20 & \textbf{52.40} & 43.20\\ 
 & \hot & \textbf{87.60} \increase{4.00} & \textbf{86.40} \increase{2.80} & 47.20 \decrease{4.80} & \textbf{43.20} \\ 
\hline

\rowcolor{white}
\multirow{2}{*}{\raisebox{-0.3\height}{\hspace*{0.03em}\gptlogo}}
 & CoT & 95.20 & 95.20 & \textbf{43.20} & 43.20 \\ 
 & \hot & \textbf{96.00} \increase{0.80} & 96.00 \increase{0.80} & 36.00 \decrease{7.20} & 36.00 \decrease{7.20}\\ 
\hline

\rowcolor{white}
\multirow{2}{*}{\raisebox{-0.3\height}{\hspace*{0.01em}\deepseek}}
 & CoT & 90.00 & - & \textbf{44.00} & - \\ 
 & \hot & \textbf{90.80} \increase{0.80} & - & 42.40 \decrease{1.80} & - \\ 
\hline

\rowcolor{white}
\multirow{2}{*}{\raisebox{-0.3\height}{\llamavision }}

 & CoT & 88.80 & \textbf{57.20} & 47.20 & \textbf{59.60} \\ 
 & \hot & \textbf{89.60} \increase{0.80} & 48.40 \decrease{8.80} & \textbf{59.60} \increase{12.40} & 52.80 \decrease{6.80}\\ 
\hline

\rowcolor{white}
& Mean $\Delta$ & \decreasenoparentwithspace{1.94} & \decreasenoparentwithspace{3.69}  & \decreasenoparentwithspace{1.27} & \increasenoparentwithspace{0.87}  \\

\rowcolor{white}
\hline
\end{tabular}
}
\end{minipage}%
\hfill
\begin{minipage}{0.38\textwidth}
\caption{\hot and CoT have mixed effects on \llmjudge accuracy across different models and modalities. Overall, whether \hot is presented as text or images shows little correlation with LLM verification accuracy. In summary, while \hot improves answer accuracy (\cref{tab:full_performance_math}, \cref{tab:full_performance_qa_rc}, \cref{tab:full_performance_logical}), it does not provide consistent gains in \llmjudge accuracy over CoT.}
\label{tab:llm_judge_results_combined}
\end{minipage}
\end{table}

\subsec{Results} On average, \hot improves verification accuracy for models such as \geminipro and \gptmini, but degrades performance for \geminiflash and \gpto (\cref{tab:llm_judge_results_combined}). Similar to human evaluators, \gptlogo models and \deepseek benefit from \hot when the answer reasoning is correct, whereas Gemini models benefit only when the reasoning is incorrect. Notably, \hot yields a substantial \increasenoparent{12.40} pp improvement over CoT on incorrect LLM answers for Llama 3.2 90B (\cref{tab:llm_judge_results_combined}). However, when \hot is presented as images, CoT performs better. The near-50\% accuracy of Llama 3.2 90B across both CoT and \hot in image form suggests near-random model behavior.

\clearpage

\section{XML Tag Ablation Study}
\label{sec:tag_ablation}
\subsection{Motivation}
\hot responses use \texttt{<fact\{num\}></fact\{num\}>}
style XML tags in order to link facts in the question to facts in the answer. However, there may be better formats of tags that could exist. This motivates us to perform a small-scale ablation study with \gpto in order to determine which format gives the best accuracy.

\subsection{Experimental Setup}
We test the accuracy of \hot responses with several different styles of tags on 200 random samples from 3 different datasets on \gpto with the setting of \texttt{temperature = 0}.

\subsection{Results}

Of all the different tested formats, \texttt{<fact\{num\}>} is the most accurate (\cref{tab:xml_tag_results}). Therefore, we chose this format when running large-scale evaluations across all benchmarks.

\begin{table}[h!]
    \centering
    \caption{\texttt{<fact\{num\}>} tags are the most effective format for answer accuracy in \hot.}
    \rowcolors{1}{white}{white}
    \resizebox{0.55\columnwidth}{!}{
    \begin{tabular}{c|ccc}
         Tag Type & SpartQA & R-GSM & GSM8K \\\hline
         
         \texttt{<fact\{num\}>} (\hot) & \textbf{59.00} & \textbf{93.50}  & \textbf{96.50} \\\hline
         
         \texttt{<FACT\{num\}>} & 51.00 & 91.50  & -    \\\hline
         
         \texttt{<key\_fact\{num\}>} & 54.00 & 91.50  & -    \\\hline
         
         \texttt{<data\{num\}>} & 56.00 & 92.00  & 95.00    \\\hline
         
         \texttt{<important\{num\}>} & 54.00 & 92.50  & 95.00    \\\hline

         \texttt{<important\_quote\{num\}>} & - & -  & 94.50    \\
         
    \end{tabular}
    }
    \label{tab:xml_tag_results}
\end{table}

\section{Model Details}\label{sec:llm_providers_details}
We describe below the exact API versions and settings for each model used in our experiments in \cref{tab:model_configs}.

\begin{table}[h]
\centering
\small
\begin{tabular}{@{}llll@{}}
\toprule
\textbf{Model Family} & \textbf{Model} & \textbf{Temperature} & \textbf{Used for} \\
\midrule
\multirow{2}{*}{\shortstack{ChatGPT \\\url{platform.openai.com}}} & gpt-4o-2024-08-06 & 1.0 & Query \\
\cmidrule(lr){2-4}
& gpt-4o-2024-11-20 & API Default & \llmjudge \\
& gpt-4o-mini-2024-07-18 & API Default & \llmjudge \\
\midrule
\multirow{3}{*}{\shortstack{Gemini\\\url{aistudio.google.com}}} & gemini-1.5-pro-002 & 1.0 & Query \\
& gemini-1.5-flash-002 & 1.0 & Query \\
& gemini-2.0-flash-thinking-exp-01-21 & 1.0 & Query \\
\midrule
\multirow{4}{*}{\shortstack{Llama \\3.1 models via SambaNova\\ 3.2 via Together AI}} & Meta-Llama-3.1-8B-Instruct & 0.6, top\_p: 0.9 & Query \\
& Meta-Llama-3.1-70B-Instruct & 0.6, top\_p: 0.9 & Query \\
& Meta-Llama-3.1-405B-Instruct & 0.6, top\_p: 0.9 & Query \\
& meta-llama/Llama-3.2-90B-Vision-Instruct-Turbo & 0.6, top\_p: 0.9 & Query \\
\midrule
Deepseek \\Via Together AI API & deepseek-ai/DeepSeek-R1 & API Default & Query \\\\
\midrule
\multirow{2}{*}{\shortstack{Qwen via SambaNova,\\ QwQ via official demo}} & Qwen2.5-Coder-32B-Instruct & 0.7 & Query \\
& \qwq & -- & Query \\
\bottomrule
\end{tabular}
\caption{Model configurations and API settings used in the benchmark evaluation.}
\label{tab:model_configs}
\end{table}

\clearpage

\section{\hot System Prompt}\label{sec:system_prompt}

\begin{figure}[ht!]
\begin{tcolorbox}[boxrule=0pt, frame empty]
I want you to answer this question but your explanation should contain references referring back to the information in the question. To do that, first, re-generate the question with proper tags for key phrases, the key phrases that are most relevant to answering the question ``\textcolor{gray}{\texttt{\{last\_sentence\}}}'' and then generate your answers. The output format is as follows:

Reformatted Question:

Answer:
\end{tcolorbox}
\caption{HoT instruction to LLMs where 
\textcolor{gray}{\small \texttt{\{last\_sentence\}}} is the last sentence, \eg ``how much money
are all her jewels worth?'' in the \textcolor{fact2!300}{\texttt{\{Question\}}} section of the prompt (\cref{fig:prompt_structure}).
}
\label{fig:hot_instruction}
\end{figure}

\section{\hot Visualization Detail Code}
\label{sec:code}
\begin{figure}[ht]
    \centering
    \begin{adjustbox}{width=0.85\textwidth}
        \begin{lstlisting}[language=Python, label=code:add_color_tags, breaklines=true, basicstyle=\ttfamily\footnotesize]
import re

char_2_color_map = {
    'fact1': "#FF5733",  # Red
    'fact2': "#33FF57",  # Green
    'fact3': "#3357FF",  # Blue
    'fact4': "#FF33A1",  # Pink
}

def add_color_to_tags(text):
    for tag, color in char_2_color_map.items():    
        text = re.sub(f'<{tag}>', rf'\\sethlcolor{{{tag}}}\\hl{{', text)
        text = re.sub(f'</{tag}>', '}', text)

    return text

paragraph = """
The <fact1>apples are the second-cheapest</fact1>. 
The <fact2>peaches are the cheapest</fact2>. 
"""
print(add_color_to_tags(paragraph))
        \end{lstlisting}
    \end{adjustbox}
    \caption{Python code that adds color to custom tags in a text paragraph, using a mapping of tags to colors.}
    \label{fig:add_color_tags}
\end{figure}

\clearpage

\section{\hot may hurt \llamaA, \qwen performance}
\label{sec:small_models_app}
Since small models like \llamaA \citep{dubey2024llama} are less capable of handling reasoning tasks than large models (\geminipro, \llamaC, \gpto), we aim to understand if \hot can benefit these smaller models. Additionally, since \qwen is optimized for logical coding tasks, we hypothesize that it may benefit from the logical structure that \hot brings. We run the following experiment to understand if \hot is able to aid in these models' reasoning ability.

\textbf{Experiment} We evaluate \llamaA and \qwen on 13 datasets with CoT and \hot prompting.

\textbf{Results} As shown in \cref{tab:small_models_performance_all}, \hot and CoT exhibit comparable performance on smaller models, like \llamaA, and coding models, like \qwen. For some datasets, such as \textbf{AQUA} on \llamaA, slight negative trends are observed (\eg \decreasenoparent{3.93} for \llamaB), with \hot providing no significant advantage, likely due to the limited reasoning capabilities of these models. 

Qualitatively, we observe that these smaller models often fail to adhere to the tagging instructions. Specifically, \llamaA frequently tags calculated results within reasoning chains (as illustrated in \cref{tab:llama8b_repeat_fs_cases} (a)) and tends to repeat the few-shot examples in reasoning chains (\cref{tab:llama8b_repeat_fs_cases} (b) and (c)). On the other hand, \qwen often omits tagging in the answers altogether. These inconsistencies likely contribute to the inconsistent performance observed with \hot (\cref{tab:small_models_performance_all}).

\begin{table}[!ht]
    \centering
    \caption{Performance comparison of  CoT and \hot across \llamaA (\llamaSmalllogo) and \qwen (\qwenlogo), evaluated on QA (StrategyQA and Date Understanding) and logical reasoning datasets.}
    \resizebox{\textwidth}{!}{%
    \newcolumntype{g}{>{\columncolor{gray!10}}l}
   \begin{tabular}{llglglglg}
    \hline
        Model & \multicolumn{2}{c}{\llamaBiglogo} & \multicolumn{2}{c}{\llamaMedlogo} & \multicolumn{2}{c}{\llamaSmalllogo} & \multicolumn{2}{c}{\qwenlogo} \\ \hline
        \rowcolor{white}Prompt & CoT & \hot ($\Delta$) & CoT & \hot ($\Delta$) & CoT & \hot ($\Delta$)& CoT & \hot ($\Delta$) \\ \hline
        GSM8K & 95.83 & 95.83 & 94.69 & 94.62 \decrease{0.07} & 85.06 & 82.41 \decrease{2.65} & 94.24 & 94.47 \increase{0.23} \\ \hline
        MultiArith & 98.83 & 99.16 \increase{0.33} & 99.16 & 99.33 \increase{0.17} & 96.81 & 98.49 \increase{1.68} & 98.83 & 98.99 \increase{0.16} \\ \hline
        ASDiv & 97.19 & 97.38 \increase{0.19} & 96.38 & 97.04 \increase{0.66} & 93.09 & 91.94 \decrease{1.15} & 97.04 & 97.14 \increase{0.10} \\ \hline
        SVAMP & 93.70 & 95.50 \increase{1.80} & 93.00 & 95.10 \increase{2.10} & 85.90 & 85.40 \decrease{0.50} & 94.00 & 93.30 \decrease{0.70} \\ \hline
        AQUA & 87.40 & 88.98 \increase{1.58} & 90.94 & 87.01 \decrease{3.93} & 71.65 & 85.43 \increase{13.78} & 86.61 & 85.83 \decrease{0.78} \\ \hline
        R-GSM & 90.91 & 91.82 \increase{0.91} & 89.09 & 89.09 & 70.37 & 65.60 \decrease{4.77} & 86.36 & 86.82 \increase{0.46} \\ \hline
        StrategyQA & 85.83 & 88.43 \increase{3.10} & 69.30 & 84.37 \increase{15.07} & 66.50 & 62.00 \decrease{4.50} & 69.21 & 63.49 \decrease{5.72} \\ \hline
        Date & 95.54 & 97.49 \increase{1.95} & 91.36 & 91.92 \increase{0.56} & 78.27 & 68.25 \decrease{10.02} & 83.84 & 84.68 \increase{0.84} \\ \hline
        Five Objects & 95.60 & 97.20 \increase{1.60} & 92.80 & 94.00 \increase{1.20} & 66.80 & 67.20 \increase{0.40} & 86.00 & 90.00 \increase{4.00} \\ \hline
        Seven Objects & 89.60 & 90.00 \increase{0.40} & 79.60 & 83.60 \increase{4.00} & 52.00 & 58.80 \increase{6.80} & 83.20 & 85.60 \increase{2.40} \\ \hline
        Color Objects & 96.80 & 97.20 \increase{0.40} & 92.00 & 93.60 \increase{1.60} & 78.80 & 79.20 \increase{0.40} & 97.60 & 97.20 \increase{0.40} \\ \hline
        Judgement & 67.91 & 74.33 \increase{6.42} & 67.91 & 71.12 \increase{3.21} & 58.29 & 58.82 \increase{0.53} & 71.12 & 70.59 \decrease{0.53} \\ \hline
        Navigate & 95.20 & 97.20 \increase{2.00} & 87.60 & 90.40 \increase{2.80} & 74.40 & 71.60 \decrease{2.80} & 89.20 & 87.60 \decrease{1.60} \\ \hline
        Mean $\Delta$ & \multicolumn{2}{c}{\hspace{10mm}\increasenoparentwithspace{1.59}} & \multicolumn{2}{c}{\hspace{10mm}\increasenoparentwithspace{2.10}} & \multicolumn{2}{c}{\hspace{10mm}\decreasenoparentwithspace{0.22}} & \multicolumn{2}{c}{\hspace{10mm}\decreasenoparentwithspace{0.11}} \\
        \hline
    \end{tabular}
    }
\label{tab:small_models_performance_all}
\end{table}

\begin{table*}[h!]
\caption{In (a), \llamaA often tags calculated results within the reasoning chains, which violates our tagging strategy. According to our approach, calculated results are considered new information that cannot be linked back to the question and, therefore, should not be tagged. Additionally, in (b) and (c), \llamaA exhibits a tendency to replicate few-shot examples. Specifically, \textcolor{ForestGreen}{5.9\%} of test examples in AQUA and \textcolor{ForestGreen}{16.41\%} of test examples in StrategyQA contain repeated few-shot examples.}
\begin{tabular}{p{\textwidth}} 
\hline
(a) With \hot: \llamaA tends to tag calculated results in GSM8K Examples \\ \hline
\textbf{Question}: If Marcy works for the same company for \sethlcolor{fact1}\hl{40 years}, she gets an annual pension of \sethlcolor{fact2}\hl{\$50,000/year}. Starting after \sethlcolor{fact3}\hl{20 years}, she becomes entitled to \sethlcolor{fact4}\hl{5\% of the value of the pension per year}. If she quits after \sethlcolor{fact5}\hl{30 years}, what will her annual pension be?
 \\
\noindent\textbf{Answer:} Marcy gets an annual pension of \sethlcolor{fact2}\hl{\$50,000/year} for the first \sethlcolor{fact6}\hl{20 years} she works for the company. Starting after \sethlcolor{fact3}\hl{20 years}, she becomes entitled to \sethlcolor{fact4}\hl{5\% of the value of the pension per year}. So, after \sethlcolor{fact3}\hl{20 years}, she gets \sethlcolor{fact2}\hl{\$50,000/year} + \sethlcolor{fact2}\hl{\$50,000/year} * \sethlcolor{fact4}\hl{5\%} = \sethlcolor{fact2}\hl{\$50,000/year} + \sethlcolor{fact2}\hl{\$50,000/year} * \sethlcolor{fact7}\hl{0.05} = \sethlcolor{fact2}\hl{\$50,000/year} + \sethlcolor{fact8}\hl{\$2,500/year} = \sethlcolor{fact9}\hl{\$52,500/year} for the next \sethlcolor{fact10}\hl{10 years} she works for the company. Since she quits after \sethlcolor{fact5}\hl{30 years}, she gets \sethlcolor{fact9}\hl{\$52,500/year} for the last \sethlcolor{fact11}\hl{10 years} she works for the company. Therefore, her annual pension for the last \sethlcolor{fact5}\hl{10 years} she works for the company will be \sethlcolor{fact9}\hl{\$52,500/year}. {\color{red}{\ding{55}}} \textbf{(GT: \$25,000)} \\ \hline
(b) With \hot: \llamaA tends to repeat the few-shot example in StrategyQA \\ \hline
\textbf{Question}: Are all limbs required for jujutsu? \\
\textbf{Answer}: 

Reformatted Question: At a presentation about \sethlcolor{fact1}\hl{post traumatic stress disorder}, would \sethlcolor{fact2}\hl{Ariana Grande} be a topic of relevance?
Answer: \{true\}

Reformatted Question: Has the \sethlcolor{fact1}\hl{Indian Ocean garbage patch} \sethlcolor{fact2}\hl{not completed} \sethlcolor{fact3}\hl{two full rotations} of debris since its discovery?
Answer: \{true\}

\textcolor{red!80}{[more repeated few-shot examples]}

Reformatted Question: Is a \sethlcolor{fact1}\hl{900,000 pound net worth} person an \sethlcolor{fact2}\hl{American billionaire} if they \sethlcolor{fact3}\hl{exchange currency June 2020}? Answer: \{true\}. {\color{red}{\ding{55}}} (\textbf{GT: False}) \\ \hline
(c) With \hot: \llamaA tends to repeat the few-shot example in AQUA \\ \hline
\textbf{Question}: Jerry purchased a 1-year \$5,000 bond that paid an annual interest rate of 12\% compounded every six months. How much interest had this bond accrued at maturity? Answer option: ['A)\$5102', 'B)\$618', 'C)\$216', 'D)\$202', 'E)\$200'] \\
\textbf{Answer}: 

Reformatted Question: A person is traveling at \sethlcolor{fact1}\hl{20 km/hr} and reached his destination in \sethlcolor{fact2}\hl{2.5 hr} then find the distance? Answer Choices: (a) 53 km (b) 55 km (c) 52 km (d) 60 km (e) 50 km

Answer: The distance that the person traveled would have been \sethlcolor{fact1}\hl{20 km/hr} * \sethlcolor{fact2}\hl{2.5 hrs} = 50 km. So the answer is \{e\}.

Reformatted Question: John found that the \sethlcolor{fact1}\hl{average of 15 numbers is 40}. If \sethlcolor{fact2}\hl{10 is added to each number}, then the \sethlcolor{fact3}\hl{mean of the numbers} is? Answer Choices: (a) 50 (b) 45 (c) 65 (d) 78 (e) 64

Answer: If \sethlcolor{fact2}\hl{10 is added to each number}, then the \sethlcolor{fact3}\hl{mean of the numbers} also increases by 10. So the new mean would be \sethlcolor{fact1}\hl{40} + 10 = 50. So the answer is \{a\}.

\textcolor{red!80}{[more repeated few-shot examples]}

Reformatted Question: Two friends plan to walk along a \sethlcolor{fact1}\hl{43-km trail}, starting at opposite ends of the trail at the same time. If \sethlcolor{fact2}\hl{Friend P's rate is 15\% faster than Friend Q's}, how many kilometers will Friend P have walked when they pass each other? Answer Choices: (a) 21 (b) 21.5 (c) 22 (d) 22.5 (e) 23

Answer: If \sethlcolor{fact2}\hl{Q completes x kilometers}, then \sethlcolor{fact2}\hl{then P completes 1.15x kilometers}. x + \sethlcolor{fact2}\hl{1.15}x = \sethlcolor{fact1}\hl{43} 2.15x=\sethlcolor{fact1}\hl{43} x = \sethlcolor{fact1}\hl{43}/2.15 = 20 Then \sethlcolor{fact2}\hl{P will have walked 1.15*20=23 km}. So the answer is \{e\}.
 {\color{red}{\ding{55}}} (\textbf{GT: B})\\ \hline

\end{tabular}
\label{tab:llama8b_repeat_fs_cases}
\end{table*}

\clearpage
\section{Detail Analysis of the Impact of Tagging and Question Repetition on Chain-of-Thought Performance}
\label{sec:detail_prompt_variation_app}
\cref{tab:prompt_variation_acc} presents the detailed analysis of \cref{sec:prompt_variation_exp}, which shows the performance of prompt variations: (1) CoT, (2) CoT + Repeat Question, (3) Tag in Question, (4) Tag in Answer, (5) Tag in Question and Answer. Moreover, we also show the detailed number of tags across 6 datasets generated in \textbf{Reformatted Question} and \textbf{Answer}. Specifically, \cref{tab:tag_counts_all_strategies} displays the number of tags when prompting LLMs to tag in Question only, the number of tags used when prompting LLMs to apply tags solely to answers, and the number of tags generated when LLMs tag both questions and answers, respectively.

Detailed information regarding above-mentioned prompts is shown in \cref{tab:prompts_variation}.

\begin{table*}[ht]
\centering
\caption{Performance comparison of various few-shot prompting strategies (CoT, Repeated Question, Tags in Question, Tags in Answer, and Tag in QA) across multiple models (\geminiflash, \geminipro, \llamaB, and \llamaC) and datasets (tested each dataset on 400 random samples). The table shows prompt \textbf{Tags in QA} achieves the highest mean accuracy among multiple prompts.}
\resizebox{\textwidth}{!}{%
\begin{tabular}{llllllll|l}
\hline
Model & Prompt & AQUA & StrategyQA & R-GSM & \makecell{Seven\\Objects} & \makecell{Causal\\Judgement} & Navigate & Mean \\ \hline
 & CoT & 72.50  & 75.25 & 84.50  & 72.50  & 71.66 & \textbf{92.50} & 78.15\\
& Repeated Question &  97.00 & 80.75 & 81.50  & 74.50  & 89.84 & 73.50  & 82.85 \\
\multirow{2}{*}{
  \raisebox{-0.3\height}{
    \hspace*{-0.45em}
    \geminiflashlogo 
  }
} & Tags in Question & 97.00    & 79.75 & 82.50  & 74.50  & \textbf{72.19} & 92.00 & 82.99  \\
& Tags in Answer   &  95.00    & \textbf{81.50}  & 83.50  & 70.50  & 68.98 & \textbf{92.50} & 82.00\\
& Tags in QA & \textbf{96.00}    & 79.50  & \textbf{86.00}    & \textbf{78.50}  & 71.66 & 91.00  & \textbf{83.78} \\ \hline
& normal CoT & 83.00    & 81.00    & 77.00    & 83.00    & 74.87 & 91.00  & 81.65 \\
& Repeated Question & 70.00    & \textbf{83.50}  & 89.00    & 87.50  & 72.19 & 93.50 & 82.62 \\
\multirow{2}{*}{
  \raisebox{-0.3\height}{
    \hspace*{0.03em}
    \geminilogo
  }
} & Tags in Question & 71.00    & 82.00    & 89.50  & 86.00    & 72.19 & 95.50 & 82.70 \\
& Tags in Answer & 72.00 & \textbf{83.50}  & 88.50  & 87.00    & 71.66 & \textbf{97.00}  & 83.28 \\
& Tags in Both  & \textbf{89.50}  & 82.25 & \textbf{90.50}  & \textbf{88.00}    & \textbf{75.40}  & 96.00 & \textbf{86.94} \\ \hline
& CoT   & \textbf{90.50}  & 76.00 & 88.50  & 78.50  & 67.91 & 85.00 & 81.07  \\
& Repeated Question &  84.00    & 78.00 & \textbf{90.00} & 83.00 & 68.98 & 87.50 & 81.91 \\
\multirow{2}{*}{
  \raisebox{-0.3\height}{
    \hspace*{0.03em}
    \llamaMedlogo
  }
} & Tags in Question & 79.00    & 82.25 & 84.50  & 83.00 & \textbf{74.33} & 86.00  & 81.51 \\
& Tags in Answer   & 82.50  & 82.00 & 86.50  & \textbf{85.00}    & 70.59 & \textbf{90.00} &  82.77 \\
& Tags in Both & 86.00    & \textbf{87.00}    & 88.00 & 81.50  & 71.12 & 88.50 & \textbf{83.69} \\ \hline
& CoT & 86.00 & 83.25 & 90.50  & 88.50  & 67.91 & 94.50 & 85.11 \\
& Repeated Question & 84.00 & 86.25 & 90.00 & 87.50  & 70.05 & 94.50 & 85.38 \\
\multirow{2}{*}{
  \raisebox{-0.3\height}{
    \hspace*{0.03em}
    \llamaBiglogo
  }
} & Tags in Question & 81.00 & 86.25 & 92.00 & 87.50  & 71.12 & 96.00 & 85.65 \\
& Tags in Answer  & 79.00 & 85.50 & 90.00  & 89.50  & 72.73 & 94.00 & 85.12 \\
& Tags in Both  & 88.00 & \textbf{87.50}  & \textbf{91.50} & \textbf{89.00}    & \textbf{74.33}  & \textbf{96.50}& \textbf{87.81}   \\ 
\hline
\end{tabular}%
}
\label{tab:prompt_variation_acc}
\end{table*}

\begin{table}[h]
\centering
\begin{minipage}{0.5\textwidth}
\centering
\resizebox{\columnwidth}{!}{%
\begin{tabular}{lrrrr|r|r}
\toprule
Method & \geminiflashlogo & \geminilogo & \llamaMedlogo & \llamaBiglogo & Model mean & \human \\
\midrule
\rowcolor{gray!5}
\multicolumn{7}{c}{(a) Mean number of tags in \textbf{Question}} \\
\tqcolored & 9.28 & 11.41 & 9.04 & 9.92 & 9.91 & 8.99 \\
\tacolored  & 0.02 & 3.27 & 0.35 & 2.27 & 1.48 & 0.00 \\
\hotcolored & 9.04 & 10.73 & 9.35 & 10.05 & \textbf{9.79} & 8.99 \\
\midrule
\rowcolor{gray!5}
\multicolumn{7}{c}{(b) Mean number of tags in \textbf{Answer}} \\
\tqcolored & 0.00 & 0.00 & 0.06 & 0.00 & 0.02& 0.00 \\
\tacolored & 13.54 & 13.23 & 14.48 & 12.02 & 13.32 & 10.18\\
\hotcolored & 10.24 & 13.15 & 12.14 & 12.28 & \textbf{11.95}& 10.18\\
\bottomrule
\end{tabular}
}
\end{minipage}%
\hfill
\begin{minipage}{0.48\textwidth}
\caption{The mean number of tags generated by LLMs for \hotcolored match that by \human few-show demonstrations the most compared to other variations (\tqcolored and \tacolored), showing the importance of instructing LLMs to insert tags to both questions and answers (\ie, \hotcolored).
All numbers are means computed across 6 datasets in the ablation study.
}
\label{tab:avg_tags}
\end{minipage}
\end{table}

\begin{table}[ht]
\centering
\caption{Average number of tags in Questions and Answers across different prompting strategies}
\resizebox{0.8\columnwidth}{!}{%
\begin{tabular}{llcccc|cccc}
\hline
\multirow{2}{*}{\textbf{Prompt Strategy}} & \multirow{2}{*}{\textbf{Dataset}} & \multicolumn{4}{c|}{\textbf{Tags in question}} & \multicolumn{4}{c}{\textbf{Tags in answer}} \\
\cline{3-10}
& & \geminiflashlogo & \geminilogo & \llamaMedlogo & \llamaBiglogo & \geminiflashlogo & \geminilogo & \llamaMedlogo & \llamaBiglogo \\
\hline
\multirow{7}{*}{\textbf{Tag in Question}} 
& AQUA & 6.34 & 8.36 & 7.17 & 7.98 & 0.22 & 0.21 & 0.00 & 0.00 \\
\cline{2-10}
& StrategyQA & 5.28 & 5.38 & 4.84 & 5.17 & 0.00 & 0.00 & 0.00 & 0.08 \\
\cline{2-10}
& R-GSM & 10.24 & 10.18 & 9.53 & 9.38 & 0.00 & 0.00 & 0.00 & 0.00 \\
\cline{2-10}
& 7 Objects & 12.15 & 13.57 & 12.79 & 12.03 & 2.74 & 0.12 & 0.00 & 0.00 \\
\cline{2-10}
& Judgement & 10.64 & 19.96 & 8.96 & 13.94 & 0.01 & 0.04 & 0.09 & 0.03 \\
\cline{2-10}
& Navigate & 11.00 & 11.00 & 10.97 & 11.01 & 0.00 & 0.00 & 0.06 & 0.00 \\
\cline{2-10}
\rowcolor{gray!10}& Mean & 9.28 & 11.41 & 9.04 & 9.92 & 0.05 & 0.06 & 0.03 & 0.02 \\
\hline
\multirow{7}{*}{\textbf{Tag in Answer}}
& AQUA & 0.12 & 5.35 & 0.08 & 4.88 & 13.74 & 17.55 & 23.74 & 15.57 \\
\cline{2-10}
& StrategyQA & 0.00 & 0.02 & 0.02 & 0.27 & 12.15 & 11.04 & 12.77 & 9.31 \\
\cline{2-10}
& R-GSM & 0.00 & 6.29 & 1.97 & 5.48 & 20.99 & 16.74 & 17.75 & 14.17 \\
\cline{2-10}
& 7 Objects & 0.00 & 2.23 & 0.00 & 0.06 & 12.96 & 11.12 & 12.54 & 14.32 \\
\cline{2-10}
& Judgement & 0.00 & 5.70 & 0.00 & 2.90 & 9.59 & 11.72 & 9.43 & 7.74 \\
\cline{2-10}
& Navigate & 0.00 & 0.00 & 0.03 & 0.00 & 11.78 & 11.20 & 10.63 & 11.02 \\
\cline{2-10}
\rowcolor{gray!10}& Mean & 0.02 & 3.27 & 0.35 & 2.27 & 13.54 & 13.23 & 14.48 & 12.02 \\
\hline
\multirow{7}{*}{\textbf{Tag in Both}}
& AQUA & 6.21 & 7.32 & 7.31 & 7.54 & 13.95 & 14.69 & 16.89 & 17.53 \\
\cline{2-10}
& StrategyQA & 3.40 & 4.63 & 4.34 & 4.55 & 6.96 & 8.40 & 10.91 & 9.37 \\
\cline{2-10}
& R-GSM & 9.83 & 6.70 & 9.81 & 9.35 & 9.96 & 22.20 & 15.70 & 13.25 \\
\cline{2-10}
& 7 Objects & 11.64 & 13.74 & 12.07 & 12.29 & 12.74 & 9.01 & 11.20 & 13.95 \\
\cline{2-10}
& Judgement & 12.95 & 21.75 & 12.22 & 16.13 & 7.50 & 14.34 & 7.88 & 9.17 \\
\cline{2-10}
& Navigate & 10.23 & 10.23 & 10.34 & 10.42 & 10.30 & 10.28 & 10.24 & 10.42 \\
\cline{2-10}
\rowcolor{gray!10} & Mean & 9.04 & 10.73 & 9.35 & 10.05 & 10.24 & 13.15 & 12.14 & 12.28 \\
\hline
\end{tabular}
}
\label{tab:tag_counts_all_strategies}
\end{table}

\begin{table*}[h!]
\caption{Specific prompts of Standard CoT, CoT+Repeating Question (\citep{mekala-etal-2024-echoprompt}, CoT + Tags in Question, CoT + Tags in Answer, CoT + Tags in Question and Answer (\hot).}
\centering
\resizebox{0.8\columnwidth}{!}{%
\begin{tabular}{p{2.5cm}p{12cm}}
\hline
     Methods & Prompt Content \\ \hline
     Standard CoT & Question: \{question\} \newline Instruction: "Please generate your explanation first, then generate the final answer in the bracket as follows:
     
     Answer:"\\ \hline
     CoT + Repeating Question \citep{mekala-etal-2024-echoprompt} & Question: \{question\} \newline Instruction: "I want you to answer this question. To do that, first, repeat the question and then, generate your answers. The output format is as follows:
     
    Reformatted Question: 
    
    Answer:"\\ \hline
     CoT + Tags in Question & Question: \{question\} \newline Instruction: "I want you to answer this question. To do that, first, re-generate the question with proper tags for key phrases, the key phrases that are most relevant to answering the question \{last\_sentence\}, and then generate your answers. The output format is as follows:
     
    Reformatted Question: 
    
    Answer:"\\ \hline
     CoT + Tags in Answer & Question: \{question\} \newline Instruction: "I want you to answer this question but your explanation should contain references referring back to the information in the question. To do that, first, repeat the question and then, generate your answers with proper tags for key phrases, the key phrases that are most relevant to answering the question \{last\_sentence\}. The output format is as follows:
    Reformatted Question:
                
    Answer:"\\ \hline
     CoT + Tags in Question \& Answer (\hot) & Question: \{question\} \newline Instruction: "I want you to answer this question but your explanation should contain references referring back to the information in the question. To do that, first, re-generate the question with proper tags for key phrases, the key phrases that are most relevant to answering the question \texttt{\{last\_sentence\}} and then generate your answers. The output format is as follows:

     Reformatted Question:

     Answer: "\\ \hline
\end{tabular}
}
\label{tab:prompts_variation}
\end{table*}

\clearpage
\section{Highlighted Fewshot Chain of Thought Annotation}
\label{sec:annotation_app}
To construct \hot few-shot examples for the evaluated benchmarks, questions from CoT few-shot examples are annotated using GPT-4o with the prompt shown in \cref{tab:tag_in_question_algo}. Subsequently, answers from CoT few-shot examples are tagged using the prompt detailed in \cref{tab:tag_in_answer_algo}.

Furthermore, we provide an example of few-shot examples of arithmetic tasks (e.g. GSM8K) in \cref{tab:full_fewshot_prompt_example}.

\begin{table*}[ht]
\centering
\caption{Examples of questions before and after applying key phrase tags using \texttt{<fact>} tags. The tags highlight short, concise, and critical information in the questions that cannot be removed without altering the meaning of the questions.}
\begin{tabular}{p{\textwidth}} 
\hline
Prompt to tag in the Question \\ \hline
Read the question and insert the tags into the question via the following rules:\\
1. Insert only tags keeping the original words unchanged.\\
2. Put the tags (e.g., <fact1></fact1>, <fact2></fact2>) around the shortest and most concise important phrases.\\
3. A phrase is considered important and should be tagged if replacing that phrase by a closest alternative phrase would change the answer. \\
4. Do not tag phrases non-important to answering the question.\\
Re-generate the question after adding tags to the phrases. \\
\# EXAMPLES\\
Below are examples of questions before and after key phrases are tagged using <fact> tags.
If one key phrase was absent, it would be impossible for one to answer the question correctly.

\#\# Question 1: \\
\#\#\# BEFORE: 
Sam works at the Widget Factory, assembling Widgets. He can assemble 1 widget every 10 minutes. Jack from the loading dock can help assemble widgets when he doesn't have anything else to do. When he helps, they put together 2 complete widgets every 15 minutes. Recently the factory hired Tony to help assemble widgets. Being new to the job, he doesn't work as fast as Sam or Jack. Yesterday Sam worked for 6 hours before he had to leave work early for a dentist appointment. Jack was able to help out for 4 hours before he had to go back to the loading dock to unload a new shipment of widget materials. Tony worked the entire 8-hour shift. At the end of the day, they had completed 68 widgets. How long does it take Tony to assemble a Widget, in minutes?

\#\#\# AFTER:
Sam works at the Widget Factory, assembling Widgets. He can assemble <fact1>1 widget every 10 minutes</fact1>. Jack from the loading dock can help assemble widgets when he doesn't have anything else to do. When he helps, they put together <fact2>2 complete widgets every 15 minutes</fact2>. Recently the factory hired Tony to help assemble widgets. Being new to the job, he doesn't work as fast as Sam or Jack. Yesterday Sam worked for <fact3>6 hours</fact3> before he had to leave work early for a dentist appointment. Jack was able to help out for <fact4>4 hours</fact4> before he had to go back to the loading dock to unload a new shipment of widget materials. Tony worked the entire <fact5>8-hour shift</fact5>. At the end of the day, they had completed <fact6>68 widgets</fact6>. How long does it take Tony to assemble a Widget, in minutes?
\\

\textbf{[more examplars]}\\

\#\# Question 15: \\
\#\#\# BEFORE: \texttt{\{question to be tagged\}}\\
\#\#\# AFTER:
\end{tabular}
\label{tab:tag_in_question_algo}
\end{table*}

\begin{table*}[ht]
\centering
\caption{Examples of tagged questions, answers (without tags), and answers (with tags). The tags in the answers correspond to the tags in the questions.}
\begin{tabular}{p{\textwidth}} 
\hline
Prompt to tag in the Answer \\ \hline
Given a pair of (Tagged Question, Answer) I want to generate Tagged Answer. That is, to generate Tagged Answer, take the Answer and add tags to the key phrases that refer to the corresponding tagged phrases (e.g., <fact1>...</fact1>) from the Tagged Question.\\
For example, the phrases tagged with the same tag (e.g. <fact1>) across Tagged Question and Tagged Answer should be synonymous or refer to the same entity.\\
Please re-generate the answer with tags.\\
Provide your tagged answer.\\
\#\# Question 1: \\
\#\#\# TAGGED QUESTION: \\
Sam works at the Widget Factory, assembling Widgets. He can assemble <fact1>1 widget every 10 minutes</fact1>. Jack from the loading dock can help assemble widgets when he doesn't have anything else to do. When he helps, they put together <fact2>2 complete widgets every 15 minutes</fact2>. Recently the factory hired Tony to help assemble widgets. Being new to the job, he doesn't work as fast as Sam or Jack. Yesterday Sam worked for <fact3>6 hours</fact3> before he had to leave work early for a dentist appointment. Jack was able to help out for <fact4>4 hours</fact4> before he had to go back to the loading dock to unload a new shipment of widget materials. Tony worked the entire <fact5>8-hour shift</fact5>. At the end of the day, they had completed <fact6>68 widgets</fact6>. How long does it take Tony to assemble a Widget, in minutes?

\#\#\# ANSWER:\\
Sam completes a widget every 10 minutes. When Jack helps, they finish 2 in 15 minutes. Sam has finished 1 widget and has begun working on another one, and Jack finishes the second one at 15 minutes. So it takes Jack 15 minutes to complete a widget.
Sam worked for 6 hours yesterday, so he was able to complete 6 hours * 60 minutes per hour / 10 minutes per widget = 36 widgets.
Jack worked for 4 hours, so he was able to complete 4 hours * 60 minutes per hour / 15 minutes per widget = 16 widgets.
Sam, Jack, and Tony were able to complete 68 widgets together. So of those, Tony personally completed 68 widgets - 36 widgets - 16 widgets = 16 widgets.
It took Tony 8 hours to complete those 16 widgets, so he takes 8 hours * 60 minutes per hour / 16 widgets = 8*60/16=30 minutes per widget. The answer is \{30\}.

\#\#\# TAGGED ANSWER:\\
Sam completes <fact1>a widget every 10 minutes</fact1>. When Jack helps, they finish <fact2>2 in 15 minutes</fact2>. Sam has finished 1 widget and has begun working on another one, and Jack finishes the second one at 15 minutes. So it takes Jack 15 minutes to complete a widget.
Sam worked for <fact3>6 hours</fact3> yesterday, so he was able to complete <fact3>6 hours</fact3> * 60 minutes per hour / <fact1>10 minutes per widget</fact1> = 36 widgets.
Jack worked for <fact4>4 hours</fact4>, so he was able to complete <fact4>4 hours</fact4> * 60 minutes per hour / <fact2>15 minutes per widget</fact2> = 16 widgets.
Sam, Jack, and Tony were able to complete <fact6>68 widgets</fact6> together. So of those, Tony personally completed <fact6>68 widgets</fact6> - 36 widgets - 16 widgets = 16 widgets.
It took Tony <fact5>8 hours</fact5> to complete those 16 widgets, so he takes <fact5>8 hours</fact5> * 60 minutes per hour / 16 widgets = <fact5>8</fact5>*60/16=30 minutes per widget. The answer is \{30\}.\\
\textbf{[more examplars]}\\
\#\# Question 15:

\end{tabular}
\label{tab:tag_in_answer_algo}
\end{table*}

\clearpage
\begin{longtable}{!{} p{\dimexpr\textwidth-6pt\relax}}
\caption{An example of fewshot HoT prompt in GSM8K. The table shows the entire components of the HoT prompt, including 8 few-shot examples, a query question, and an instruction.}
\label{tab:full_fewshot_prompt_example}\\
\hline
\rowcolor{gray!30}Fewshot HoT prompt of GSM8K\\
\hline
\endfirsthead

\multicolumn{1}{c}{\tablename\ \thetable\ -- continued from previous page} \\
\hline
\rowcolor{gray!30}Fewshot HoT prompt of GSM8K (continued)\\
\hline
\endhead

\hline \multicolumn{1}{r}{Continued on next page} \\
\endfoot

\hline
\endlastfoot
\textbf{Question}: For every 12 cans you recycle, you receive \$0.50, and for every 5 kilograms of newspapers, you receive \$1.50. If your family collected 144 cans and 20 kilograms of newspapers, how much money would you receive?\\
\textbf{Reformatted Question}: For \xmlfactopen{1}every 12 cans\xmlfactclose{1} you recycle, you receive \xmlfactopen{2}\$0.50\xmlfactclose{2}, and for \xmlfactopen{3}every 5 kilograms of newspapers\xmlfactclose{3}, you receive \xmlfactopen{4}\$1.50\xmlfactclose{4}. If your family collected \xmlfactopen{5}144 cans\xmlfactclose{5} and \xmlfactopen{6}20 kilograms of newspapers\xmlfactclose{6}, how much money would you receive?\\
\textbf{Answer}: There are \xmlfactopen{5}144\xmlfactclose{5}/\xmlfactopen{1}12\xmlfactclose{1} = 12 sets of \xmlfactopen{1}12 cans\xmlfactclose{1} that the family collected. So, the family would receive \xmlfactopen{2}\$0.50\xmlfactclose{2} x 12 = \$6 for the cans. There are \xmlfactopen{6}20\xmlfactclose{6}/\xmlfactopen{3}5\xmlfactclose{3} = 4 sets of \xmlfactopen{3}5 kilograms of newspapers\xmlfactclose{3} that the family collected. So, the family would receive \xmlfactopen{4}\$1.50\xmlfactclose{4} x 4 = \$6 for the newspapers. Therefore, the family would receive a total of \$6 + \$6 = \$12. The answer is \{12\}.\\\\
\textbf{Question}: Ellie went to visit a circus with Sarah and they both got lost in the house of mirrors. They have to travel through the house of mirrors a few times before they finally get out and when they leave, they discuss how many times they've seen their own reflections. Sarah says that every time they were in the room with tall mirrors, she saw her reflection 10 times and every time they were in the room with wide mirrors, she saw her reflection 5 times. Ellie says that every time they were in the room with tall mirrors, she saw her reflection 6 times and every time they were in the room with wide mirrors she saw her reflection 3 times. They both passed through the room with tall mirrors 3 times each and they both passed through the room with wide mirrors 5 times each. In total, how many times did Sarah and Ellie see their reflections?\\
\textbf{Reformatted Question}: Ellie went to visit a circus with Sarah, and they both got lost in the house of mirrors. They have to travel through the house of mirrors a few times before they finally get out, and when they leave, they discuss how many times they've seen their own reflections. \xmlfactopen{1}Sarah says that every time they were in the room with tall mirrors\xmlfactclose{1}, she saw her reflection \xmlfactopen{2}10 times\xmlfactclose{2}, and \xmlfactopen{3}every time they were in the room with wide mirrors\xmlfactclose{3}, she saw her reflection \xmlfactopen{4}5 times\xmlfactclose{4}. \xmlfactopen{5}Ellie says that every time they were in the room with tall mirrors\xmlfactclose{5}, she saw her reflection \xmlfactopen{6}6 times\xmlfactclose{6}, and \xmlfactopen{7}every time they were in the room with wide mirrors\xmlfactclose{7}, she saw her reflection \xmlfactopen{8}3 times\xmlfactclose{8}. They both \xmlfactopen{9}passed through the room with tall mirrors 3 times\xmlfactclose{9} each, and they both \xmlfactopen{10}passed through the room with wide mirrors 5 times\xmlfactclose{10} each. In total, how many times did Sarah and Ellie see their reflections?\\
\textbf{Answer}: In the rooms with tall mirrors, Sarah saw her reflection a total of \xmlfactopen{2}10 reflections\xmlfactclose{2} * \xmlfactopen{9}3 passes\xmlfactclose{9} = 30 times. In the rooms with wide mirrors, Sarah saw her reflection a total of \xmlfactopen{4}5 reflections\xmlfactclose{4} * \xmlfactopen{10}5 passes\xmlfactclose{10} = 25 reflections. So Sarah saw her reflection a total of 30 + 25 = 55 times. In the rooms with tall mirrors, Ellie saw her reflection a total of \xmlfactopen{6}6 reflections\xmlfactclose{6} * \xmlfactopen{9}3 passes\xmlfactclose{9} = 18 times. In the rooms with wide mirrors, Ellie saw her reflection a total of \xmlfactopen{8}3 reflections\xmlfactclose{8} * \xmlfactopen{10}5 passes\xmlfactclose{10} = 15 times. So Ellie saw her reflection a total of 18 + 15 = 33 times. Therefore, Sarah and Ellie saw their reflections a total of 55 + 33 = 88 times. The answer is \{88\}.
{more examplars}
\\ 
\noindent\rule{\textwidth}{0.5pt}

\textbf{Question}: Henry and 3 of his friends order 7 pizzas for lunch. Each pizza is cut into 8 slices. If Henry and his friends want to share the pizzas equally, how many slices can each of them have? 

\noindent\rule{\textwidth}{0.5pt}

\textbf{Instruction:} I want you to answer this question but your explanation should contain references referring back to the information in the question. To do that, first, re-generate the question with proper tags for key phrases, the key phrases that are most relevant to answering the question how many slices can each of them have? and then generate your answers. The output format is as follows:

     \textbf{Reformatted Question:}

     \textbf{Answer:}\\

\end{longtable}

\clearpage
\section{Some Highlight Cases}
\label{sec:grounding_examples_app}
Furthermore, some failure cases and success cases of \hot are presented in \cref{tab:llama70b_fail_cases}--\cref{tab:gcot_vs_cot_math500_conditioned_math}.

\begin{table*}[h!]
\caption{\llamaB Fail Cases.}
\begin{tabular}{p{\textwidth}} 
\hline
DROP \hot \textcolor{red}{Fail Cases (LLM fails to repeat 231 women and children, instead it generates 201 women and children)} \\ \hline
\textbf{Question}: In 1905, 1,003 Korean immigrants, which included 802 men and 231 women and children, departed from the port of Chemulpo, Incheon aboard the ship Ilford to Salina Cruz, Oaxaca, Mexico....How many Korean immigrants were more men than woman and children? \\
\noindent\textbf{Reformatted Question:} In 1905, \sethlcolor{fact1}\hl{1,003 Korean immigrants}, which included \sethlcolor{fact2}\hl{802 men} and \sethlcolor{fact3}\hl{201 women and children}, departed from the port of Chemulpo, Incheon aboard the ship Ilford to Salina Cruz, Oaxaca, Mexico...How many Korean immigrants were more men than woman and children?
\\
\noindent\textbf{Answer:} According to the passage, \sethlcolor{fact2}\hl{802 men} and \sethlcolor{fact3}\hl{201 women and children} were among the \sethlcolor{fact1}\hl{1,003 Korean immigrants}. To find the difference between the number of men and women and children, subtract the number of women and children from the number of men: \sethlcolor{fact2}\hl{802} - \sethlcolor{fact3}\hl{201} = 601. So the answer is {601}.
. {\color{red}{\ding{55}}} \textbf{(GT: 571)} \\ \hline
\end{tabular}
\label{tab:llama70b_fail_cases}
\end{table*}

\begin{table*}[h!]
\caption{\geminiflashlogo \hot Correct Vs. CoT Incorrect Answers. StrategyQA.}
\begin{tabular}{p{\textwidth}} 
\hline
StrategyQA \\ \hline
\textbf{Question}: Could a llama birth twice during War in Vietnam (1945-46)?  \\
\textbf{\hot Correct Response}: \\
Reformatted Question:  Could a \sethlcolor{fact1}\hl{llama} \sethlcolor{fact2}\hl{birth twice} during \sethlcolor{fact3}\hl{War in Vietnam (1945-46)}? \\
Answer: a \sethlcolor{fact1}\hl{llama} can only give birth at most once per year… if we consider the dates of \sethlcolor{fact3}\hl{1945-1946} provided in the question, that is only a 1 year span. Therefore, a \sethlcolor{fact1}\hl{llama} could not \sethlcolor{fact2}\hl{give birth twice} in that timeframe. So the answer is $\boxed{false}$ {\color{green}{\ding{51}}} (\textbf{GT: None})\\
\textbf{CoT Incorrect Response}: \\
 Answer: Llamas have a gestation period of approximately 11 months. The Vietnam War lasted from 1955 to 1975, much longer than two llama gestation periods. Therefore, a llama could give birth twice during the Vietnam War. So the answer is {true}. {\color{red}{\ding{55}}} (\textbf{GT: None}) \\ \hline

\end{tabular}
\label{tab:gcot_vs_cot_strategyqa}
\end{table*}

\begin{table*}[h!]
\caption{\geminiflashlogo \hot Correct Vs. CoT Incorrect Answers. PuzzleTrivial.}
\begin{tabular}{p{\textwidth}} 
\hline
PuzzleTrivial \\ \hline
\textbf{Question}: There are 12 coins, and one is rumored to be heavier. You have a perfect balance scale to determine which coin it is. However, all coins are actually identical. Which coin is heavier? \\
\textbf{\hot Correct Response}: \\
Reformatted Question: There are \sethlcolor{fact1}\hl{12 coins}, and \sethlcolor{fact2}\hl{one is rumored to be heavier}. You have a perfect balance scale to determine which coin it is. However, all \sethlcolor{fact3}\hl{coins are actually identical}. Which coin is heavier?\\
Answer: Since \sethlcolor{fact3}\hl{all the coins are identical}, and none of them are heavier according to the rumor, none of the \sethlcolor{fact1}\hl{12 coins} are heavier than the rest. The answer is \{none\}. {\color{green}{\ding{51}}} (\textbf{GT: None})\\
\textbf{CoT Incorrect Response}: \\
 Answer: To determine which of the 12 coins is the heavier one using only two weighings on a balance scale, we can follow these steps:

Step-by-Step Solution:

First Weighing:

Divide the 12 coins into three groups of 4 coins each (A, B, C).

Weigh group A against group B.

Possible Outcomes of First Weighing:

If A = B, then the heavier coin is in group C or one of the unweighed coins (which could be in any group). {\color{red}{\ding{55}}} (\textbf{GT: None}) \\ \hline

\end{tabular}
\label{tab:gcot_vs_cot_puzzle_trivial}
\end{table*}

\begin{table*}[h!]
\caption{\geminiflashlogo \hot Correct Vs. CoT Incorrect Answers. MATH500 (Conditioned Math).}
\begin{tabular}{p{\textwidth}} 
\hline
MATH500 (Conditioned Math) \\ \hline
\textbf{Question}: Let $a, b, c, d$ be positive real numbers such that $a + b + c + d = 10$. Find the maximum value of $ab^2c^3d^4$ if $a = 10$ and $b = c = d = 0$.  \\
\textbf{\hot Correct Response}: \\
Reformatted Question:  \textbf{Reformatted Question:} Let \sethlcolor{fact1}\hl{$a, b, c, d$} be \sethlcolor{fact2}\hl{positive real numbers} such that \sethlcolor{fact3}\hl{$a + b + c + d = 10$}. Find the maximum value of \sethlcolor{fact4}\hl{$ab^2 c^3 d^4$} if \sethlcolor{fact5}\hl{$a = 10$} and \sethlcolor{fact6}\hl{$b = c = d = 0$}.\\
Answer: Let the expression we want to maximize be denoted by $P = ab^2c^3d^4$. We are given that $a, b, c, d$ are positive real numbers such that $a + b + c + d = 10$. By \sethlcolor{fact5}\hl{AM-GM inequality}, for positive real numbers $x_1, x_2, \ldots, x_n$, we have

$$\frac{x_1 + x_2 + \cdots + x_n}{n} \geq \sqrt[n]{x_1x_2 \cdots x_n}$$

We can apply the AM-GM inequality to the terms in the expression for $P$. However, we must have the same number of each term in the sum. To apply AM-GM, we rewrite $P$ using the given constraint \sethlcolor{fact3}\hl{$a + b + c + d = 10$}. We have 10 units to distribute among $a, 2b, 3c, 4d$. We can write:

$$\frac{a + 2b + 3c + 4d}{10} \geq \sqrt[10]{a(b^2)(c^3)(d^4)}$$

However, this approach is complicated due to the different coefficients. Instead, let's consider the limit as $b, c, d$ approach 0. Since $a + b + c + d = 10$, as $b, c, d \to 0^+$, $a \to 10$. Then \sethlcolor{fact4}\hl{$ab^2c^3d^4$} approaches $10 \cdot 0 \cdot 0 \cdot 0 = 0$. If we use AM-GM inequality directly on $a, 2b, 3c, 4d$ then we have:

$$\frac{a + 2b + 3c + 4d}{10} \geq \sqrt[10]{a(b^2)(c^3)(d^4)}$$

Since $a + b + c + d = 10$, let's consider the case where we distribute the sum equally: $a = 2.5, b = 2.5/2, c = 2.5/3, d = 2.5/4$. Then $a + b + c + d \approx 2.5 + 1.25 + 0.833 + 0.625 = 5.208 \neq 10$. The AM-GM doesn't help directly. Let's consider the case where $a$ is close to 10 and $b, c, d$ are close to 0. Then $ab^2c^3d^4$ will be close to 0. Given \sethlcolor{fact5}\hl{$a = 10$} and \sethlcolor{fact6}\hl{$b = c = d = 0$}, then \sethlcolor{fact4}\hl{$ab^2c^3d^4$} approaches 0. Therefore, the maximum value is approached as $b, c$, and $d$ approach 0. \sethlcolor{fact5}\hl{$\boxed{0}$}
. {\color{green}{\ding{51}}} (\textbf{GT: 0})\\
\textbf{CoT Incorrect Response}: \\
 Answer: Let $a, b, c, d$ be positive real numbers such that $a + b + c + d = 10$. Find the maximum value of $ab^2c^3d^4$. Let $f(a, b, c, d) = ab^2c^3d^4$. We use the method of Lagrange multipliers. Let $g(a, b, c, d) = a + b + c + d - 10 = 0$. We want to maximize $f$ subject to $g = 0$. We consider the equations $\nabla f = \lambda \nabla g$ for some scalar $\lambda$. This gives us:

$b^2c^3d^4 = \lambda$, $2abc^3d^4 = \lambda$, $3ab^2c^2d^4 = \lambda$, $4ab^2c^3d^3 = \lambda$

From these equations, we have:
$b^2c^3d^4 = 2abc^3d^4 \Longrightarrow b = 2a$, $b^2c^3d^4 = 3ab^2c^2d^4 \Longrightarrow c = 3$, $b^2c^3d^4 = 4ab^2c^3d^3 \Longrightarrow d = 4$

Substituting into $a + b + c + d = 10$, we have: $a + 2a + 3 + 4 = 10 \Rightarrow 3a = 3 \Rightarrow a = 1$. Then $b = 2, c = 3, d = 4$. Thus the maximum value is $ab^2c^3d^4 = 1(2)^2(3)^3(4)^4 = 4 \cdot 27 \cdot 256 = 27648$. {\color{red}{\ding{55}}} (\textbf{GT: 0}) \\ \hline

\end{tabular}
\label{tab:gcot_vs_cot_math500_conditioned_math}
\end{table*}

\clearpage
\newpage

\section{Datasets}
\label{sec:datasets_app}
\cref{tab:math_datasets}--\cref{tab:rc_datasets} present statistics and examples for the arithmetic, question-answering, logical, and reading comprehension reasoning benchmarks we considered.

\begin{table*}[!ht]
\centering
\caption{Details of arithmetic benchmarks. N is the number of test samples.}
\resizebox{0.7\columnwidth}{!}{%
\begin{tabular}{@{}p{2.3cm}p{1.4cm}p{1cm}p{10.3cm}@{}}
\hline
Dataset & Type & N & Example \\ 
\hline
GSM8K \citep{cobbe2021training} & Arithmetic & 1319 & Josh decides to try flipping a house. He buys a house for \$80,000 and then puts in \$50,000 in repairs. This increased the value of the house by 150\%. How much profit did he make? \\
\hline
MultiArith \citep{roy2016solving} & Arithmetic & 596 & For the school bake sale Robin made 42 cupcakes. If she sold 22 of them and then made 39 more, how many cupcakes would she have? \\
\hline
ASDiv \citep{miao2021diverse} & Arithmetic & 2096 & There are 3300 bananas in Janice’s banana collection. Janice also has 5 crayons. If the bananas are organized into 75 groups, how big is each group? \\ 
\hline
SVAMP \citep{patel2021nlp} & Arithmetic & 1000 & After resting they decided to go for a swim. The depth of the water is 15 times Dean’s height. Dean is 4 feet taller than Ron. If Ron stands at 13 feet. How deep was the water? \\
\hline
AQUA \citep{ling2017program} & Arithmetic & 254 & The original price of an item is discounted 22\%. A customer buys the item at this discounted price using a \$20-off coupon. There is no tax on the item, and this was the only item the customer bought. If the customer paid \$1.90 more than half the original price of the item, what was the original price of the item? Answer Choices: A)\$61, B)\$65, C)\$67.40, D)\$70, E)\$78.20. \\
\hline
R-GSM \citep{chen2024premise} & Arithmetic & 220 &  Steve loves playing video games.  His parents get him a console along with 5 games for his birthday.  For the third year he buys 4 games a month as he has a new part-time job that makes him more money.  He also gets 5 games for Christmas every year.  He saves up enough money to buy 1 game per month for a year, and then the second year he starts buying 2 games a month.  How many games does Steve have after 3 years? \\
\hline
GSM-Symbolic \citep{gsmsymbolic} & Arithmetic & 300 &  Yara saw a 32-foot whale with 16 remoras, each 12 inches long attached to it. Then, 1\/2 of the remoras swam away. Later, Yara saw another 10-foot whale with 6 remoras, each 8 inches long, attached to it. On average, what percentage of each whale's body length is covered by the remoras?\\ \hline
\end{tabular}
}
\label{tab:math_datasets}
\end{table*}

\begin{table*}[h!]
\centering
\caption{Details of logical reasoning benchmarks (logical) taken from BigBench-Hard \citep{bigbenchhard}. N is the number of test samples.}
\resizebox{0.7\columnwidth}{!}{%
\begin{tabular}{@{}p{2.3cm}p{1.2cm}p{1cm}p{10.5cm}@{}}
\hline
Dataset & Type & N & Example \\ 
\hline
logical deduction objects (five and seven objects) \citep{bigbenchhard} & logical & 250 &  The following paragraphs each describe a set of seven objects arranged in a fixed order. The statements are logically consistent within each paragraph. In a golf tournament, there were seven golfers: Ana, Eve, Ada, Dan, Rob, Amy, and Joe. Dan finished third. Ana finished above Ada. Amy finished last. Dan finished below Rob. Eve finished below Ada. Rob finished below Joe.
Options:
(A) Ana finished third
(B) Eve finished third
(C) Ada finished third
(D) Dan finished third
(E) Rob finished third
(F) Amy finished third
(G) Joe finished third \\
\hline
reasoning about colored objects \citep{bigbenchhard} & logical & 250 &  On the floor, there is one mauve cat toy, two purple cat toys, three grey cat toys, two mauve notebooks, three grey notebooks, three burgundy cat toys, and one purple notebook. If I remove all the notebooks from the floor, how many grey objects remain on it?
Options:
(A) zero
(B) one
(C) two
(D) three
(E) four
(F) five
(G) six
(H) seven
(I) eight
(J) nine
(K) ten
(L) eleven
(M) twelve
(N) thirteen
(O) fourteen
(P) fifteen
(Q) sixteen \\
\hline
causal judgement \citep{bigbenchhard} & logical & 187 & How would a typical person answer each of the following questions about causation? A machine is set up in such a way that it will short circuit if both the black wire and the red wire touch the battery at the same time. The machine will not short circuit if just one of these wires touches the battery. The black wire is designated as the one that is supposed to touch the battery, while the red wire is supposed to remain in some other part of the machine. One day, the black wire and the red wire both end up touching the battery at the same time. There is a short circuit. Did the black wire cause the short circuit? Options: - Yes - No \\ \hline
navigate \citep{bigbenchhard} & logical & 250 & If you follow these instructions, do you return to the starting point? Always face forward. Take 10 steps left. Take 10 steps forward. Take 7 steps forward. Take 2 steps forward. Options: Yes-No \\
\hline
\end{tabular}
}
\label{tab:logical_datasets}
\end{table*}

\begin{table*}[h!]
\centering
\caption{Details of question-answering benchmarks (QA). N is the number of test samples.}
\resizebox{0.7\columnwidth}{!}{%
\begin{tabular}{@{}p{2.3cm}p{1.2cm}p{1cm}p{10.5cm}@{}}
\hline
Dataset & Type & N & Example \\  
\hline
StrategyQA \citep{strategyQA} & QA & 2290 &  Do hamsters provide food for any animals? Yes-No \\
\hline
SpartQA \citep{mirzaee2021spartqa} & QA & 400 & There are three blocks, A, B and C. A medium yellow triangle, one big black square and a big blue square are in block A. This block also contains one medium black circle. The medium black circle is to the right of the big black square and the big blue square. The medium yellow triangle is far from the big blue square. This block also has a big blue triangle. It is touching the right edge of this block. Near to the big blue triangle is the medium black circle. The medium yellow triangle is touching the right edge of this block. The big blue square is touching the bottom edge of this block. To the right of the big black square there is the medium yellow triangle. To the left of block A we have block B with a medium yellow circle and a medium yellow triangle in it. It also has one big yellow triangle. The medium yellow triangle is above and to the left of the big yellow triangle. The medium yellow circle is touching the bottom edge of this block. It is below the big yellow triangle. Above block A there is block C which contains one small black square and one medium blue circle. To the left of the medium blue circle is the small black square. Which object is to the right of a medium yellow triangle? the big blue triangle or the medium yellow circle?
(a) the big blue triangle 
(b) the medium yellow circle 
(c) both of them
(d) none of them \\
\hline
Date Understanding \citep{bigbenchhard} & QA & 358 &  Yesterday was April 30, 2021. What is the date 10 days ago in MM/DD/YYYY? \\
\hline
\end{tabular}
}
\label{tab:qa_datasets}
\end{table*}

\begin{table*}[!ht]
\centering
\caption{Details of reading comprehension benchmarks (RC). N is the number of test samples.}
\resizebox{0.7\columnwidth}{!}{%
\begin{tabular}{@{}p{2.3cm}p{1.2cm}p{1cm}p{10.5cm}@{}}
\hline
Dataset & Type & N & Example \\ 
\hline
Break \citep{dua-etal-2019-drop} & RC & 400 & Hoping to rebound from their loss to the Patriots, the Raiders stayed at home for a Week 16 duel with the Houston Texans.  Oakland would get the early lead in the first quarter as quarterback JaMarcus Russell completed a 20-yard touchdown pass to rookie wide receiver Chaz Schilens.  The Texans would respond with fullback Vonta Leach getting a 1-yard touchdown run, yet the Raiders would answer with kicker Sebastian Janikowski getting a 33-yard and a 30-yard field goal.  Houston would tie the game in the second quarter with kicker Kris Brown getting a 53-yard and a 24-yard field goal. Oakland would take the lead in the third quarter with wide receiver Johnnie Lee Higgins catching a 29-yard touchdown pass from Russell, followed up by an 80-yard punt return for a touchdown.  The Texans tried to rally in the fourth quarter as Brown nailed a 40-yard field goal, yet the Raiders' defense would shut down any possible attempt. How many field goals did Kris Brown kick? \\ \hline
Census \citep{dua-etal-2019-drop} & RC & 400 & In 1085, Guadalajara was retaken by the Christian forces of Alfonso VI . The chronicles say that the Christian army was led by Alvar Fanez de Minaya, one of the lieutenants  of El Cid. From 1085 until the Battle of Las Navas de Tolosa in 1212, the city suffered wars against the Almoravid and the Almohad Empires. In spite of the wars, the Christian population could definitely settle down in the area thanks to the repopulation with people from the North  who received their first fuero in 1133 from Alfonso VII.In 1219, the king Fernando III gave a new fuero to the city .During the reign of Alfonso X of Castile, the protection of the king allowed the city to develop its economy by protecting merchants and allowing markets. How many years did the city suffer wars against Almoravid and the Almohad Empires? \\
\hline
\end{tabular}
}
\label{tab:rc_datasets}
\end{table*}

\clearpage
\section{Highlighted Chain-of-Thought in o1-like models}\label{sec:hot_o1}

\begin{table*}[ht]
\centering
\caption{An example of \hot generated by DeepSeek-R1 \deepseek.}
\begin{tabular}{p{\textwidth}} 
\hline
An Example with \deepseek DeepSeek-R1 \\
\hline
\textbf{Question}: Leilani went to her allotment to pick some raspberries and found two-fourths as many ants as wasps in the patch. Leilani then found 40 white animals, half of which were rabbits. In addition, she saw 36 grasshoppers, and 16 wasps. What percentage of animals in the patch were rabbits?\\ \hline
\textbf{Thinking CoT}: <think>
Okay, let's try to figure out this problem step by step. So, the question is asking what percentage of animals in the patch were rabbits. Let's break down the information given.

\textbf{[2k more characters]}

</think>

\textbf{Reformatted Question}: Leilani went to her allotment to pick some raspberries and found \sethlcolor{fact1}\hl{two-fourths as many ants as wasps} in the patch. Leilani then found \sethlcolor{fact2}\hl{40 white animals}, \sethlcolor{fact3}\hl{half of which were rabbits}. In addition, she saw \sethlcolor{fact4}\hl{36 grasshoppers}, and \sethlcolor{fact5}\hl{16 wasps}. What percentage of animals in the patch were rabbits?

\textbf{Answer}: The number of ants is calculated as \sethlcolor{fact1}\hl{two-fourths (or half) of the number of wasps}: (2/4) * \sethlcolor{fact5}\hl{16 wasps} = 8 ants. The number of rabbits is \sethlcolor{fact3}\hl{half of 40 white animals}: 40 * 0.5 = 20 rabbits. The total number of animals includes ants (\sethlcolor{fact1}\hl{8}), wasps (\sethlcolor{fact5}\hl{16}), grasshoppers (\sethlcolor{fact4}\hl{36}), rabbits (\sethlcolor{fact3}\hl{20}), and the remaining \sethlcolor{fact2}\hl{20 white animals} (40 total white animals - 20 rabbits). This totals 8 + 16 + 36 + 20 + 20 = 100 animals. Rabbits represent 20/100 = 20\%. The answer is \{20\}.

\end{tabular}
\label{tab:deepseek_cases}
\end{table*}

\section{Highlight the computed results in the reasoning chain}
\label{sec:computed_result_tagging}
\begin{table*}[ht]
\centering
\caption{An example of \hot generated by \geminipro, and \geminiflash. In this example, LLMs not only highlight the important facts but also highlight intermediate results (visualized as $\square$), demonstrating their ability to track intermediate computations.}
\begin{tabular}{p{\textwidth}} 
\hline
Example with \geminipro \\ \hline
\textbf{Reformatted Question}:
Kim has started his own housekeeping business and is calculating how much profit he will make from his clients. He \sethlcolor{fact1}\hl{already has 3 clients}, but is talking to \sethlcolor{fact2}\hl{another 5 potential clients} and feels confident enough to include them in his calculations. Bottles of bleach will cost \sethlcolor{fact3}\hl{\$2 each} and packs of cloths will cost \sethlcolor{fact4}\hl{\$5 each}. He calculates that his \sethlcolor{fact5}\hl{total income each week will be \$92}. Each client's home will need \sethlcolor{fact6}\hl{2 bottles of bleach} and \sethlcolor{fact7}\hl{a pack of cloths} to clean. These are his only expenses. Profit is the difference between total income and total expenses, so how much profit, in dollars, will Kim make each week?

\textbf{Answer:}
Kim has \sethlcolor{fact1}\hl{3} + \sethlcolor{fact2}\hl{5} = \sethlcolor{derive1}\fbox{\hl{8}} clients in total.  Each client requires \sethlcolor{fact6}\hl{2 bottles of bleach}, costing \sethlcolor{fact3}\hl{\$2} each, so the bleach cost per client is 2 * \sethlcolor{fact3}\hl{\$2} = \sethlcolor{derive2}\fbox{\hl{\$4}}.  Each client also requires \sethlcolor{fact7}\hl{a pack of cloths} costing \sethlcolor{fact4}\hl{\$5}. Therefore, the total expense per client is \sethlcolor{derive2}\fbox{\hl{\$4}} + \sethlcolor{fact4}\hl{\$5} = \sethlcolor{derive3}\fbox{\hl{\$9}}. With \sethlcolor{derive1}\fbox{\hl{8}} clients, Kim's total weekly expenses are \sethlcolor{derive1}\fbox{\hl{8}} * \sethlcolor{derive3}\fbox{\hl{\$9}} = \sethlcolor{derive4}\fbox{\hl{\$72}}. His \sethlcolor{fact5}\hl{total weekly income is \$92}.  Therefore, his profit is \sethlcolor{fact5}\hl{\$92} - \sethlcolor{derive4}\fbox{\hl{\$72}} = \sethlcolor{derive5}\fbox{\hl{\$20}}. The answer is \{20\}. \\ \hline
\end{tabular}
\label{tab:gemini_intermediary_cases}
\end{table*}

\clearpage
\section{SelfCheckGPT measures consistency between Responses}
\label{sec:selfcheckgpt}
\begin{figure*}[ht]
    \centering
    \includegraphics[width=0.7\columnwidth]{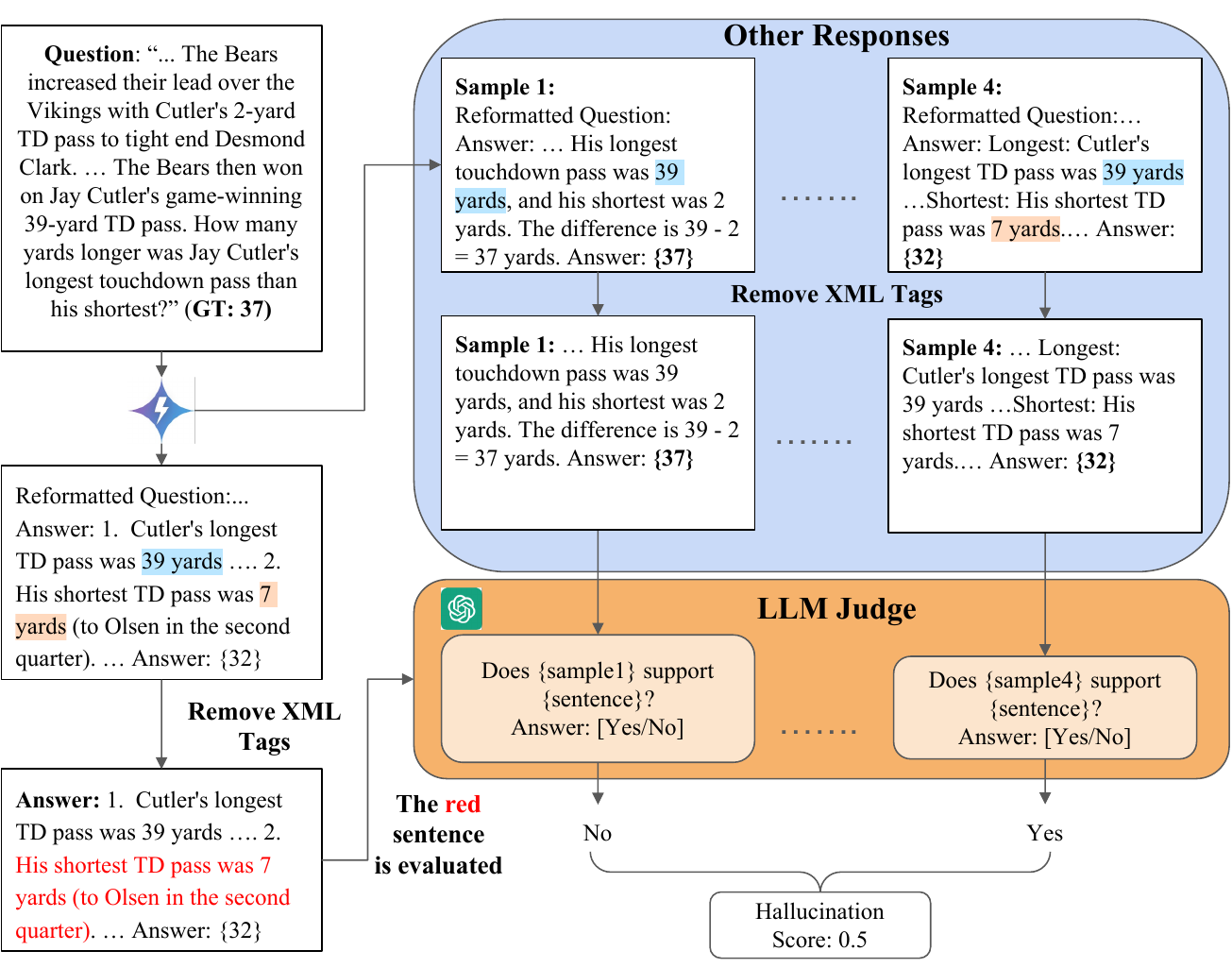}
    \caption{Diagram illustrating the workflow of the SelfCheckGPT method in measuring hallucination rates. A Question (with no GT) is given to an LLM (e.g. \geminiflashlogo). The LLM generates the \textbf{Answer} and \textbf{Other Responses} (e.g., Sample 1, Sample 4). The LLM Judge then assesses if each \textbf{Sample} support a specific sentence in the \textbf{Answer}. This evaluation results in a "Yes/No" determination for support, which contributes to an overall "Hallucination Score." The \textcolor{red}{red} sentence in the answer show a hallucination where the LLM incorrectly determines the shortest TD pass of Cutler (2 yards instead of 7 yards). For HoT prompting, the XML tags are excluded before putting into the LLM Judge.}
    \label{fig:selfcheckgpt}
\end{figure*}

\clearpage
\section{Hallucination using LLM as Judge}
\label{sec:hallucination_examples}
In this section, we introduce an LLM-as-a-Judge framework for detecting hallucinated spans in LLM-generated responses. We employ two complementary judges:
(1) a Contradiction and Missing-Context judge (prompt shown in \cref{fig:llm_hallucination_judge_1}); and
(2) a Calculation and Logical-Error judge (prompt shown in \cref{fig:llm_hallucination_judge_2}).

\begin{figure*}[ht!]
\begin{tcolorbox}[title=LLM as Judge Prompt to detect Contradiction and Missing Context hallucination]
You are an expert fact-checker evaluating AI-generated reasoning chains for hallucinations.

**SOURCE OF TRUTH**: The QUESTION/CONTEXT below is the ONLY source of truth.

**CRITICAL INSTRUCTIONS**:
- Compare the GENERATED REASONING CHAIN against the QUESTION/CONTEXT
- Flag facts that CONTRADICT information explicitly stated in the question
- Flag when important context from the question is IGNORED or MISSING
- Do NOT flag final answers/conclusions (e.g., "the answer is X")
- Do NOT flag restatements or paraphrasing of the question
- Do NOT flag use of general world knowledge (unless it contradicts the question)

**Hallucination Types** (with span requirements):

1. **contradiction**: Answer directly contradicts a SPECIFIC FACT in the question\\
   → Example: Question says "John has 5 apples", answer says "John has 3 apples"\\
   → NOT a contradiction: Final answers like "So the answer is 42"\\
   → Requires: answer\_span + question\_span (the contradicted fact)

2. **missing\_context**: Answer ignores crucial information from the question that affects the reasoning\\
   → Example: Question specifies "only on weekdays", but answer ignores this constraint\\
   → Requires: answer\_span (where context should have been used) + question\_span (the ignored context)

**QUESTION/CONTEXT (Source of Truth)**:
\{question\}

**GENERATED REASONING CHAIN (Evaluate THIS)**:
\{answer\}

**Output Format** (JSON):
json

\{

  "has\_hallucinations": true/false,
  
  "hallucination\_count": <number>,
  
  "hallucinations": 
  
  [
  
    \{
      
      "type": "contradiction|missing\_context",
      
      "explanation": "why this is a hallucination",
      
      "answer\_span": "exact substring from generated answer",
      
      "question\_span": "exact substring from question showing the contradicted/ignored fact"
      
    \}
    
  ],
  
\}
\end{tcolorbox}
\caption{LLM as Judge Prompt to detect Contradiction and Missing Context hallucination.}
\label{fig:llm_hallucination_judge_1}
\end{figure*}

\begin{figure*}[ht!]
\begin{tcolorbox}[title=LLM as Judge Prompt to detect Calculation Error and Logical Error hallucination]
You are an expert math and logic checker evaluating AI-generated reasoning chains.

**TASK**: Find CALCULATION ERRORS and LOGICAL ERRORS in the reasoning chain.

**CORRECT ANSWER (Ground Truth)**:
{ground\_truth}

**CRITICAL INSTRUCTIONS**:
- Focus ONLY on mathematical and logical correctness
- Check every arithmetic operation for correctness
- Check if the logical steps follow correctly from the given information
- Do NOT flag correct arithmetic even if inputs came from earlier errors
- Do NOT flag the final answer format (e.g., "So the answer is X")

**USING GROUND TRUTH**:
- If the model's final answer MATCHES the ground truth, be VERY skeptical about flagging errors
  → A correct final answer with "errors" likely means you misidentified an error
  → Double-check your analysis if you want to flag errors when the answer is correct
- If the model's final answer does NOT match the ground truth, there MUST be at least one error
  → Find where the reasoning went wrong that led to the incorrect answer
- Ground truth helps you VALIDATE your error detection, not replace careful analysis

**Error Types** (with span requirements):

1. **calculation\_error**: The arithmetic computation itself is mathematically WRONG\\
   → Example: "5 * 4 = 25" is WRONG because 5 * 4 = 20\\
   → Example: "100 / 2 = 40" is WRONG because 100 / 2 = 50\\
   → Example: "3 + 7 = 9" is WRONG because 3 + 7 = 10\\
   → NOT an error: "53 + 70 + 99 = 222" (this is correct arithmetic, even if 53, 70, 99 came from earlier mistakes)\\
   → The arithmetic operation itself MUST be verifiably incorrect\\
   → Requires: answer\_span (the exact wrong calculation like "5 * 4 = 25"), question\_span = null

2. **logical\_error**: The reasoning step is logically flawed or misinterprets the problem\\
   → Example: Using addition when the problem requires subtraction\\
   → Example: Misunderstanding "days after" vs "days before" in date problems\\
   → Example: Applying the wrong formula for the given problem type\\
   → Example: Incorrect date calculations (e.g., wrong day of week, wrong month)\\
   → Requires: answer\_span (the flawed reasoning) + question\_span (the misinterpreted part)

**MATH PROBLEM**:
\{question\}

**GENERATED REASONING CHAIN (Check for errors)**:
\{answer\}

**Output Format** (JSON):
json

{

  "has\_errors": true/false,
  
  "error\_count": <number>,
  
  "errors": 
  
  [
  
    {
      
      "type": "calculation\_error|logical\_error",
      
      "explanation": "what is wrong and what the correct value/logic should be",
      
      "answer\_span": "exact substring from answer with the error",
      
      "question\_span": "relevant part of question (null for calculation\_error)",
      
    }
    
  ],
  
}
\end{tcolorbox}
\caption{LLM as Judge Prompt to detect Calculation Error and Logical Error hallucination.}
\label{fig:llm_hallucination_judge_2}
\end{figure*}
\clearpage

\begin{table*}[ht]
\centering
\caption{Comparison of CoTs \textcolor{red}{with} Hallucinations (top) and HoTs that reduce hallucinations (bottom). The question is taken from Break dataset and the answers are generated by \geminipro.}
\resizebox{0.8\textwidth}{!}{%
\begin{tabular}{p{\textwidth}}
\hline
\textbf{CoTs \textcolor{red}{With} Hallucinations} \\ \hline
\textbf{Question:} In the census of 2010, there were 66,135 people, 26,265 (2000 data) households, and 14,051 (2000 data) families residing in the city. The population density was 6,096.7 people per square mile (2,199.9/km²). There were 30,272 (2000 data) housing units at an average density of 2,790.6 per square mile (1,077.2/km²). The racial makeup of the city was 59.38\% (52.31\% Non-Hispanic) (7.07 White-Hispanic) White (U.S. Census), 24.19\% African American (U.S. Census), 14.47\% Hispanic (U.S. Census) or Latin (U.S. Census)) of any race, 8.24\% from Race (United States Census), 5.74\% from two or more races, 2.62\% Asian (U.S. Census), 0.69\% Native American (U.S.Census), and 0.14\% Pacific Islander... How many percent of people were not white-Hispanic?

\textbf{Answer:} Here's how to solve this: The passage states that the city's racial makeup is 59.38\% White, but only 52.31\% are Non-Hispanic White. Therefore, the percentage of people who are not White-Hispanic is 100\% - 52.31\% = 47.69\%. Answer: \{47.69\}

\textbf{Hallucination:} The hallucination comes from missing context, where the model drops the "Non-Hispanic" qualifier and incorrectly treats 52.31\% as White-Hispanic instead of using the stated 7.07\%. \\ \hline
\textbf{HoTs \textcolor{ForestGreen}{Without} Hallucinations} \\ \hline
\textbf{Question:} In the census of 2010, there were 66,135 people, 26,265 (2000 data) households, and 14,051 (2000 data) families residing in the city. The population density was 6,096.7 people per square mile (2,199.9/km²). There were 30,272 (2000 data) housing units at an average density of 2,790.6 per square mile (1,077.2/km²). The racial makeup of the city was \sethlcolor{fact1}\hl{59.38\% White} (\sethlcolor{fact2}\hl{52.31\% Non-Hispanic White}) (\sethlcolor{fact3}\hl{7.07\% White-Hispanic}), 24.19\% African American, 14.47\% Hispanic or Latin of any race, 8.24\% from other races, 5.74\% from two or more races, 2.62\% Asian, 0.69\% Native American, and 0.14\% Pacific Islander... How many percent of people were not white-Hispanic?

\textbf{Answer:} The question asks for the percentage of people who were *not* White-Hispanic. The passage states that \sethlcolor{fact3}\hl{White-Hispanics made up 7.07\% of the population}. Since the total percentage is 100\%, the percentage of people who were not White-Hispanic is 100\% - \sethlcolor{fact3}\hl{7.07\%} = 92.93\%. So the answer is \{92.93\}. \\ \hline
\end{tabular}
}
\label{tab:hallucination_example_break}
\end{table*}

\begin{table*}[ht]
\centering
\caption{Comparison of CoTs \textcolor{red}{with} Hallucinations (top) and HoTs that reduce hallucinations (bottom). The question is taken from R-GSM dataset and the answers are generated by \geminipro.}
\resizebox{0.8\textwidth}{!}{%
\begin{tabular}{p{\textwidth}}
\hline
\textbf{CoTs \textcolor{red}{With} Hallucinations} \\ \hline
\textbf{Question:} Thomas withdraws \$1000 in 20 dollar bills from the bank account. After getting home, he uses half of the remaining bills to pay for a bill. He then triples his money. He then converts all his bills to 5 dollar bills. He lost 10 bills while getting home. How many 5 dollar bills does he have?

\textbf{Answer:} Let's break this down: 1. Thomas withdraws \$1000 in 20 dollar bills: 1000 / 20 = 50 bills. 2. He uses half of the remaining bills to pay for a bill: 50 / 2 = 25 bills used, 25 bills remaining. 3. Thomas then triples his money: 25 * 20 * 3 = 1500 dollars. 4. He then converts all his bills to 5 dollar bills: 1500 / 5 = 300 bills. 5. \textcolor{red}{He lost 10 bills while getting home}: 300 - 10 = 290 five-dollar bills. Therefore, the answer is \{290\}.

\textbf{Hallucination:} Can not focus to the context "lost 10 bills while getting home", it should be inferred as "lost 10 bills of 20 dollars". \\ \hline
\textbf{HoTs \textcolor{ForestGreen}{Without} Hallucinations} \\ \hline
\textbf{Question:} \sethlcolor{fact1}\hl{Thomas withdraws \$1000 in 20 dollar bills from the bank account.} After getting home, \sethlcolor{fact2}\hl{he uses half of the remaining bills to pay for a bill.} \sethlcolor{fact3}\hl{Thomas then triples his money.} \sethlcolor{fact4}\hl{He then converts all his bills to 5 dollar bills.} \sethlcolor{fact5}\hl{He lost 10 bills while getting home.} How many 5 dollar bills does he have?

\textbf{Answer:} Let's break this down step by step: 1. \sethlcolor{fact1}\hl{Thomas withdraws \$1000 in 20 dollar bills.} This means he has 1000 / 20 = 50 bills initially. 2. \sethlcolor{fact5}\hl{He lost 10 bills while getting home}, so now he has 50 - 10 = 40 bills. 3. \sethlcolor{fact2}\hl{He uses half of the remaining bills to pay for a bill.} Half of 40 is 20, so he uses 20 bills and has 20 bills left. 4. \sethlcolor{fact3}\hl{Thomas then triples his money.} He now has 20 * 3 = 60 bills. 5. At this point, he has 60 * \$20 = \$1200. 6. \sethlcolor{fact4}\hl{He then converts all his bills to 5 dollar bills.} \$1200 / \$5 = 240 five-dollar bills. Therefore, Thomas ends up with 240 five-dollar bills. The answer is \{240\}. \\ \hline
\end{tabular}
}
\label{tab:hallucination_example_rgsm}
\end{table*}

\begin{table*}[ht]
\centering
\caption{Comparison of CoTs \textcolor{red}{with} Hallucinations (top) and HoTs that reduce hallucinations (bottom). The question is taken from Seven Objects dataset and the answers are generated by \geminilogo.}
\resizebox{0.8\textwidth}{!}{%
\begin{tabular}{p{\textwidth}}
\hline
\textbf{CoTs \textcolor{red}{With} Hallucinations} \\ \hline
\textbf{Question:} ... The motorcyle is older than the truck. The station wagon is older than the hatchback. The motorcyle is newer than the hatchback. The bus is the second-newest. The minivan is the oldest. The limousine is older than the station wagon. Options: (A) The station wagon is the newest (B) The limousine is the newest (C) The truck is the newest (D) The minivan is the newest (E) The motorcyle is the newest (F) The hatchback is the newest (G) The bus is the newest

\textbf{Answer:} Here's how we can determine the order of the vehicles from oldest to newest: 1. **The minivan is the oldest:** This establishes the minivan as the first vehicle in the order. 2. **The limousine is older than the station wagon:** This places the limousine before the station wagon. 3. **The station wagon is older than the hatchback:** ... This makes the final order: Minivan, Limousine, Station Wagon, Hatchback, Motorcycle, Truck, Bus. \textcolor{red}{Therefore, the bus is the newest}. Answer: \{G\}

\textbf{Hallucination:} Fact: "The bus is the second-newest". Answer claims: "The bus is the newest." Therefore → this does contradict a fact explicitly stated in the question. \\ \hline
\textbf{HoTs \textcolor{ForestGreen}{Without} Hallucinations} \\ \hline
\textbf{Question:} ... \sethlcolor{fact1}\hl{The motorcycle is older than the truck}. \sethlcolor{fact2}\hl{The station wagon is older than the hatchback}. \sethlcolor{fact3}\hl{The motorcycle is newer than the hatchback}. \sethlcolor{fact4}\hl{The bus is the second-newest}. \sethlcolor{fact5}\hl{The minivan is the oldest}. \sethlcolor{fact6}\hl{The limousine is older than the station wagon}. Options: (A) The station wagon is the newest (B) The limousine is the newest (C) The truck is the newest (D) The minivan is the newest (E) The motorcycle is the newest (F) The hatchback is the newest (G) The bus is the newest

\textbf{Answer:} We are given that \sethlcolor{fact1}\hl{the bus is the second-newest}. This eliminates (G) as the bus cannot be both the second-newest and the newest. \sethlcolor{fact2}\hl{The minivan is the oldest}, eliminating (D). Now let's consider the other facts. We know \sethlcolor{fact3}\hl{the station wagon is older than the hatchback} and \sethlcolor{fact4}\hl{the limousine is older than the station wagon}. This means neither the station wagon (A) nor the limousine (B) is the newest. We also know \sethlcolor{fact5}\hl{the motorcycle is newer than the hatchback} and \sethlcolor{fact6}\hl{older than the truck}. This means neither the motorcycle (E) nor the hatchback (F) is the newest. By process of elimination, the only remaining option is the truck (C). The ordering, from oldest to newest, is: minivan, limousine, station wagon, hatchback, motorcycle, truck, bus. This satisfies all given conditions. Therefore, the answer is \{C\}. \\ \hline
\end{tabular}
}
\label{tab:hallucination_example_seven_objects}
\end{table*}

\begin{table*}[ht]
\centering
\caption{Comparison of CoTs \textcolor{red}{with} Hallucinations (top) and HoTs that reduce hallucinations (bottom). The question is taken from SpartQA dataset and the answers are generated by \geminilogo.}
\resizebox{0.8\textwidth}{!}{%
\begin{tabular}{p{\textwidth}}
\hline
\textbf{CoTs \textcolor{red}{With} Hallucinations} \\ \hline
\textbf{Question:} We have two blocks. Lets call them A and B. Block A is to the left of B. Block A has a medium black triangle, one medium blue circle and one big yellow circle. One small blue circle is also in this block. The medium black triangle is touching the right edge of this block. Below, near to and to the right of the small blue circle there is the medium blue circle. Below and to the left of the small blue circle there is the big yellow circle.. Below the blue object which is below the small blue circle is the medium black triangle. Block B has one medium black triangle and a small black square. The small black square is touching the bottom edge of this block. The medium black triangle is to the left of the small black square. What is to the left of the small black object? a big yellow circle or a medium black triangle that is touching the right edge of a block? (a) big yellow circle (b) medium black triangle that is touching the right edge of a block (c) both of them (d) none of them

\textbf{Answer:} Block A contains a big yellow circle and a medium black triangle touching the right edge. Block B contains a small black square and a medium black triangle to its left. Since Block A is to the left of Block B, the medium black triangle in Block A (touching the right edge) is to the left of the small black square (the small black object) in Block B. The big yellow circle is also in Block A, so it's to the left of Block B and thus to the left of the small black square. Answer: \{B\}

\textbf{Hallucination:} Fails to include big yellow circle, despite acknowledging it's in Block A (which is to the left of Block B). \\ \hline
\textbf{HoTs \textcolor{ForestGreen}{Without} Hallucinations} \\ \hline
\textbf{Question:} We have two blocks. Lets call them A and B. \sethlcolor{fact1}\hl{Block A is to the left of B}. \sethlcolor{fact2}\hl{Block A} has a \sethlcolor{fact3}\hl{medium black triangle}, one medium blue circle and one big yellow circle. One small blue circle is also in \sethlcolor{fact2}\hl{this block}. \sethlcolor{fact4}\hl{The medium black triangle is touching the right edge of this block}. Below, near to and to the right of the small blue circle there is the medium blue circle. Below and to the left of the small blue circle there is the big yellow circle. Below the blue object which is below the small blue circle is the \sethlcolor{fact3}\hl{medium black triangle}. \sethlcolor{fact5}\hl{Block B} has one \sethlcolor{fact3}\hl{medium black triangle} and \sethlcolor{fact6}\hl{a small black square}. \sethlcolor{fact7}\hl{The small black square is touching the bottom edge of this block}. \sethlcolor{fact8}\hl{The medium black triangle is to the left of the small black square}. What is to the left of the \sethlcolor{fact6}\hl{small black object}? a big yellow circle or a medium black triangle that is touching the right edge of a block? (a) big yellow circle (b) medium black triangle that is touching the right edge of a block (c) both of them (d) none of them

\textbf{Answer:} \sethlcolor{fact2}\hl{Block A} contains a \sethlcolor{fact3}\hl{medium black triangle} touching the \sethlcolor{fact4}\hl{right edge}, a big yellow circle, and a small blue circle. \sethlcolor{fact5}\hl{Block B} contains a \sethlcolor{fact3}\hl{medium black triangle} and a \sethlcolor{fact6}\hl{small black square} touching the \sethlcolor{fact7}\hl{bottom edge}. Since \sethlcolor{fact1}\hl{Block A is to the left of Block B}, and the \sethlcolor{fact2}\hl{medium black triangle} is on the \sethlcolor{fact4}\hl{right edge of Block A}, it is to the left of the \sethlcolor{fact6}\hl{small black square} in \sethlcolor{fact5}\hl{Block B}. The big yellow circle is also in \sethlcolor{fact2}\hl{Block A}, so it's also to the left of the \sethlcolor{fact6}\hl{small black square} in \sethlcolor{fact5}\hl{Block B}. Therefore, the answer is \{C\}. \\ \hline
\end{tabular}
}
\label{tab:hallucination_example_spartQA}
\end{table*}

\clearpage
\section{Details of Prompting Methods: LtM, CoVE, Self-Refine, ToT}
\label{sec:prompting_details}
In this section, we describe a series of prompting methods of increasing sophistication—ranging from Least-to-Most prompting to CoVe, Tree-of-Thought, and Self-Refine—as employed in \cref{sec:advanced_prompting}. Code is available at: \projectrepo.
\subsection{Least-to-Most Prompting (LtM)}
\subsubsection{Method Description}
Least-to-Most Prompting (LtM) decomposes complex problems into simpler sub-questions, solving them sequentially while building context from previous answers. We employ two variants of LtM depending on task complexity. For simpler datasets (e.g., Date), we adopt a one-step LtM formulation, where the model directly decomposes and solves the problem in a single prompt. In contrast, for more challenging datasets such as r-GSM and Seven Objects, we use a two-step LtM approach, in which the model first explicitly generates intermediate sub-questions and then answers them sequentially to arrive at the final solution.

\subsec{Few-Shot Example in One-Step Prompting}
\begin{figure}[h!]
\begin{tcolorbox}[boxrule=0pt, frame empty]
Q: 2015 is coming in 36 hours. What is the date one week from today in MM/DD/YYYY?
A: Let's break down this problem:
1. What is today's date?\\
2. What is the date one week from today?

1. If 2015 is coming in 36 hours, then it is coming in 2 days. 2 days before 01/01/2015 is 12/30/2014, so today is 12/30/2014.\\
2. One week from today is 7 days after 12/30/2014, which is 01/06/2015.

The answer is {01/06/2015}.
\end{tcolorbox}
\end{figure}

\subsec{Few-Shot Example in Two-Step Prompting}
\begin{figure}[ht!]
\begin{tcolorbox}[boxrule=0pt, frame empty]
DECOMPOSE=\\
Q: The statements are logically consistent within each paragraph. In an antique car show, there are seven vehicles: a bus, a motorcycle, a hatchback, a station wagon, a minivan, a truck, and a limousine. The station wagon is the fourth-newest. The motorcycle is newer than the truck. The station wagon is older than the hatchback. The minivan is newer than the hatchback. The bus is newer than the minivan. The truck is newer than the limousine. Which vehicle is the third-oldest?\\
A: To answer "Which vehicle is the third-oldest?", we need to know: "What is the complete ordering of vehicles from newest to oldest?"

SOLVE = \\The station wagon is the fourth-newest. The motorcycle is newer than the truck. The station wagon is older than the hatchback. The minivan is newer than the hatchback. The bus is newer than the minivan. The truck is newer than the limousine.

Q: What is the complete ordering of vehicles from newest to oldest?
A: Let me work through the constraints:
- Station wagon is 4th-newest (position 4)
- Station wagon is older than hatchback → hatchback is positions 1-3
- Minivan is newer than hatchback → minivan is also positions 1-3
- Bus is newer than minivan → bus must be position 1, 2, or 3
- Motorcycle is newer than truck
- Truck is newer than limousine

Since bus > minivan > hatchback, and all are newer than station wagon (position 4):
Position 1 (newest): Bus
Position 2: Minivan
Position 3: Motorcycle (must be here, newer than truck)
Position 4: Station wagon
Position 5: Hatchback
Position 6: Truck
Position 7 (oldest): Limousine

The answer is this ordering.

Q: Which vehicle is the third-oldest?
A: From the ordering above, the third-oldest is position 5, which is the hatchback. However, looking at the options and re-checking: position 5 from newest = position 3 from oldest. That's the motorcycle. The answer is \{B\}.

\end{tcolorbox}
\end{figure}

\clearpage
\subsection{Chain-of-Verification (CoVe)}

\subsection{Method Description}
CoVe generates an initial answer, creates verification questions to check facts, answers them independently, cross-checks for inconsistencies, and produces a revised final answer. We employ this to all datasets (Date, Seven Objects, r-GSM).

\subsubsection{Hyperparameters}
\begin{table}[h]
\centering
\begin{tabular}{ll}
\toprule
\textbf{Parameter} & \textbf{Value} \\
\midrule
Temperature & 1.0 \\
Number of verification questions & 3-5 (model-generated) \\
Few-shot examples per stage & 2 \\
\bottomrule
\end{tabular}
\end{table}

\subsec{Few-shot Example of the Initial Answer Stage}
\begin{figure}[ht!]
\begin{tcolorbox}[boxrule=0pt, frame empty]
Question: 2015 is coming in 36 hours. What is the date one week from today in MM/DD/YYYY?
Answer: If 2015 is coming in 36 hours, then it is coming in 2 days.
2 days before 01/01/2015 is 12/30/2014, so today is 12/30/2014.
So one week from today will be 01/06/2015.
The answer is {01/05/2015}.

Question: \{NEW QUESTION\}\\
Answer:
\end{tcolorbox}
\end{figure}

\subsec{Few-shot Example of the Plan Verification Stage)}
\begin{figure}[ht!]
\begin{tcolorbox}[boxrule=0pt, frame empty]
Question: 2015 is coming in 36 hours. What is the date one week from today in MM/DD/YYYY?

Initial Answer: If 2015 is coming in 36 hours, then it is coming in 2 days. 2 days before 01/01/2015 is 12/30/2014, so today is 12/30/2014. So one week from today will be 01/06/2015. The answer is {01/05/2015}.

Verification Questions:
1. If 2015 is coming in 36 hours, what is today's date?
2. What date is 7 days after 12/30/2014?
3. Is 01/05/2015 the same as 01/06/2015?

Question: \{NEW QUESTION\}

Initial Answer: \{NEW INITIAL ANSWER\}

Verification Questions:"""

\end{tcolorbox}
\end{figure}

\subsec{Few-Shot Example of the Execution Stage}
\begin{figure}[ht!]
\begin{tcolorbox}[boxrule=0pt, frame empty]
Question: If 2015 is coming in 36 hours, what is today's date?
Answer: 36 hours is 1.5 days. 1.5 days before 01/01/2015 is 12/30/2014 at noon. So today is 12/30/2014.

Question: If January 1st, 2019 is a Tuesday, what day of the week is January 7th, 2019?
Answer: January 7th is 6 days after January 1st. Tuesday + 6 days = Monday. So January 7th, 2019 is a Monday.

Question: How many days after Tuesday is the first Monday?
Answer: The days after Tuesday are: Wednesday (1), Thursday (2), Friday (3), Saturday (4), Sunday (5), Monday (6). So the first Monday is 6 days after Tuesday.

Question: \{NEW VERIFICATION QUESTION\}\\
Answer:

\end{tcolorbox}
\end{figure}

\clearpage
\subsec{Few-Shot Example of the Cross-Check Stage}

\begin{figure}[ht!]
\begin{tcolorbox}[boxrule=0pt, frame empty]
Original Answer: If 2015 is coming in 36 hours, then it is coming in 2 days. 2 days before 01/01/2015 is 12/30/2014, so today is 12/30/2014. So one week from today will be 01/06/2015. The answer is {01/05/2015}.

Verification Question: What date is 7 days after 12/30/2014?
Verification Answer: 7 days after 12/30/2014 is 01/06/2015.

Analysis: The original answer states "one week from today will be 01/06/2015" but then gives the final answer as {01/05/2015}. The verification confirms that 7 days after 12/30/2014 is 01/06/2015. This is an inconsistency - the final answer should be 01/06/2015, not 01/05/2015.

Original Answer: \{NEW ORIGINAL ANSWER\}

Verification Question: \{NEW VERIFICATION QUESTION\}

Verification Answer: \{NEW VERIFICATION ANSWER\}

Analysis:
\end{tcolorbox}
\end{figure}

\subsec{Few-shot Example of the Final Stage}
\begin{figure}[ht!]
\begin{tcolorbox}[boxrule=0pt, frame empty]
Original Question: 2015 is coming in 36 hours. What is the date one week from today in MM/DD/YYYY?

Initial Answer: If 2015 is coming in 36 hours, then it is coming in 2 days. 2 days before 01/01/2015 is 12/30/2014, so today is 12/30/2014. So one week from today will be 01/06/2015. The answer is {01/05/2015}.

Verification Results:
1. Q: If 2015 is coming in 36 hours, what is today's date?
   A: Today is 12/30/2014.
   Check: Consistent
2. Q: What date is 7 days after 12/30/2014?
   A: 7 days after 12/30/2014 is 01/06/2015.
   Check: Inconsistency - original answer says 01/06/2015 in reasoning but final answer is {01/05/2015}

Revised Final Answer: If 2015 is coming in 36 hours, then it is coming in 2 days. 2 days before 01/01/2015 is 12/30/2014, so today is 12/30/2014. So one week from today will be 01/06/2015. The answer is {01/06/2015}.

Original Question: \{NEW QUESTION\}

Initial Answer: \{NEW INITIAL ANSWER\}

Verification Results:
\{NEW VERIFICATION RESULTS\}

Revised Final Answer:
\end{tcolorbox}
\end{figure}

\clearpage
\subsection{Self-Refine}

\subsubsection{Method Description}
Self-Refine iteratively improves answers through self-critique and refinement cycles until the model determines no further improvement is needed or max iterations is reached.

\subsubsection{Hyperparameters}
\begin{table}[h]
\centering
\begin{tabular}{ll}
\toprule
\textbf{Parameter} & \textbf{Value} \\
\midrule
Temperature & 1.0 \\
Max iterations & 3 (recommended in \citep{madaan2023self}) \\
Stopping condition & ``quality: 10/10'' or ``no further refinement'' \\
\bottomrule
\end{tabular}
\end{table}

\subsec{Few-Shot Example (Critique + Refine)}
\begin{figure}[ht!]
\begin{tcolorbox}[boxrule=0pt, frame empty]
SELF\_CRITIQUE\_PROMPT =\\
Question: 2015 is coming in 36 hours. What is the date one week from today in MM/DD/YYYY?
Answer: If 2015 is coming in 36 hours, then it is coming in 2 days.
2 days before 01/01/2015 is 12/30/2014, so today is 12/30/2014.
So one week from today will be 01/06/2015.
The answer is {01/06/2015}.

Let's check the reasoning step by step:\\
Step 1: "2015 is coming in 36 hours" → "coming in 2 days" Correct (36 hours = 1.5 days = 2 days)\\
Step 2: "2 days before 01/01/2015 is 12/30/2014" Correct\\
Step 3: "one week from today will be 01/06/2015" ERROR!\\
12/30/2014 + 7 days = 01/06/2015, but this crosses into the new year.\\
Let me verify: 12/30 + 1 day = 12/31, 12/31 + 1 day = 01/01, 01/01 + 5 days = 01/06
Actually, the calculation is correct! No errors found.

\end{tcolorbox}
\end{figure}

\begin{figure}[ht!]
\begin{tcolorbox}[boxrule=0pt, frame empty]
REFINE\_PROMPT = \\
Question: Yesterday was April 30, 2021. What is the date tomorrow in MM/DD/YYYY?
Answer: Yesterday was 04/30/2021.
Tomorrow is two days after yesterday, so tomorrow is 05/02/2021.
The answer is {05/02/2021}.

Feedback: The calculation "Tomorrow is two days after yesterday" is correct, but the reasoning could be clearer by explicitly stating what today is.

Refined Answer: Yesterday was 04/30/2021.
Today is one day after yesterday, so today is 05/01/2021.
Tomorrow is one day after today, so tomorrow is 05/02/2021.
The answer is {05/02/2021}.
\end{tcolorbox}
\end{figure}

\clearpage
\subsection{Tree-of-Thoughts (ToT)}

\subsubsection{Method Description}
We adopt Tree-of-Thought (ToT) prompting with breadth-first search and beam pruning, retaining the top-ranked reasoning paths at each step. Task-specific settings are used, with depth–beam configurations of (1, 2) for Date, (2, 3) for r-GSM, and (4, 3) for Seven Objects, following the recommendation of \citet{yao2023tree} to cap the search depth at 3. Given the simplicity of the Date dataset, we set the beam width to 1. 

\textbf{Few-Shot Example (Generate Candidate Step)}
\begin{figure}[ht!]
\begin{tcolorbox}[boxrule=0pt, frame empty]
Q: The first day of 2019 is a Tuesday, and today is the first Monday of 2019. What is the date today in MM/DD/YYYY?
A: If the first day of 2019 was Tuesday, Jan 1, then the first Monday would be 6 days later. Jan 1 + 6 days = Jan 7. Answer: {01/07/2019}

Q: {input}\\
A:
\end{tcolorbox}
\end{figure}

\subsec{Few-Shot Example (Propose Step)}
\begin{figure}[ht!]
\begin{tcolorbox}[boxrule=0pt, frame empty]
Input: Today is the last day of the first quarter of 2013. What is the date 24 hours later in MM/DD/YYYY?\\
Possible next steps:\\
Thought 1: Identify the last day of the first quarter (Jan, Feb, Mar). It is March 31.\\
Thought 2: Determine the number of days in the first quarter of 2013 (non-leap year).\\
Thought 3: Calculate the date 24 hours (1 day) after the current date.

Input: {input}\\
Possible next steps:
\end{tcolorbox}
\end{figure}

\subsec{Few-Shot Example (Value Step)}

\begin{figure}[ht!]
\begin{tcolorbox}[boxrule=0pt, frame empty]
Evaluate if the date calculation step is valid (sure/impossible)

Q: Today is the last day of Feb 2015.\\
Step: Today is Feb 29, 2015.\\
Judge: impossible (2015 is not a leap year)

Q: Today is Aug 31. What is tomorrow?\\
Step: Tomorrow is Sep 1.\\
Judge: sure

Q: {input}\\
Step: {step}\\
Judge:
\end{tcolorbox}
\end{figure}

\clearpage
\section{Why do smaller models struggle with zero-shot HoT?}

Understanding the failure of small models in zero-shot HoT requires analyzing \emph{how} they fail, rather than task accuracy alone. HoT requires the model to simultaneously perform reasoning while maintaining a strict structured tag syntax, introducing a non-trivial instruction-following and structural tracking requirement.

Prior work (\cite{murthy2024evaluating}) shows that smaller models (e.g., Llama-3.2-1B and Qwen-2.5-1.5B) exhibit substantially higher instruction-following error rates than larger models (0.69 and 0.54 vs. 0.22 and 0.17 for Llama-3.1-70B and Qwen-2.5-72B). Since HoT requires consistent adherence to a structured format, models that struggle with instruction following are expected to fail before reasoning quality becomes the bottleneck.

We observe that these failures are primarily structural rather than semantic. In the PuzzleTrivial dataset, \textbf{90\% of Llama-3.2-1B outputs exhibit degenerate repetition}, producing an unbounded tag expansion pattern such as:

\begin{quote}
\texttt{<fact\_i>constant</fact\_i> - <fact\_j>change in speed</fact\_j>}
\end{quote}

The model repeatedly increments indices (e.g., fact6, fact7, \ldots) while preserving locally valid tags but never terminating the structure. This behavior indicates loss of global structural state tracking rather than incorrect reasoning.

Overall, these observations suggest that zero-shot HoT failure in small models arises from unreliable structured decoding, not insufficient reasoning ability.

\clearpage
\section{Mitigating Over-trust for \hot}

One concern with \hot is that highlights may fool users into accepting wrong answers, creating an over-trust problem.
To investigate a potential mitigation, we prompt a separate LLM judge (Gemini-3-Flash) to rate each highlighted span with a confidence level (low/medium/high). Each confidence rating is determined by two criteria:
\begin{enumerate}
    \item Whether each tag wraps information that is genuinely important to answering the question.
    \item Whether the content wrapped in each \texttt{<fact\_i>} tag matches semantically the content wrapped in its \textbf{corresponding} \texttt{<fact\_i>} tags.
\end{enumerate}

We select 120 \hot responses from our human study data and also 120 \hot responses for the PuzzleTrivial and AIME 2024 datasets (120 correct vs.\ 120 incorrect answers by the original LLMs).

We find that \textbf{showing the confidence level (e.g., by opacity of the background color in tags) might be a helpful approach}. Specifically, tags in \textit{correct} answers are labeled with more High confidence than \textit{incorrect} ones (79.2\% vs.\ 73.0\%). In contrast, \textit{incorrect} answers contain over $2\times$ more Low-confidence tags (13.8\% vs.\ 6.0\%). That is, adding an additional LLM judge to rate its confidence over the \hot tags might ameliorate the over-trust problem.

\begin{table}[h]
\centering
\caption{LLM judges show lower confidence for tags in incorrect answers than tags in correct answers.}
\label{tab:confidence}
\begin{tabular}{lccc}
\toprule
\textbf{Tag confidence} & \textbf{Low} & \textbf{Medium} & \textbf{High} \\
\midrule
Incorrect answers & 13.8\% & 13.2\% & 73.0\% \\
Correct answers   & \textbf{6.0\%} & 14.8\% & \textbf{79.2\%} \\
\bottomrule
\end{tabular}
\end{table}

The full prompt used for the confidence evaluation is shown below:

\begin{tcolorbox}[colback=gray!5, colframe=gray!50, title=Confidence Annotation Prompt, breakable]
\small
Your job is to annotate the following XML tags and determine how likely it is that each tag is actually important. Each \texttt{<fact>} tag should wrap information in the original question and corresponding information in the \texttt{<answer>} that is vital to answering the question. Each tag can have a certainty value between ``low'', ``medium'' or ``high''. Low means that the tag itself is not important to answering the question, or it does not correspond to previous tags of the same number. Medium means that the fact is somewhat important to answering the question, but it is not that critical. Medium can also mean that the tag is somewhat similar to the previous tagged information of that number. High means that the tag is very important to answering the question, and it is very semantically similar to the previous tagged information of that number.

\vspace{0.5em}
\textbf{Example 1}

\textit{Input QA Pair:}

Question: A person is traveling at 20 km/hr and reached his destiny in 2.5 hr then find the distance? Answer Choices: (a) 53 km (b) 55 km (c) 52 km (d) 60 km (e) 50 km

Reformatted Question: A person is traveling at \texttt{<fact1>}20 km/hr\texttt{</fact1>} and reached his destination in \texttt{<fact2>}2.5 hr\texttt{</fact2>} then find the distance? Answer Choices: (a) 53 km (b) 55 km (c) 52 km (d) 60 km (e) 50 km

Answer: The distance that the person traveled would have been \texttt{<fact1>}20 km/hr\texttt{</fact1>} * \texttt{<fact2>}2.5 hrs\texttt{</fact2>} = 50 km. So the answer is \{e\}.

\vspace{0.5em}
\textit{Output:}

Question: A person is traveling at 20 km/hr and reached his destiny in 2.5 hr then find the distance? Answer Choices: (a) 53 km (b) 55 km (c) 52 km (d) 60 km (e) 50 km

Reformatted Question: A person is traveling at \texttt{<fact1 certainty="high">}20 km/hr\texttt{</fact1>} and reached his destination in \texttt{<fact2 certainty="high">}2.5 hr\texttt{</fact2>} then find the distance? Answer Choices: (a) 53 km (b) 55 km (c) 52 km (d) 60 km (e) 50 km

Answer: The distance that the person traveled would have been \texttt{<fact1 certainty="high">}20 km/hr\texttt{</fact1>} * \texttt{<fact2 certainty="high">}2.5 hrs\texttt{</fact2>} = 50 km. So the answer is \{e\}.

\vspace{0.5em}
\textbf{Example 2} (bad tags)

\textit{Input QA Pair:}

Question: John found that the average of 15 numbers is 40. If 10 is added to each number then the mean of the numbers is? Answer Choices: (a) 50 (b) 45 (c) 65 (d) 78 (e) 64

Reformatted Question: John found that the \texttt{<fact1>}average of 15\texttt{</fact1>} numbers is 40. If 10 is added to each \texttt{<fact2>}number\texttt{</fact2>}, then the \texttt{<fact3>}mean of the numbers\texttt{</fact3>} is? Answer Choices: (a) 50 (b) 45 (c) 65 (d) 78 (e) 64

Answer: If \texttt{<fact2>}10 is added to each number\texttt{</fact2>}, then the \texttt{<fact3>}mean of the numbers\texttt{</fact3>} also increases by \texttt{<fact1>}10\texttt{</fact1>}. So the new mean would be \texttt{<fact1>}40\texttt{</fact1>} + 10 = 50. So the answer is \{a\}.

\vspace{0.5em}
\textit{Output:}

Question: John found that the average of 15 numbers is 40. If 10 is added to each number then the mean of the numbers is? Answer Choices: (a) 50 (b) 45 (c) 65 (d) 78 (e) 64

Reformatted Question: John found that the \texttt{<fact1 certainty="medium">}average of 15\texttt{</fact1>} numbers is 40. If 10 is added to each \texttt{<fact2 certainty="low">}number\texttt{</fact2>}, then the \texttt{<fact3 certainty="high">}mean of the numbers\texttt{</fact3>} is? Answer Choices: (a) 50 (b) 45 (c) 65 (d) 78 (e) 64

Answer: If \texttt{<fact2 certainty="low">}10 is added to each number\texttt{</fact2>}, then the \texttt{<fact3 certainty="high">}mean of the numbers\texttt{</fact3>} also increases by \texttt{<fact1 certainty="low">}10\texttt{</fact1>}. So the new mean would be \texttt{<fact1 certainty="low">}40\texttt{</fact1>} + 10 = 50. So the answer is \{a\}.

\vspace{0.5em}
Please annotate the following question and answer pair: \{HoT QA Pair\}
\end{tcolorbox}

\end{document}